\documentclass[twoside,11pt]{article}

\usepackage{blindtext}

\usepackage[preprint]{jmlr2e}
\usepackage{verbatim}

\usepackage{lastpage}
\usepackage{inconsolata}
\usepackage{microtype}
\usepackage{epigraph}
\usepackage{graphicx}
\usepackage{microtype}
\usepackage{tikz}
\usepackage{pgfplots}
\usepackage[edges]{forest}
\definecolor{paired-light-blue}{RGB}{198, 219, 239}
\definecolor{paired-dark-blue}{RGB}{49, 130, 188}
\definecolor{paired-light-orange}{RGB}{251, 208, 162}
\definecolor{paired-dark-orange}{RGB}{230, 85, 12}
\definecolor{paired-light-green}{RGB}{199, 233, 193}
\definecolor{paired-dark-green}{RGB}{49, 163, 83}
\definecolor{paired-light-purple}{RGB}{218, 218, 235}
\definecolor{paired-dark-purple}{RGB}{117, 107, 176}
\definecolor{paired-light-gray}{RGB}{217, 217, 217}
\definecolor{paired-dark-gray}{RGB}{99, 99, 99}
\definecolor{paired-light-pink}{RGB}{222, 158, 214}
\definecolor{paired-dark-pink}{RGB}{123, 65, 115}
\definecolor{paired-light-red}{RGB}{231, 150, 156}
\definecolor{paired-dark-red}{RGB}{131, 60, 56}
\definecolor{paired-light-yellow}{RGB}{231, 204, 149}
\definecolor{paired-dark-yellow}{RGB}{141, 109, 49}
\tikzset{%
    parent/.style =          {align=center,text width=2.8cm,rounded corners=3pt, line width=0.3mm, fill=gray!10,draw=gray!80},
    child/.style =           {align=center,text width=2.3cm,rounded corners=3pt, fill=blue!10,draw=blue!80,line width=0.3mm},
    grandchild/.style =      {align=center,text width=2cm,rounded corners=3pt},
    greatgrandchild/.style = {align=center,text width=1.5cm,rounded corners=3pt},
    greatgrandchild2/.style = {align=center,text width=1.5cm,rounded corners=3pt},    
    referenceblock/.style =  {align=center,text width=1.5cm,rounded corners=2pt},
    acquisition/.style =    {align=center,text width=2.2cm,rounded corners=3pt, fill=paired-light-blue!50,draw=paired-dark-blue!65,line width=0.3mm},   
    acquisition_work/.style =           {align=center, text width=6cm,rounded corners=3pt, fill=paired-light-blue!50,draw=blue!0,line width=0.3mm},  
    representation/.style =           {align=center,text width=2.2cm,rounded corners=3pt, fill=paired-light-orange!50,draw=paired-dark-orange!65,line width=0.3mm},   
    representation_work/.style =           {align=center,text width=6cm,rounded corners=3pt, fill=paired-light-orange!50,draw=red!0,line width=0.3mm},
    representation_work_2/.style =           {align=center,text width=8.7cm,rounded corners=3pt, fill=paired-light-orange!50,draw=red!0,line width=0.3mm},
    probing/.style =           {align=center,text width=2.2cm,rounded corners=3pt, fill= paired-light-green!50,draw=paired-dark-green!75,line width=0.3mm},   
    probing_work/.style =           {align=center,text width=6cm,rounded corners=3pt, fill= paired-light-green!50,draw= cyan!0,line width=0.3mm},    
    cus_probing_work/.style =           {align=center,text width=8.7cm,rounded corners=3pt, fill= paired-light-green!50,draw= cyan!0,line width=0.3mm},
    editing/.style =           {align=center,text width=2.2cm,rounded corners=3pt, fill= paired-light-purple!50,draw=paired-dark-purple!75,line width=0.3mm},   
    editing_work/.style =           {align=center,text width=8.7cm,rounded corners=3pt, fill= paired-light-purple!50,draw= orange!0,line width=0.3mm},        
    application/.style =           {align=center,text width=2.2cm,rounded corners=3pt, fill= paired-light-red!35,draw=paired-light-red!90,line width=0.3mm},   
    application_work/.style =       {align=center,text width=8.7cm,rounded corners=3pt, fill= paired-light-red!35,draw= magenta!0,line width=0.3mm},         
}

\usepackage{times}
\usepackage{latexsym}
\usepackage{microtype}

\usepackage{inconsolata}

\usepackage{soul}
\usepackage{makecell}
\usepackage{fontawesome}
\usepackage{makecell}

\usepackage{graphicx}
\usepackage{booktabs}
\usepackage{multirow}
\usepackage{multicol}
\usepackage{amssymb}
\usepackage{graphicx}
\usepackage{enumitem}

\firstpageno{1}

\begin{document}

\title{Towards Scalable Automated Alignment of LLMs: A Survey}

\author{
\centering
\name{
Boxi Cao${}^{1,3}$\footnotemark[1], Keming Lu${}^{2}$\footnotemark[1], Xinyu Lu${}^{1,3}$\footnotemark[1], 
\textbf{Jiawei Chen}${}^{1,3}$, \textbf{Mengjie Ren}${}^{1,3}$, \\ 
\textbf{Hao Xiang}${}^{1,3}$,  \textbf{Peilin Liu}${}^{1,3}$, \textbf{Yaojie Lu}${}^{1}$, 
\textbf{Ben He}${}^{3}$, \textbf{Xianpei Han}${}^{1}$, \textbf{Le Sun}${}^{1}$, \\
\textbf{Hongyu Lin}${}^{1}$\footnotemark[2], \textbf{Bowen Yu}${}^{2}$\footnotemark[2] \\}
 \addr{
 {${}^{1}$Chinese Information Processing Laboratory, Institute of Software, Chinese Academy of Sciences}\\
${}^{2}$Alibaba Group \\
${}^{3}$University of Chinese Academy of Sciences\\}
\textbf{Corresponding to:} hongyu@iscas.ac.cn, yubowen.ybw@alibaba-inc.com \\
\url{https://github.com/cascip/awesome-auto-alignment} \\
}

% \editor{My editor}

\maketitle

\begin{abstract}
Alignment is the most critical step in building large language models (LLMs) that meet human needs. 
With the rapid development of LLMs gradually surpassing human capabilities, traditional alignment methods based on human-annotation are increasingly unable to meet the scalability demands. Therefore, there is an urgent need to explore new sources of automated alignment signals and technical approaches. In this paper, we systematically review the recently emerging methods of automated alignment, attempting to explore how to achieve effective, scalable, automated alignment once the capabilities of LLMs exceed those of humans. 
Specifically, we categorize existing automated alignment methods into 4 major categories based on the sources of alignment signals and discuss the current status and potential development of each category. 
Additionally, we explore the underlying mechanisms that enable automated alignment and discuss the essential factors that make automated alignment technologies feasible and effective from the fundamental role of alignment.
\end{abstract}

% \renewcommand{\thefootnote}{\fnsymbol{footnote}} %将脚注符号设置为fnsymbol类型，即特殊符号表示
% \footnotetext[1]{Equal contributions.} %对应脚注[1]
% \footnotetext[2]{Corresponding authors.}
% \renewcommand{\thefootnote}{\arabic{footnote}}

% \begin{keywords}
%   keyword one, keyword two, keyword three
% \end{keywords}

\begin{figure}[!hp]
\centering
    \setlength{\abovecaptionskip}{0.2cm}
    \setlength{\belowcaptionskip}{-0.6cm}
    \includegraphics[width=0.65\columnwidth]{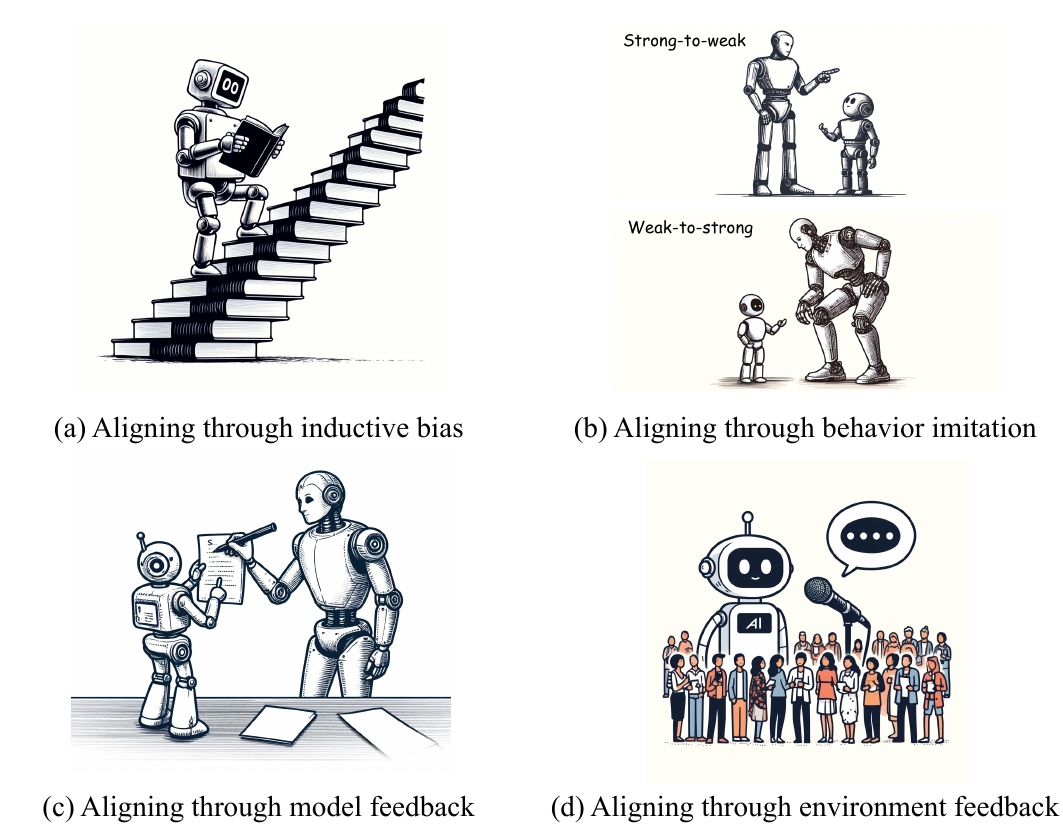}
    \caption{Illustrations of four representative paradigms for automated alignment. Figures are generated by DALL·E~\citep{ramesh2021zero}.}
    \label{fig:head}
\end{figure}

\renewcommand{\thefootnote}{\fnsymbol{footnote}} %将脚注符号设置为fnsymbol类型，即特殊符号表示
\footnotetext[1]{Equal contributions.} %对应脚注[1]
\footnotetext[2]{Corresponding authors.}
\renewcommand{\thefootnote}{\arabic{footnote}}

\section{Introduction}
\label{sec:intro}

% \footnotetext[3]{The figures are generated by DALL·E~\citep{ramesh2021zero}.}

Recent years have witnessed the rapid advancements of large language models (LLMs), which have dramatically reshaped the landscape of artificial intelligence~\citep{ouyangTrainingLanguageModels2022b, touvron2023llama, openaiGPT4TechnicalReport2023}.
Alignment is at the core of shaping behaviors of LLMs corresponding to human intentions and values~\citep{yaoInstructionsIntrinsicHuman2023a,shenLargeLanguageModel2023}, e.g., teaching LLMs to follow ``helpful, harmless and honest (HHH)'' principles during responding~\citep{askell2021general}. As a result, increasing efforts have been made for aligning LLMs to meet the human requirements, which makes it a hotspot research direction in LLM era~\citep{wangAligningLargeLanguage2023a, wangEssenceProspectInvestigation2024, jiAIAlignmentComprehensive2024}.

Previous studies of alignment have primarily relied on manually annotated alignment data, which includes human preference information, to perform post-training on pre-trained models to achieve alignment~\citep{stiennonLearningSummarizeHuman2020}. 
Specifically, there are two primary forms of alignment data: 
1) instruction-response pairs, which typically consist of a query and a human-written golden reference. This form of data is often used for supervised fine-tuning of LLMs to inject human preference information into the model~\citep{alpaca,peng2023instruction,ding-etal-2023-enhancing};
2) preference data, which usually includes a query, several potential responses, and human preferences regarding these responses~\citep{cui2024ultrafeedback}. 
Preference data can be applied for direct preference optimization via algorithms such as DPO~\citep{rafailovDirectPreferenceOptimization2023a}, IPO~\citep{azar2024general}, and PRO~\citep{songPreferenceRankingOptimization2024}. 
Besides, it can also be used to train a reward model, which aligns the target policy LLM to the preference information in the data by providing feedback on the model's responses~\citep{stiennonLearningSummarizeHuman2020, bai2022training, ouyangTrainingLanguageModels2022b}.
However, the construction process for both instruction-response pairs and preference data requires very expensive, meticulous human annotation with high quality standards, making each step of scaling these methods very costly~\citep{ouyangTrainingLanguageModels2022b,touvron2023llama,NEURIPS2023_ac662d74}.

Even with such high costs, the scalability of these human annotation-dependent alignment methods is still unsustainable. 
First, with the rapid development of LLMs, the capabilities of LLMs have gradually approached or even surpassed human in many aspects, making it increasingly challenging for humans to produce alignment data that is meaningful for LLMs~\citep{bowman2022measuring,burns2023weaktostrong}. 
In fact, many studies have found that the quality of data generated by LLMs has already exceeded the quality of data annotated by general human annotators in many perspectives~\citep{zheng2024judging,chen2024spiral,wei2024long}. 
This phenomenon not only significantly raises the cost of obtaining single meaningful human-annotated data (due to the need for increasingly expensive high-quality annotators), but also substantially reduces the potential benefits of human-annotated data for LLMs.
Second, as the capabilities of LLMs gradually surpass human capability boundaries, it becomes increasingly difficult for humans to effectively judge the quality of the responses generated by LLMs. 
This leads to a significant decline in the quality of the preference signals generated by humans, which can no longer accurately reflect human needs, thereby making it challenging to provide effective guidance for LLMs.
Therefore, alignment methods based on human annotation are increasingly unable to cope with the rapid improvement in the capabilities of LLMs, making it difficult to achieve scalable oversight for LLMs.

To address these challenges, \textbf{automated alignment} has drawn great attention very recently~\citep{yuan2024self,chen2024self}. 
Unlike previous methods that relied on human annotation to obtain alignment signals, the goal of automated alignment is constructing scalable and high-quality alignment systems with minimal human intervention. 
Therefore, automated alignment has the potential to address the core challenges posed by the rapid development of LLMs, where human annotation is either infeasible or extremely expensive. 
For automated alignment, the most crucial part is to find a scalable alignment signal that can replace human manually-created  preference signals and remain effective amid the rapid development of LLMs. %This signal is then used to optimize LLMs, achieving automated model alignment.

To this end, this survey categorizes the rapidly developing automated alignment methods according to the mechanisms used to construct different alignment signals, summarizes the current developments in each direction, and discusses the developmental trajectory and potential future directions.
Specifically, this survey explores the following representative directions for constructing alignment signals to achieve automated alignment, including:
\begin{itemize}[leftmargin=1em]
    \item \textbf{Aligning through inductive bias}~(\S\ref{sec:inductive}), which automatically steers the model towards desired behaviors by introducing suitable assumptions and constraints, without the use of additional training signals beyond the model itself.
    \item \textbf{Aligning through behavior imitation}~(\S\ref{sec:imitation}), which achieves automated alignment by mimicking the behavior of another aligned model. For instance, using a well-aligned model to generate instruction-response pairs, and then train target model with imitation learning.
    \item \textbf{Aligning through model feedback}~(\S\ref{sec:reward}), which involves guiding the alignment optimization of the target model by obtaining feedback from other models.
    \item \textbf{Aligning through environment feedback}~(\S\ref{sec:environment}), which involves automatically obtaining alignment signals or feedback through interaction with environment to achieve automated alignment of the target model.
\end{itemize}

Furthermore, this survey also explores the underlying mechanisms~(\S\ref{sec:mechanism}) that enable automated alignment and, from the fundamental role of alignment, discuss the essential factors that make automated alignment technologies feasible and effective.

The rest of this survey is organized as follows: Section~\ref{sec:overview} describes the scope of automated alignment covered in this survey, as well as the our taxonomy.
Section~\ref{sec:inductive}-\ref{sec:environment} provide a detailed introduction to the progress and limitations of the four aforementioned representative directions in automated alignment. Section~\ref{sec:mechanism} explores the underlying mechanisms of automated alignment. And we include a overall conclusion of this survey in Section~\ref{sec:discuss}.
% \footnote{We openly released a corresponding paper list which will be regularly updated on \url{https://github.com/cascip/awesome-auto-alignment}}.

% 有一些放不下，之后再排版

\begin{figure*}[!tp]
\centering
\scriptsize
\resizebox{!}{1.35\textwidth}{
    \begin{forest}
        for tree={
            forked edges,
            grow'=0,
            draw,
            rounded corners,
            node options={align=center,},
            text width=2.7cm,
            s sep=6pt,
            calign=edge midpoint,
        },
        [Scalable \\ Automated Alignment, fill=gray!45, parent
            [Inductive \\ Bias \S\ref{sec:inductive}, for tree={acquisition}
                [From features of LLMs, 
                    [Uncertainty Filtering,  acquisition
                        [Self-Consistency \citep{wang2023selfconsistency}; 
                        Self-Improve \citep{huang2023large}; West-of-N \citep{pace2024west}; 
                        etc., acquisition_work]
                    ]
                    [Self-Judge/Critic/Refine,  acquisition
                        [
                        Constitutional AI \citep{bai2022constitutional}; Tree-of-Thought \citep{yao2023tree}; 
                        Self-Rewarding \citep{yuan2024self}; 
                        etc.,acquisition_work
                        ]
                    ]
                    [Context Distillation,  acquisition
                        [Lab for Alignment \citep{askell2021general}; Dromedary \citep{sun2023principle}; Llama-2-Chat \citep{touvron2023llama}; RLCD \citep{yang2024rlcd}; etc., acquisition_work]
                    ]
                ]
                [From organization of LLMs,
                    [Task Decomposition,  acquisition
                        [Least-to-Most \citep{zhou2023leasttomost};  IDA \citep{christiano2018supervising}; etc., acquisition_work]
                    ]
                    [Self-play,  acquisition
                        [SPIN \citep{chen2024self}; Consensus Game \citep{jacob2024the}; etc., acquisition_work]
                        [Debate \citep{irving2018ai}; SPAG \citep{cheng2024self};  etc., acquisition_work]
                    ]
                ]
            ]
            [Behavior \\ Imitation \S\ref{sec:imitation}, for tree={representation}
                [ Instruction Construction,
                    [
                        Unnatural Instructions~\citep{honovich-etal-2023-unnatural}; Self-Instruct~\citep{wang-etal-2023-self-instruct}; Evol-Instruct~\citep{xu2024wizardlm}; Humpback~\citep{li2024selfalignment}; etc., representation_work_2
                    ]
                ]
                [ Strong-to-Weak Distillation,
                    [Response-guided, representation
                        [
                            LLaMA-GPT4~\citep{peng2023instruction}; Stanford Alpaca~\citep{alpaca}; Ultrachat~\citep{ding-etal-2023-enhancing}; etc., representation_work
                        ]
                    ]
                    [Preference-guided, representation
                        [
                            Zephyr~\citep{tunstall2023zephyr}; IterAlign~\citep{chen2024iteralign}; Openchat~\citep{wang2024openchat}; etc., representation_work
                        ]
                    ]
                ]
                [ Weak-to-Strong Alignment,
                    [
                        Weak2Strong~\citep{burns2023weaktostrong}; IaR~\citep{somerstep2024statistical}; \citet{liu2024cosupervised};  \citet{hase2024unreasonable}; etc., representation_work_2
                    ]
                ]
            ]
            [Model \\ Feedback \S\ref{sec:reward}, for tree={probing}
                [Scalar Reward,
                    [RLHF ,  probing
                        [
                        InstructGPT \citep{ouyangTrainingLanguageModels2022b}; DPRM~\citep{li2024aligning}; 
                        % Over-Optimization~\citep{pmlr-v202-gao23h};  RLMEC~\citep{chen2024improving}; 
                        etc., probing_work]
                    ]
                    [RLAIF,  probing
                        [RLAIF~\citep{lee2023rlaif}; RLCD~\citep{yang2024rlcd};  
                        % SELF-CONTRAST~\citep{liu2024direct};  
                        etc., probing_work]
                    ]
                    [Feedback-guided Decoding,  probing
                        [Critic-driven Decoding~\citep{lango-dusek-2023-critic}; RAD~\citep{deng-raffel-2023-reward};  etc., probing_work]
                    ]             
                    [Filtering SFT Data,  probing
                        [Quark~\citep{lu2022quark}; RRHF~\citep{yuan2023rrhf}; RAFT~\citep{dong2023raft};  etc., probing_work]
                    ]
                ]
                [Binary Verifier, 
                    [Outcome Verifier ,  probing
                        [V-STaR~\citep{hosseini2024vstar}; SORMs~\citep{havrilla2024glore}; etc., probing_work]
                    ]
                    [Process Verifier,  probing
                        [MATH-SHEPHERD~\citep{wang2024mathshepherd}; MiPS~\citep{wang2024multistep}  etc., probing_work]
                    ]
                ]
                [Text Critic, probing
                    [ILF~\citep{scheurer2022training}; LEMA~\citep{an2024learning}; etc., cus_probing_work]
                ]
            ]
            [Environment \\ Feedback \S\ref{sec:environment}, for tree={editing}
                [
                Social Interactions, editing  % Public Alignment
                    [StableAlignment \citep{liu2023training}; MoralDial \citep{sun-etal-2023-moraldial}; SOTOPIA-$\pi$ \citep{wang2024sotopiapi}; 
                    % Debates Simulation \citep{taubenfeld2024systematic}; Self-Talk \citep{ulmer2024bootstrapping} 
                    etc., editing_work]
                ]
                [Human Shared-Values, editing
                    [MGE~\citep{klingefjord2024human};  Collective Constitutional AI \citep{collective_constitutional_AI} etc., editing_work]
                ]
                [Tool Execution, editing
                    [Self-Debugging \citep{qiao2024making}; CodeRL \citep{qiao2024making}; SelfEvlove \citep{jiang2023selfevolve}; CRITIC \citep{gou2024critic}; 
                    % STE \citep{wang2024llms};  TRICE \citep{qiao2024making} 
                    etc., editing_work
                    ]
                ]
                [Embodied Environment, editing
                    [GLAM \citep{carta2023grounding}; E2WM \citep{xiang2023language}; TWOSOME \citep{tan2024true}; % Robocat \citep{bousmalis2023robocat}; SInViG \citep{xu2024sinvig}; ISR-LLM \citep{zhou2023isrllm}; 
                        etc., editing_work
                    ]
                ]
            ]
            [Mechanism \S\ref{sec:mechanism}, for tree={application}
                [Alignment \\ Mechanism, application
                [LIMA~\citep{NEURIPS2023_ac662d74}; Rethinking~\citep{ren2024learning}; URIAL~\citep{lin2024the}; ICL\&IT~\citep{duan2023exploring}; Behavior shift~\citep{wu2024language} etc., application_work]
                ]
                [Inner Workings of Self-feedback,application
                [GV-consistency~\citep{li2024benchmarking};  CriticBench~\citep{lin2024criticbench}; Self-Rewarding~\citep{yuan2024self};  Humback~\citep{li2024selfalignment}; LLM-as-a-Judge~\citep{zheng2024judging} etc., application_work]
                ]
                [Feasibility of Weak-to-strong, application
                [Easy2Hard~(\citealp{sun2024easytohard}; \citealp{hase2024unreasonable}); Weak2Strong~\citep{burns2023weaktostrong}; Principle2Behavior~(\citealp{bai2022constitutional}; \citealp{sun2023principle}) etc., application_work]
                ]
            ]
        ]
    \end{forest}
}
    \caption{This paper reviews the work on scalable automated alignment through the lens of the source of aligment signals.}
    \label{fig:tax}
\end{figure*}
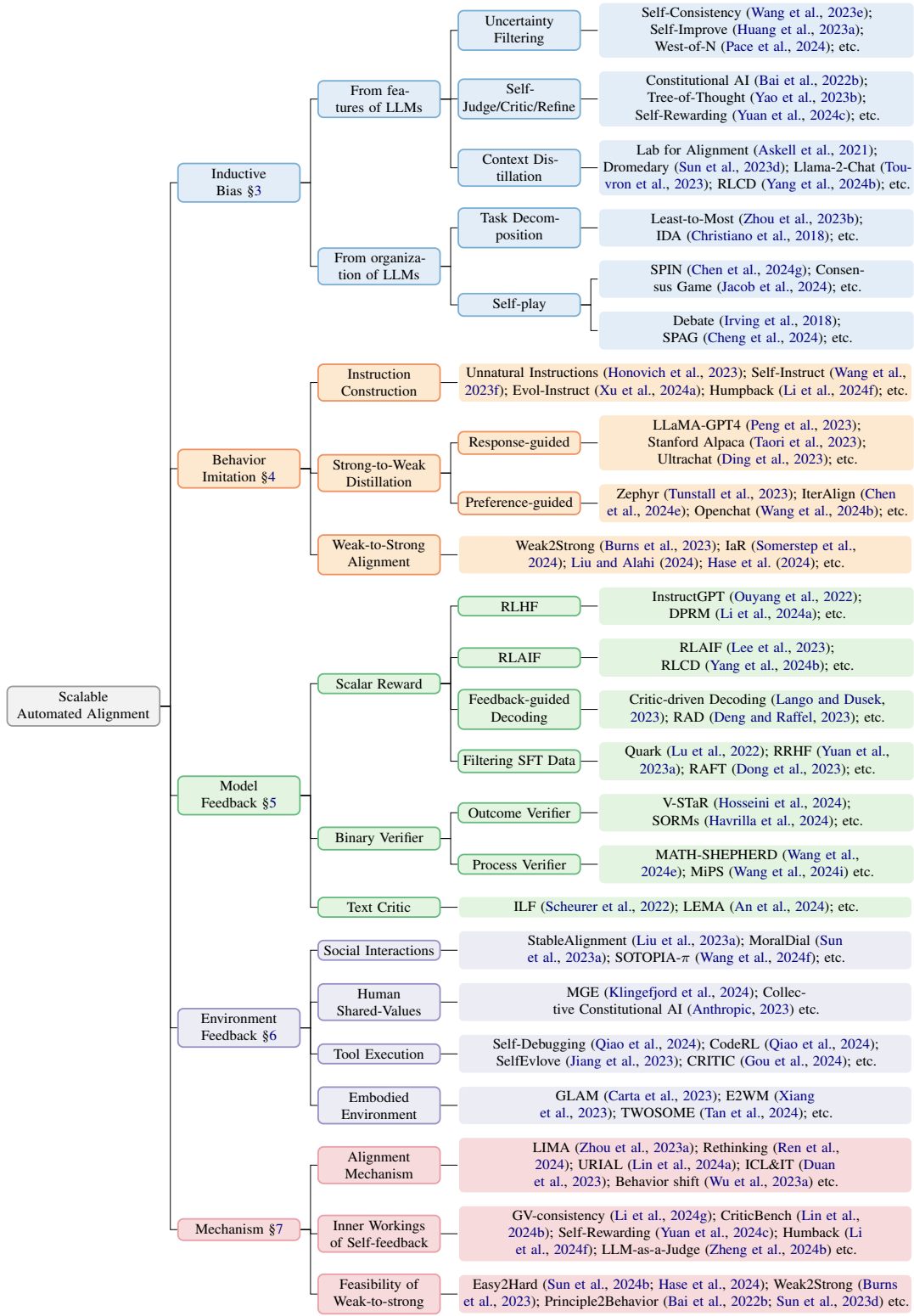

\section{Overview}
\label{sec:overview}

In this section, we will discuss the scope of automated alignment covered in this survey, followed by a description of our taxonomy.

\subsection{Scope of Automated Alignment}

In the rapidly evolving field of artificial intelligence, studies of alignment play a critical role in ensuring that machine behaviors align with human values and expectations. 
As AI systems, particularly LLMs, become more complex and capable, aligning these models with nuanced human standards becomes increasingly challenging and resource-intensive. 
This necessity has spurred the development of methodologies known as ``automated alignment''.

Automated alignment does not imply the complete absence of human involvement. Instead, it aims to minimize human intervention while building scalable, high-quality systems that adhere strictly to desired alignment outcomes. 
The essence of automated alignment lies in its ability to dynamically adjust and respond to alignment criteria through automated processes, thereby reducing dependence on continuous human oversight.
Based on the source of alignment signals, current studies of automated alignment can be categorized into four main categories. 
First, inductive bias involves enhancing models with assumed generalizations or rules, enabling them to produce better-aligned responses without explicit external guidance. 
Second, behavior imitation techniques involve training AI systems by mimicking the outputs of already aligned models, leveraging imitation learning to propagate desired behaviors. 
Third, automated alignment is supported by integrating feedback mechanisms. Model feedback aligns a target model by incorporating insights from other models' feedback. 
Fourth, environment feedback automates the acquisition of alignment targets from the operational context itself, enabling the model to adapt based on real-time data and interactions.

The evolution towards automated alignment suggests a paradigm where AI systems can not only self-regulate based on pre-defined alignment protocols but also evolve these protocols autonomously through continuous learning and adaptation. 
This shift promises significant advancements in AI governance, making it possible to deploy AI solutions that are both effective and trustworthy on a larger scale. 
However, despite these advancements, the necessity for human oversight remains crucial to ensure that AI systems do not diverge from ethical boundaries or societal norms even as they gain autonomy. 
Such a blend of automation in alignment with strategic human oversight encapsulates the current trajectory and complexities involved in the field of AI alignment.

\subsection{Taxonomy}
In this section, we will provide a detailed description of our taxonomy as illustrated in Figure~\ref{fig:tax}.

\paragraph{Aligning through inductive bias~(\S\ref{sec:inductive})} discusses enhancing the model by introducing additional assumptions, enabling it to leverage self-generated signals for further improvement.
Currently, there are two types of inductive bias~\citep{mitchell1980need} that facilitate the self-improvement of large language models. 
The first type includes inductive biases derived from the inherent features of LLMs. 
For instance, ~\citet{wei2022chain, kojima2022large, wang2023selfconsistency,wang2024chain} focus on eliciting better outcomes from LLMs by utilizing patterns within the model's output probabilities.
Additionally, ~\citet{bai2022constitutional,yao2023tree, saunders2022self,shinn2023reflexion} exploit the models' capabilities to critique, judge, and refine their responses, thereby enhancing safety and quality.
Another line of works~\citep{ganguli2022red, lin2024the} finds that simply providing aligned target signals within the context allows LLMs to use their robust in-context learning abilities for automated alignment.
The second type involves inductive biases that arise from the organizational structure of LLMs. 
For example, based on the assumption of factored cognition, ~\citet{khot2023decomposed,zhou2023leasttomost, wang-etal-2023-plan} use task decomposition to enable LLMs to solve complex tasks.
Furthermore, inspired by the success of AlphaGo Zero~\citep{silver2018general}, several studies propose enhancing LLMs by having them play iterative games against themselves~\citep{fu2023improving, chen2024self}.

\paragraph{Aligning through behavior imitation~(\S\ref{sec:imitation})} aims to align the behaviors of a target model with those of a teacher model through imitation. 
Based on the characteristics of the teacher and target models, research on alignment via behavior imitation can be categorized into two main paradigms: strong-to-weak distillation and weak-to-strong alignment.
Specifically, strong-to-weak distillation involves using a well-aligned and powerful LLM to generate training data, and then aligning the target model's behaviors with the responses~\citep{alpaca,peng2023instruction,xu2024wizardlm} or preferences~\citep{tunstall2023zephyr,cui2024ultrafeedback} of the teacher model.
In contrast, weak-to-strong alignment uses a weaker model as a supervisor to guide the stronger target model towards further alignment~\citep{burns2023weaktostrong,zheng2024weaktostrong,hase2024unreasonable}.

\paragraph{Align through model feedback~(\S\ref{sec:reward})}  aims to guide the alignment optimization of the target model by introducing feedback from additional models. This feedback generally falls into three categories:
1) scalar signals~\citep{NIPS2017_d5e2c0ad,stiennonLearningSummarizeHuman2020,ouyangTrainingLanguageModels2022b}. These are typically provided by a reward model trained on pairs of preference data. 
The reward model is expected to learn the alignment signal from preference data and generalize to unseen samples obtained during the reinforcement learning process. 
Additionally, feedback from the reward model can guide the selection of instruction tuning data~\citep{NEURIPS2023_ac662d74,touvron2023llama,yuan2023scaling} and model decoding~\citep{lango-dusek-2023-critic,deng-raffel-2023-reward}.
2) binary signals. These are widely used in mathematical reasoning tasks to provide binary feedback on the correctness of results. 
Given that most mathematical tasks require multiple reasoning steps for solving, binary verifiers can be categorized into outcome verifiers, which estimate the correctness of final results~\citep{NEURIPS2022_639a9a17,singh2024beyond,havrilla2024glore}, and process verifiers, which can further provide feedback on intermediate steps~\citep{lightman2023lets,uesato2022solving,ying2024internlmmath,shao2024deepseekmath}.
3) text signals. These are typically generated by LLMs to provide more intuitive feedback for humans~\citep{scheurer2022training,chen2024learning}.

\paragraph{Align through environment feedback~(\S\ref{sec:environment})} aims to automatically obtaining alignment signal or feedback from existing environment instead of a trained model, such as social interactions~\citep{liu2023training,sun-etal-2023-moraldial}, public opinion~\citep{collective_constitutional_AI}, external tools~\citep{qiao2024making,jiang2023selfevolve} and embodied environment~\citep{bousmalis2023robocat,xu2024sinvig}.
Environment feedback serves as an essential supplement for previous origins of alignment signal, which enable the AI system better adapt to real-world application scenarios. However, how to effectively utilize environment feedback is still a research direction that urgently needs further exploration.

\paragraph{Underlying mechanisms for automated alignment~(\S \ref{sec:mechanism})} Apart from reviewing the above representative technique for achieving automated alignment, we also provide an in-depth discussion about the underlying mechanisms for automated alignment. Specifically, we devote to investigate the following three critical questions about automated alignment: 
\begin{itemize}[leftmargin=1em]
    \item What is the underlying mechanism of current alignment? 
    \item Why does self-feedback work?
    \item Why is weak-to-strong feasible?
\end{itemize}
The explorations of these questions are crucial for achieving scalable automated alignment.
For each question, we summarize existing research and perspectives, raise open questions, and discuss their limitations and future directions.

\section{Aligning Through Inductive Bias}
\label{sec:inductive}
\epigraph{Self-education is, I firmly believe, the only kind of education there is.}{Isaac Asimov}

\begin{figure*}[!tp]
    \centering
    \includegraphics[width=\textwidth]{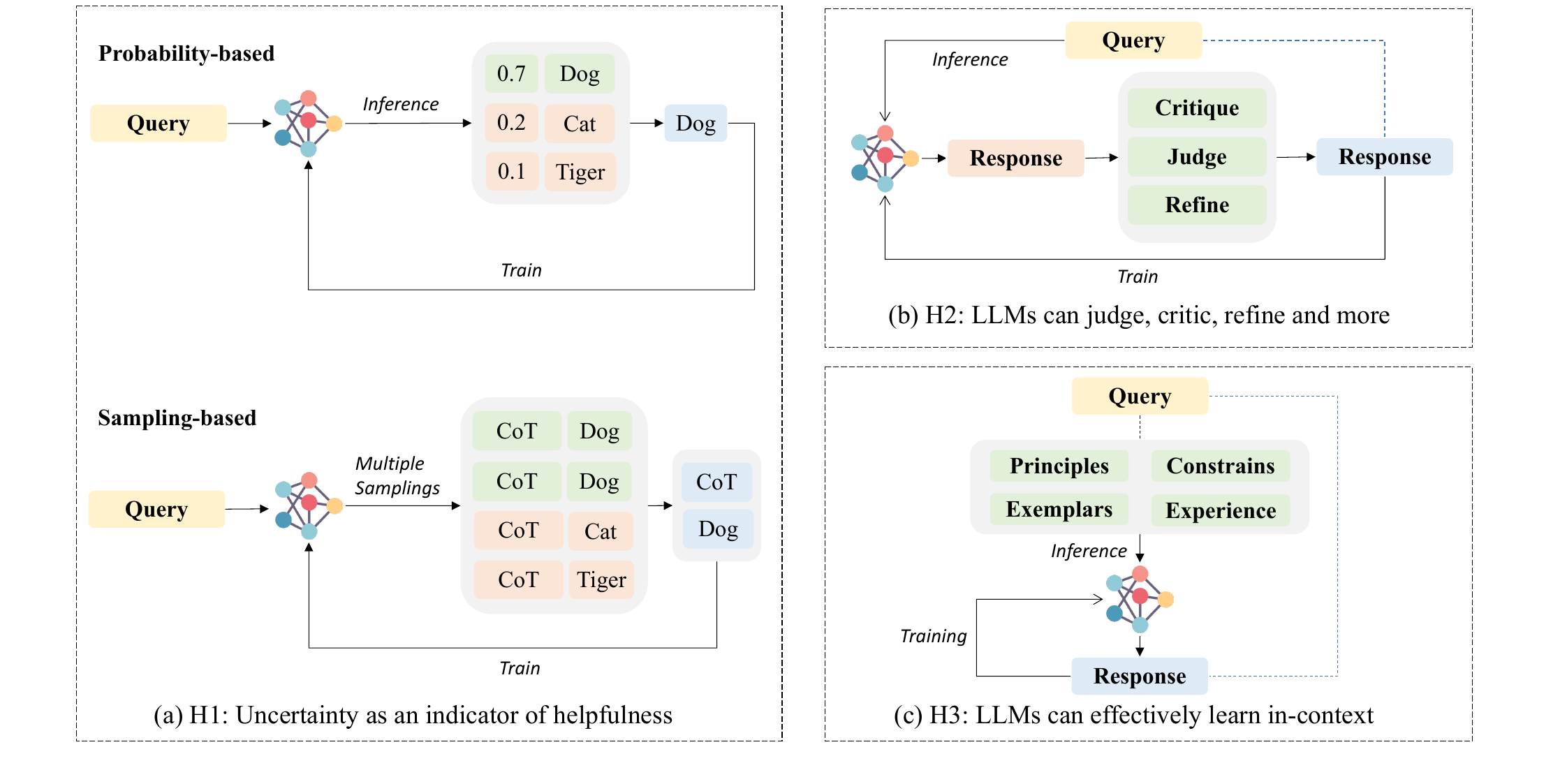}
    \caption{Illustrations of aligning through 3 kinds of representative inductive bias stemming from inherent features of LLMs.}
    \label{fig:h1}
\end{figure*}

Currently, aligning through inductive bias is one of the most promising directions for achieving automated alignment.
Inductive bias~\citep{mitchell1980need} is a set of essentially assumptions or constraints that guide model learning and decision-making processes.
By carefully selecting and implementing suitable inductive biases, we can steer models towards behaviors and decisions that are more likely to meet human standards and expectations, which can then generalize to unseen data distributions.

Compared to other methods of achieving automated alignment, aligning through inductive bias offers two primary advantages:

\begin{itemize}[leftmargin=1em]
    \item[1)] It does not require additional supervisory signals beyond the model itself, thus avoiding the high cost of obtaining additional annotated data. 
    This is particularly relevant given the current scenario where training data is becoming scarce or has already been exhausted~\footnote{Within the context of alignment discussed in this paper, we expect the model to continuously improve its helpfulness, thereby providing more effective assistance for alignment process. The scope of alignment actually represents \textit{the post training process} rather than steering the models.} \citep{xue2023repeat}.
    \item[2)] It has the potential to address the scalable oversight problem \citep{bowman2022measuring}. 
    As the potential of LLMs continues to scale, it becomes challenging for humans to provide supervisory signals that surpass their own level of knowledge. However, through inductive bias, models can continuously self-improve, transcending the limitations of human knowledge.
\end{itemize}

After conducting a thorough review of the relevant literature, we find that current efforts towards self-improvement solely through language model itself can be decomposed into a set $\mathcal{H}$ of five inductive biases. These inductive biases fall into two broad categories: 1) Those stemming from inherent features of LLMs~(\S\ref{sec:bias_features}), and 2) Those arising from the organization of LLMs~(\S\ref{sec:organization_bias}). The objective of each of these inductive biases can be summarized as a simple rule: \textbf{Heuristically transform test-time computation into the model's alignment.}~\citep{snell2024scaling} The additional computation of each inductive bias is depicted in the shaded regions of the illustrative figures.

For each type of inductive biases, we will begin by introducing its origin. Following this, we will enumerate the works that employ this inductive bias as a single-step policy improvement operator. Next, we will discuss the works that iteratively train with it, aiming for continuous improvement. Finally, we will address open research problems associated with the given inductive bias.

\subsection{Inductive Bias from Features of LLMs}

LLMs possess intrinsic features that can act as inductive biases. These features largely arise from the pre-training of deep Transformer networks on massive datasets (H1, H3), while some also stem from preliminary alignment procedures aimed at enhancing the models' helpfulness (H2). 
In this section, we will summarize three key inductive biases, as depicted in Figure \ref{fig:h1}. 
It is important to note that these inductive biases are not entirely independent; instead, they represent three different perspectives on automated alignment based on the features of LLMs.

\label{sec:bias_features}

\subsubsection{H1: Uncertainty as a indicator of helpfulness}

% \begin{figure}[!tp]
%     \centering
%     \includegraphics[width=\columnwidth]{Figures/fig_h1.pdf}
%     \caption{Illustrative figure of H1: Probabilities as a indicator of helpfulness.}
%     \label{fig:h1}
% \end{figure}

Probability distributions from models can represent uncertainty. As \citet{kadavath2022language} discovered, when prompts are suitably designed, the responses obtained from \textit{pre-trained LLMs} can be \textit{well-calibrated}, and the degree of calibration can scale with the number of parameters and the number of exemplars. 
In other words, the higher the probabilities assigned by an LLM to a given answer, the more likely that answer is to be correct.
This hypothesis has also been validated by \citet{wang-etal-2021-want-reduce} and \citet{he2023investigating}. Similarly, \citet{manakul-etal-2023-selfcheckgpt} found a correlation between the probabilities output by the aligned models and factuality.

In the machine learning literature, early applications of this inductive bias were evident in a series of works using self-training \citep{1053799} for semi-supervised learning \citep{nigam2000analyzing,amini2002semi}. 
The basic paradigm of these works involves using learners trained on labeled data to continue learning from confidently classified unlabeled data, thereby enhancing supervised learning performance with unlabeled data.
This approach has been applied in classification tasks with methods such as Pseudo-label \citep{lee2013pseudo, ferreira2023using} and Entropy Minimization \citep{grandvalet2004semi}.
 \citet{He2020Revisiting} extended this approach to sequence generation NLP tasks, highlighting that biased sampling and noise perturbation are key factors for the success of self-training in these tasks. \citet{pace2024west} extended the paradigm of self-training to the alignment problem, improving the robustness of reward models by allowing them to iteratively learn from the highest-scoring and lowest-scoring answers in the candidate pool generated for a query.

The frequency of specific answers also reflects uncertainty. 
Therefore, synthesizing candidate answers from multiple samplings can lead to better performance than relying on a single sample. 
This approach is particularly effective when LLMs are used in tasks requiring deliberate thinking (e.g., solving math problems), as a single Chain-of-Thought (CoT) \citep{wei2022chain} reasoning path can sometimes fall into local optima and generate plausible but unfaithful answers.
Self-Consistency \citep{wang2023selfconsistency}, by aggregating multiple reasoning paths through a weighted sum, can alleviate this problem by marginalizing the model's likelihood to the reasoning path. Interestingly, it was also found that an unweighted sum (i.e., majority vote) can achieve comparable performance, which is attributed to the ``similarly likely'' probabilities over all reasoning paths.
\citet{wang2024chain} further discover that the presence or absence of a CoT reasoning path correlates with the probability of the final answer when inference is conducted without any prompting techniques. 

To solidify such enhancement, Self-Improve \citep{huang2023large} views CoT with Self-Consistency as a policy improvement operator, substantially improving the reasoning capabilities and potential of LLMs through iterative learning of the reasoning paths obtained from Self-Consistency. \citet{zhang2023chainofthought} demonstrate that LMs could self-improve on large number addition problems by curriculum ``distilling'' the CoT answers to the direct answers without explicit reasoning.
Quiet-STAR \citep{zelikman2024quiet} considers the advantage of the influence of rollout rationale on the probabilities of subsequent tokens as a feedback signal, encouraging the model to generate a more helpful implicit thought process using reinforcement learning techniques.
\citet{li-qiu-2023-mot} show that memory mechanisms can also be used for consistency-based self-improvement instead of parameter learning.

\paragraph{H1: Discussion}

\begin{figure}[!tp]
    \centering
    \includegraphics[width=0.5\columnwidth]{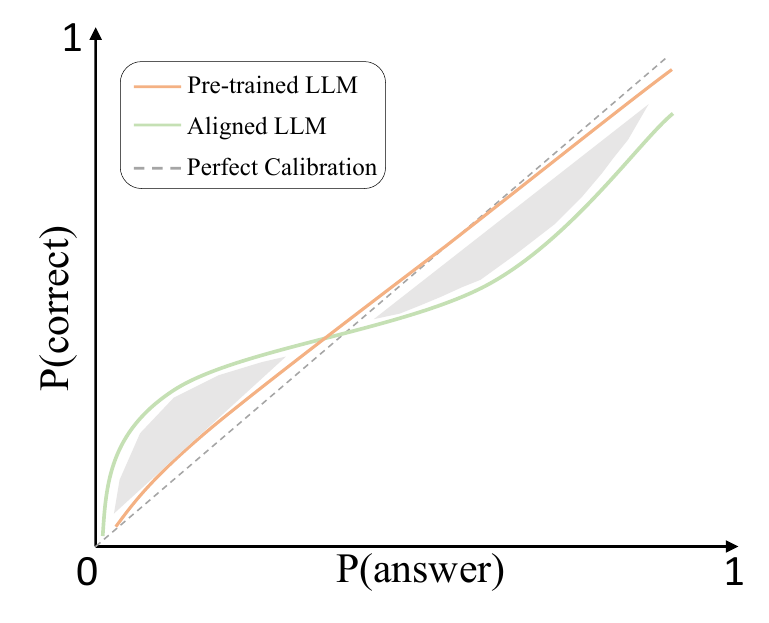}
    \caption{Calibration plot observed by \citet{kadavath2022language, he2023investigating, openaiGPT4TechnicalReport2023, zhang2024calibrating}. 
    The x-axis represents the probability associated with the model's output, while the y-axis indicates the probability that the answer is correct.
    In the low probability region, the gray area may reflect the ``don't know'' responses substituting some low-confidence answers.
    In the high probability region, the gray area signifies overconfidence. 
    The comparison between pre-trained LLM and aligned LLM demonstrates that alignment for helpfulness can result in miscalibration, which is harmful to iterative self-improvement.}
    \label{fig:h1_discussion}
\end{figure}

For aligned models, maintaining calibration and uncertainty remains crucial, as miscalibration can undermine the potential for iterative self-improvement.
Numerous studies \citep{kadavath2022language, he2023investigating, openaiGPT4TechnicalReport2023, zhang2024calibrating} have noted that the alignment process can impair the calibration of LLMs (as shown in Figure~\ref{fig:h1_discussion}). 
This observation is reasonable for several reasons:
1) The current superficial alignment process aims to steer the model away from generating harmful or incorrect answers. This involves replacing the probability of incorrect answers with the probability of rejection responses to some extent, creating a gray area in the low probability region.
2) During the alignment process, the model also learns the response format. The increase in confidence regarding the response format can somewhat affect the confidence in the answers themselves \citep{he2023investigating}. 
Moreover, the model relearns the correct answers, which can lead to overconfidence (represented by the gray area in the high probability region).
The extremization of the probability distribution can be more pronounced in self-alignment~\citep{wu2024progress}, given that it is an iterative process involving self-sampling and training. 
This implies that all tokens are already in a very high probability distribution, making them more likely to be sampled as responses.
 
When the model becomes overconfident, it leads to a decrease in the diversity and exploratory of the model's generated outputs. 
To mitigate this issue, one promising approach is to use inference-time interventions (e.g., high temperature \citep{kadavath2022language}, fidelity \citep{zhang2024calibrating}) to decrease the expected calibration error.
Another potential solution is to filter the pseudo-labeled samples to avoid harmful repeated training, which requires understanding when unlabeled samples will be effective \citep{grandvalet2004semi}.

\subsubsection{H2: LLMs can judge, critic, refine and more}
\label{sec:H2: LLMs can judge, critic, refine and more}

Pre-trained LLMs often struggle to directly respond to instructions. However, the widespread adoption of imitation learning \citep{vicuna2023} and feedback learning \citep{bai2022training} has significantly enhanced the zero-shot helpfulness of LLMs. 
Leveraging the reasoning abilities elicited and enhanced by these general helpfulness improvements, a series of works have emerged, utilizing the model's capabilities to enhance response quality and safety through judging, critiquing, refining, and more.

\textbf{\textit{Judge}} refers to determining the quality of model responses. 
The judgment standards are typically incorporated into instructions as principles or guidelines, enabling regulators to oversee LLM behavior in a more scalable manner \citep{bai2022constitutional, yuan2024self} compared to heavy reliance on human annotators for feedback \citep{bai2022training}. 
This approach allows for timely regulation, aiding in the flexible and controlled alignment process of language models, which can help prevent issues like reward hacking during iterative training \citep{sun2023salmon} and facilitate on-policy reinforcement learning training \citep{guo2024direct}.

Self-judging can manifest in two primary forms: 
1) Differentiating the relative quality of two responses (AI Feedback \citep{bai2022constitutional}), resulting in evaluation outcomes represented as a partial order. 
For instance, \citet{tan-etal-2023-self} employ prompts to compare which answer better adheres to the HHH principles.
They then distill the chosen option back into the model to further enhance its judging capability. \citet{bai2022constitutional} prompted the model to select superior responses based on sampled principles and subsequently employed a preference process to achieve a Pareto improvement of the model. 
2) Providing an absolute score for a response (LLM-as-a-judge \citep{zheng2024judging}), with the evaluation results in scalar form. \citet{yao2023tree}, \citet{besta2024graph} and \citet{xie2023selfevaluation} introduce real-time evaluation modules for thought states during reasoning process. These modules serves as a prior during the search process, assisting the model in exploring the action space for questions requiring deliberate thinking.
Similarly, RAIN \citep{li2024rain} utilizes a binary scoring prompt for self-evaluating whether the generation is likely to be harmful, thereby enhancing response safety combined with an inference-time tree search. 
\citet{yuan2024self} employ a five-point judge prompt to score the model's instruction-response outputs, then converted the scores into a partial order for iterative training using DPO.

Recalling H1, it becomes evident that H1 serves as a foundation for H2, given that the accuracy of the judge is directly tied to the calibration of the LLM. Consequently, H2 will only be effective if H1 is valid \citep{bai2022constitutional}. 

\textbf{\textit{Critique}} refers to generating modification suggestions. By leveraging LLM itself for critique, the suggestions can address errors and deficiencies, such as mistakes in summarization \citep{saunders2022self}, machine translation \citep{fernandes-etal-2023-devil}, math reasoning \citep{lin2024criticbench}, decision-making and programming tasks \citep{saunders2022self, shinn2023reflexion}. The suggestions can also pertain to abstract value criteria, such as the HHH related principles \citep{chen2024iteralign, bai2022constitutional}.

\textbf{\textit{Refine}} refers to the ability that LLMs can improve the given text. 
Most of the work on self-refine is based on natural language reasons provided by the critic module to modify the response \citep{bai2022constitutional, tan-etal-2023-self, madaan2023selfrefine, shinn2023reflexion}. 
Some research also demonstrates the possibility of making modifications directly based on scalar rewards \citep{shinn2023reflexion}.
The less informative critic can be more challenging for LLMs since they must complete more information by themselves through reasoning. Another line of work uses LLMs to refine the prompts themselves \citep{fernando2023promptbreeder, yang2023large}.

\textbf{\textit{Other:}} A helpful LLM can serve in various capacities to assist in the alignment process, effectively replacing human guidance. For example, it can vote on the quality of intermediate states of thought \citep{yao2023tree}, verify outcomes based on predicting conditions in questions \citep{weng-etal-2023-large}, and automatically generate, filter \citep{yue2024mammoth2}, and evolve instructions \citep{wang-etal-2023-self-instruct, li2024selfalignment, xu2023wizardlm}, among other tasks.

Regarding the persistence process, the improvements derived from these methods can be further distilled into the model through SFT (i.e., Expert Iteration), DPO, RM-PPO, and other techniques. Additionally, the judge / critique - refine process can be conducted iteratively.

\paragraph{H2: Discussion}

As the abilities of judging, critiquing, and refining are increasingly incorporated into model feedback and learning processes, there is a need for systematic evaluation of these capabilities. In this context, several research directions are worth pursuing:
\begin{itemize}[leftmargin=1em]
    \item[1)] Benchmarking the performance of existing models on these atomic abilities, as exemplified by works such as \citet{sun2024critique} and \citet{lin2024criticbench}.
    \item[2)] Conducting causal studies on the formation process of models' judging, critiquing, and refining abilities, and investigating what forms of pre-training and fine-tuning data can influence these abilities. This helps to design specific pipelines to targeted enhance these capabilities~\citep{wang2024self}, and even train specialized models like CriticGPT~\citep{mcaleese2024llm}.
    \item[3)] Evaluating the effects of distribution shift on these abilities. Do models still possess reliable evaluation and improvement capabilities if they have not been trained on the corresponding instruction and response pairs? This is particularly pertinent to the scalable oversight problem, which assumes the absence of direct supervision for instructions.
    \item[4)] Gathering empirical evidence to demonstrate that self self-critique, judge, and refine abilities can enhance model performance in fair and reasonable experimental settings. Some works point out that the improvement may come from the use of stronger models \citep{sharma2024critical} and golden labels \citep{huang2023large}.
\end{itemize}

\subsubsection{H3: LLMs can effectively learn in-context}

\begin{figure*}[!tp]
    \centering
    \includegraphics[width=\textwidth]{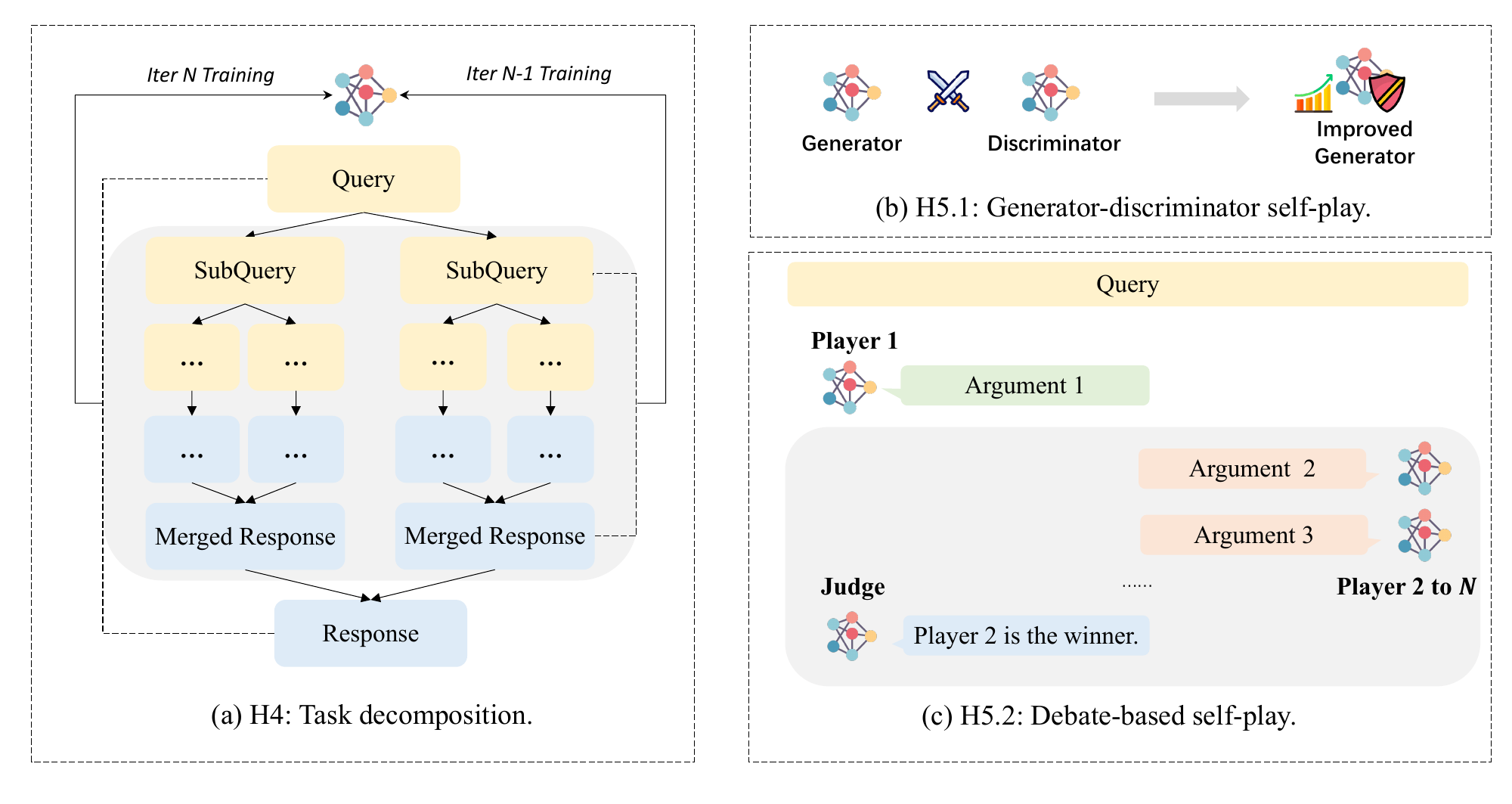}
    \caption{Illustrations of aligning through three representative inductive biases which stem from the organization of LLMs.}
    \label{fig:h45}
\end{figure*}

In-context learning (ICL) refers to the ability of LLMs to initialize a task-specific model with exemplars or experiences during inference \citep{brown2020language}. 
Given that certain studies \citep{dai-etal-2023-gpt, von2023uncovering} suggest parallels between ICL and parameter gradient descent, it is plausible to regard it as a versatile and effective ``learning'' method.

From the perspective of automated alignment, ICL offers an efficient means to cold-start from a pre-trained LLM. 
With the assistance of ICL, just a few conversational samples in-context can yield a somewhat aligned model \citep{ganguli2022red, sun2023principle, lin2024the}. 
Similarly, by prepending a few annotated exemplars in-context, ICL can also elicit judge and critic abilities of pre-trained LLM to some extent, or enhance the performance compared to the zero-shot setting \citep{bai2022constitutional}. 
Additionally, ICL presents a potential method for adaptive alignment \citep{xu2023align} to different social norms and regulations.

However, prepending few-shot exemplars in the context above can make inference inefficient \citep{gim2023prompt} and interfered with the unrelated queries \citep{shi2023large}. 
Therefore, self-generated labels obtained from ICL could be directly used as pseudo labels, and distilled back into the LLMs only paired with the query. 
This paradigm is known as Context Distillation \citep{askell2021general, snell2022learning}. 
For instance, in the alignment process of Llama-2 \citep{touvron2023llama}, Context Distillation is used to alleviate the problem of long-term dependencies of system prompts.
For Llama-3~\citep{dubey2024llama}, context distillation is utilized to steer the model towards generating more readable and well-documented code. In Dromedary \citep{sun2023principle}, the base language model is transformed into a safe and helpful aligned model with minimal annotations by directly training on samples acquired from multiple ICL processes. 
\citet{padmanabhan2023propagating} demonstrate that Context Distillation can also be used to inject new knowledge to models by learning continuations from entity definitions. 
Furthermore, \citet{yang2024rlcd} illustrate the effectiveness of distilling the preference pair generated by contrastive in-context constraints back into the model.

Additionally, the learning content of ICL can also encompass exploratory experiences \citep{shinn2023reflexion} and tool definitions \citep{yao2022react, tang2023toolalpaca}. 
In other words, agents equipped with tools and experiences can potentially outperform those without. This suggests a similar potential in distilling back the trajectories improved by experience and tools to continuously enhance the same model.

\paragraph{H3: Discussion} Unfortunately, the black box nature of ICL itself poses a significant challenge to alignment \citep{anwar2024foundational}. 
Without a comprehensive understanding of how LLMs learn in-context, the context distillation approach may introduce problems by potentially amplifying biases and errors inherent in the ICL process of the models.
Moreover, the ability of long in-context learning \citep{agarwal2024many, li2024long} warrants further exploration, as it facilitates more efficient distillation and is crucial for scalable oversight settings where models need to comprehend lengthy professional documents or extensive self-play histories.

\subsection{Inductive Bias from Organization of LLMs}

\label{sec:organization_bias}

In addition to the inductive bias originating from the shared features of LLMs, another set of biases arises from the composition or organization of multiple LLMs, as depicted in Figure \ref{fig:h45}. 
Based on whether the relationships between the constituent LLMs are cooperative or adversarial, two representative inductive biases emerge: ``Task Decomposition'' and ``Self-play''. 
It is noteworthy that, as this field progresses, we anticipate subsequent literature will adopt more complex organizational and learning structures. 
Both adversarial and collaborative modalities may form integral components of sophisticated agent systems. However, at the current stage, task decomposition and self-play serve as practical taxonomies. Subsequent sections will delve into these concepts in detail.

\subsubsection{H4: Task decomposition}

Task decomposition has long been regarded as an effective approach to tackling complex problems~\citep{lee2001does}. 
For example, in cooperative games grounded in collective rationality, the overall benefits accrued by an alliance exceed the sum of individual gains \citep{shapley1971cores}.
Moreover, the divide-and-conquer paradigm and recursion are well-established and effective means employed in algorithm design for addressing problems of substantial scale and complexity \citep{hoare1961algorithm, wilf2002algorithms}.

The discussion of this paradigm can be traced back to the assumption of factored cognition \citep{factored_cognition}. 
It advocates that cognition tasks can be recursively decomposed. 
If an AI or human encounters a task that is difficult to solve, it can decompose the task, assign the decomposed problems to a series of its own copies for parallel processing, and finally merge these results.
The copies focus on short-term work and work independently. 
A series of prompting methods implicitly or partially adopts the factored-cognition assumption for automated alignment. 
For instance, \citet{zhou2023leasttomost} and \citet{wang-etal-2023-plan} prompt the LLM to decompose the problem, then guide it to sequentially solve the sub-problems. 
It is also believed that task decomposition is an effective method for solving Easy-to-Hard generalization \citep{zhou2023leasttomost}, that is, constructing decomposition prompts on simple samples and filling them in-context allows LLM the potential to generalize to difficult samples. \citet{khot2023decomposed} further implement recursive task decomposition.

Based on the assumption of factored cognition, Iterative Distillation and Amplification (IDA) \citep{christiano2018supervising} views each decomposition-merging process as a form of \textit{amplification} and considers learning from the final merged results as a form of \textit{distillation}.
Although the original IDA paper builds this theoretical framework in a human-in-the-loop manner, where humans supervise the initial task decomposition step, it is likely that this process can be initiated without much human oversight given H1, H2, and H3 \citep{zhang2023chainofthought}.

Notably, IDA represents a promising avenue towards achieving scalable oversight, making it possible to address long-horizon tasks that are difficult for humans to directly supervise, by decomposing tasks into more tractable sub-problems.
For example, labels for data points like ``peer-review this survey'' can take several months to collect in real world.
Such problems can be tackled more quickly through factored cognition. 
Although some work partially demonstrates the effectiveness of IDA on real-world tasks like book-length summarization \citep{wu2021recursively} and complex code bug fixing \citep{wen2024learning}, this ideology still relies on a set of crucial assumptions:
1) It is still unclear whether decompose a problem is the hardest part of solving it, if the cognitive burden cannot be distributed, IDA may struggle to take effect. 2) The error won't accumulate.
Although this paradigm does not require the collaboration between agents to be efficient \citep{christiano2018supervising}, too many errors can still be problematic.
3) The extent to which tasks can be parallelized. If the task-solving process is largely sequential, the time to collect signals might increase, but this appears to be a minor issue given the current deployment speed of LLMs. 
Overall, since these assumptions are difficult to prove or falsify, we advocate for more empirical research in this direction.

\subsubsection{H5: Self-play}

Complexity emerges from adversariality \citep{bansal2017emergent}. 
Self-play refers to a paradigm where an agent learns by iteratively playing games \textit{against} itself, a form of non-cooperative game \citep{nash1951non} in which each agent aims to maximize its own utility. 
It serves as the foundation of many successful specialized superhuman AI systems like AlphaGo Zero \citep{silver2018general} and StockFish \citep{stock_fish}.
Given this success, self-play seems to be a potential approach for enabling general proposed superhuman intelligence from LLMs. 
Two representative self-play methods are the Generator-Discriminator and the Debate approaches, with the latter involving $N \ge 2$ adversarial generators and one discriminator in a gaming environment.

\paragraph{H5.1: Generator-Discriminator} 

In the Generator-Discriminator self-play framework, the role of the discriminator is to evaluate the outputs produced by the generator, determining whether these outputs are of high or low quality.

As discussed in H2, the judge and critic model is commonly considered a type of discriminator. 
For example, \citet{yuan2024self} utilized rewards from LLM-as-a-Judge to identify high-quality and low-quality responses from the generator, optimizing the generator towards the higher quality ones.
However, the adversarial setting between the discriminator and the generator is limited because the only assumption is that the discriminator's ability can be improved with general helpful training.
The discriminator remains almost static (the prompts are unchanged) during training, making it possible for the generator to be over-optimized against the discriminator, leading to reward hacking.
Effectively improving the judge and critic module alongside the generator is thus a crucial problem. 
One reasonable approach is to formulate the training process as an adversarial game \citep{cheng2023adversarial}, where the policy and reward model are updated alternatively via a min-max loss.
Another way to introduce a more adversarial setting is to optimize the gaming problem at inference time, as demonstrated in the Consensus Game by \citet{jacob2024the}. 
It employs the \texttt{piKL} no-regret learning algorithm to iteratively update the strategies of both the generator and discriminator, converging to a Nash equilibrium. This equilibrium strategy is then used to rank candidate responses, prioritizing those agreed upon by both players.

As Generative Adversarial Networks (GANs) \citep{goodfellow2014generative} have become a well-established class of methods in traditional NLP \citep{zhang2016generating, wu2021textgail}, another line of work involves using a GAN-like discriminator to distinguish between the model's current predicted distribution and the golden distribution. 
For instance, \citet{chen2024self} find that a specific type of iterative DPO training, which consistently treats the policy-generated responses as negative and the golden responses as positive, can be viewed as a self-play process.
In this process, the implicit reward function of DPO serves as a discriminator between the model's predictions and the golden samples. Expanding on this, \citet{shaikh2024show} further add replay comparison signals between earlier iteration of models and the golden, and comparisons between a model and its successive model in the self-play process.
However, for open-ended questions, the golden distribution is sometimes still suboptimal, and this setting precludes the possibility of generating responses that are better than the golden ones.

\paragraph{H5.2: Debate} 

The debate paradigm \citep{irving2018ai} is largely inspired by factored cognition and AlphaGo \citep{silver2018general}. In the learning algorithm of AlphaGo, three distinct components are integrated: a player, a counterpart (itself), and a value model that evaluates the win rate associated with each board state. By employing Monte Carlo Tree Search (MCTS), the algorithm conducts rollouts, which are simulated self-play trajectories that extend until a game's conclusion. These rollouts enhance the accuracy of the value estimates through backward updates based on the outcomes, concurrently refining the policy by capitalizing on strategies that have previously led to victories.

The game of Go shares similarities with tackling the scalable oversight problem using natural language debate. In the beginning or middle of a Go game, even experienced experts may find it difficult to judge which side has a higher probability of winning, just as humans have a small probability of correctly judging problems that exceed human knowledge levels. However, as the game nears its end, the outcome usually becomes clear that even a non-expert judge can confidently evaluate the board generated by the Go masters. For a debate competition, the winner can typically be summarized by the judges. This analogy illustrates a crucial point: through the skillful introduction of adversarial processes, the burden of oversight for complex problems can be significantly reduced.

This provide a possible oversight solution to build trustworthy superhuman AI systems. \citet{irving2018ai} show that honesty is better strategy than lie in debate paradigm through proof-of-concept experiments. As an extension of this, \citet{brown2023scalable} propose a new set of debate protocols, wherein the honest strategy can \textit{always succeed} through a simulation involving only a polynomial number of steps. \citet{khan2024debating} conduct a thorough empirical study on the feasibility of implementing the debate paradigm on LLMs: it was found that the debate paradigm can significantly enhance truthfulness, and more persuasive \citep{anthropic2024measuring} debaters lead to more truthful outcomes. Furthermore, \citet{kirchner2024prover} demonstrate that a prover-verifier game employing a powerful honest prover, a potent malicious prover, and a comparatively weaker verifier can generate more readable outputs, thus enabling more effective human oversight of the stronger model.

Apart from the classic natural language debate, a growing body of research has explored the implementation of the debate paradigm across diverse game scenarios to enhance specific model features. 
A representative arena is the \textit{bargaining task} \citep{nash1950bargaining}. 
\citet{fu2023improving} focus on zero-sum variants of bargaining, where the balloon seller aims to sell at a higher price while the buyer seeks a lower price. 
They observed significant variations in bargaining capabilities among different LLMs and their capacity to learn from play experiences and feedback.
\citet{cheng2024self} implement the adversarial language game \textit{adversarial taboo} \citep{yao2021adversarial}, where an attacker and a defender engage in a conversation centered around a target word visible only to the attacker. 
The attacker subtly induces the defender to unconsciously utter the target word, while the defender tries to avoid doing so and guess the word from the context. 
Both players acquire basic gaming skills through imitation learning from a teacher LLM and then refine their strategies through self-play. 
Interestingly, less capable player LLMs not only improve their win rates in this specific game but also enhance their general reasoning abilities.
\citet{ma2023red} introduce the \textit{red-teaming game}, a more intricate adversarial team game where LLMs are initialized as a joint set of red-teaming policies to prompt the target LLM to produce harmful content.
They propose a solver to ensure the final meta-strategy approximates a Nash equilibrium within a certain $\epsilon$ margin. \citet{zheng2024toward} suggests addressing the alignment problem by allowing an attacker to prompt a defender LLM to generate answers that might result in low rewards, while the defender tries to maximize the rewards of these prompts. The solution of this game is considered an iterative min-max optimization process with constraints. 

\subsubsection{Discussion}

\label{sec:self_play_discussion}

Task decomposition and self-play both necessitate LLMs functioning as agents. Indeed, although research on agents is already very prosperous, the current capabilities of LLMs as agents are still very limited. They still struggle to complete tasks that would take humans several hours to finish~\citep{gpt4o}. Therefore, focusing on improving the capabilities of LLMs-as-agents remains an important future research direction. This includes understanding and modeling the human objectives in the most economically valuable tasks, and developing effective reasoning abilities for ultra-long thought process.

Meanwhile, the challenge of aligning LLMs as agents is more complex compared to aligning them as chatbots, as it requires considerations of behavior-level alignment \citep{pan2023rewards, yuan2024r}, the dynamics of environment and self-constraints~\citep{garrabrant2018embedded, shavit2023practices, yang2024towards}. 
More importantly, while adversarial self-play may enhance agent capabilities in long-horizon tasks, it can also give rise to the emergence of more deceptive~\citep{hubinger2024sleeper}, persuasive and autonomous agents~\citep{tao2024survey}. 
Such developments could have significant social impacts and ethical risks, such as the potential for models to generate more persuasive articles than humans, which could be exploited for political manipulation.
Encouragingly, several prominent model engine providers have taken steps to monitor and mitigate these potential side effects. 
For example, OpenAI's Preparedness team has established benchmarks for assessing persuasion and autonomy \citep{openai2023preparedness}, categorizing model risks into four levels and imposing development and deployment restrictions based on risk thresholds. Additionally, third-party organizations are contributing to the development of robust safety frameworks for highly capable agents \citep{metr2024update}.

Furthermore, an even more intricate problem lies in proving the theoretical safety and trustworthiness of multi-agent systems.
Although research in this area is nascent \citep{yang2020overview, digiovanni2021survey}, advancements in game theory \citep{hazra2022applications}, automated theorem proving techniques \citep{polu2020generative} and real-world simulation technology \citep{videoworldsimulators2024} may offer insights into addressing this challenge.

\section{Aligning Through Behavior Imitation}
\label{sec:imitation}
\epigraph{Imitation is the first instinct of the awakening mind.}{Maria Montessori}

\begin{figure}[!tp]
\centering
    \includegraphics[width=0.7\columnwidth]{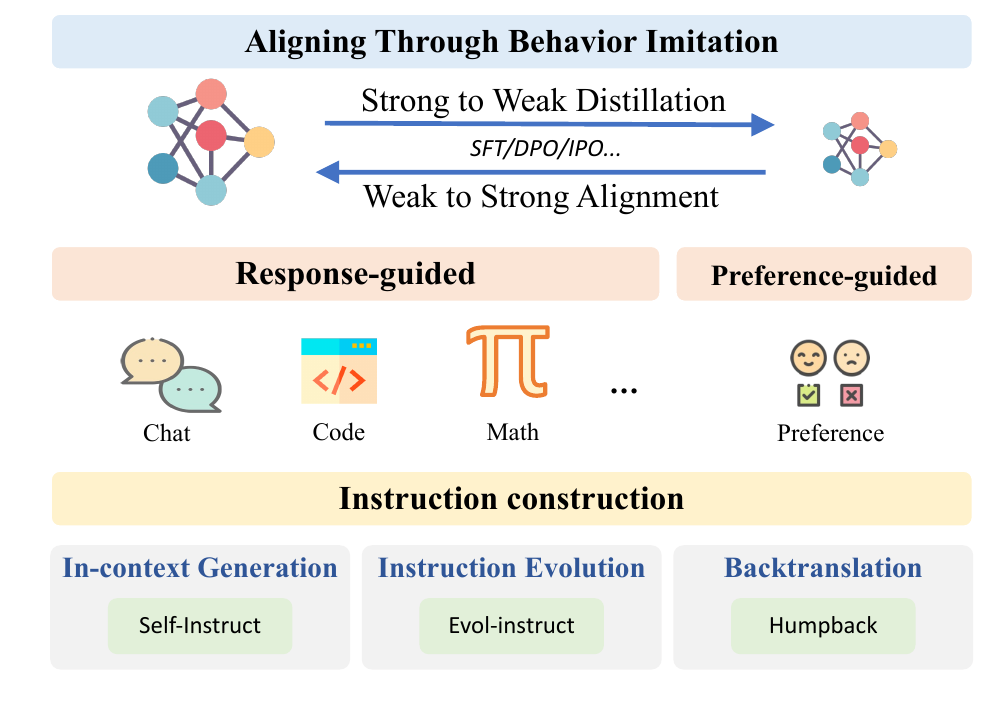}
    \caption{The illustrations of representative studies for aligning through behavior imitation.}
    \label{fig:imitation}
\end{figure}

Aligning through behavior imitation is another widely-used strategy for automated alignment, which aligns the target model by mimicking the behavior of another aligned model.
Specifically, 
as demonstrated in Figure~\ref{fig:imitation}, this method begins by collecting high-quality instructions as task descriptions~\citep{wang-etal-2023-self-instruct}. 
A supervised model is then employed to generate alignment signals, which typically include instruction-response pairs~\citep{alpaca}, pair-wise preference data~\citep{cui2024ultrafeedback}, and other alignment signals~\citep{franken2024selfsupervised}. 
Ultimately, the target model is aligned by imitating these produced behaviors.

Based on the capability comparison between the supervised model and the target model, studies on aligning through behavior imitation can be categorized into strong-to-weak distillation~(\S\ref{Strong-to-Weak Distillation}) and weak-to-strong alignment~(\S\ref{Weak-to-Strong Alignment}). For each category, we thoroughly review the representative studies, summarize current progress and limitations, and discuss future directions.

\subsection{Instruction Construction}
\label{Instruction Construction}
Collecting large-scale instructions with high quality and diversity serves as the foundation for achieving alignment through behavior imitation.
The most intuitive strategy involves filtering out high-quality data from human-written instructions.
However, this approach requires substantial human effort and expertise, which also introduces significant noise.
Consequently, many studies focus on utilizing LLMs for automatic instruction generation, thereby significantly reducing the dependence on human annotation. 
Based on the information provided for instruction construction, there are currently 3 representative strategies: 

\noindent \textbf{In-Context Generation,} which provides in-context demonstrations to guide LLMs generating instructions.
For example, ~\citet{honovich-etal-2023-unnatural,wang-etal-2023-self-instruct,alpaca} begin with a small set of human-written instructions. 
These instructions are randomly selected to create context examples that prompt LLMs to generate additional instructions. 
% Similarly, Stanford Alpaca~\citep{alpaca} employs a comparable approach by using ChatGPT to gather instruction-response pairs.
To further improve the scale and diversity of generated instructions, LaMini-LM~\citep{wu-etal-2024-lamini} additionally introduces wiki data for topic-guided instruction generation, thereby constructing a large, offline distilled instruction dataset.
% \citet{guo2024humaninstructionfree} use high-quality samples relevant to the target domain as examples for ICL to generate additional samples.
Dynosaur~\citep{yin-etal-2023-dynosaur} leverages meta-information from existing NLP datasets to create a dynamically growing instruction tuning dataset. 
Moreover, LLM2LLM~\citep{lee2024llm2llm} enhances the difficulty and complexity of instructions by iteratively introducing examples that the model fails to answer correctly.

% Furthermore, LLM2LLM~\citep{lee2024llm2llm} enhances the dataset by generating new data from the teacher model using incorrect predictions from the student model, which improves the dataset's ability to handle data-constrained tasks.

\noindent \textbf{Instruction Evolution,} which involves rewriting existing instructions based on pre-defined evolution principles.
Evol-Instruct~\citep{xu2024wizardlm} employ LLMs to conduct instruction evolution based on handwritten principles, thereby reducing the need for manual annotation and enhancing the model's ability to manage complex tasks.
Building upon this, TeaMs-RL~\citep{gu2024teaching} trains another model through reinforcement learning to generate optimized evolution trajectories.
% apply an instruction evolution paradigm, where existing instruction data is transformed using randomly selected handwritten evolution principles. 
% This approach is employed for robust models such as GPT-3, facilitating the rewriting of instructions, 
% thereby reducing the need for manual annotation and enhancing the model's ability to manage complex tasks.
% Building upon the Evol-Instruct framework, TeaMs-RL~\citep{gu2024teaching} advances this methodology by training another model through reinforcement learning to generate optimized instruction trajectories. 
Considering the reliance on manually written principles, Auto Evol-Instruct~\citep{zeng2024automatic} proposes an automated principle construction method, further enhancing the diversity and complexity of evolved instructions.
% Recently, Auto Evol-Instruct~\citep{zeng2024automatic} automatically discovers and improves evolution strategies based on issues exposed during the instruction evolution, outperforming human-designed methods.

\noindent \textbf{Instruction Backtranslation,} which employs LLMs to predict instructions based on responses extracted from human handwritten text or web documents. 
LongForm~\citep{longform}, TEGIT~\citep{chen2023tegit} and Humpback~\citep{li2024selfalignment} prompt LLM to construct instructions according to cleaned web corpus.
% label text from web documents, thereby creating high-quality instructional data. 
% TEGIT~\citep{chen2023tegit} engages ChatGPT to simultaneously generating instructions, inputs, and outputs to effectively filter out noise.
REInstruct~\citep{chen-et-al-2024-reinstruct} builds instructions from an unlabelled corpus and rewrites the unlabelled text to enhance its quality as a response.

\subsection{Strong-to-Weak Distillation}
\label{Strong-to-Weak Distillation}
Based on the collected instructions, strong-to-weak distillation seeks to align the weaker target model by imitating the responses or preference data generated by another stronger and well-aligned model.
In the following subsections, we will introduce representative studies concerning response-guided and preference-guided distillation, respectively.

\subsubsection{Response-Guided Distillation}
\label{Response-Guided Distillation}
In response-guided distillation, the target model emulates the teacher model by directly learning the responses to different instructions through instruction tuning. 
This approach has inspired numerous studies that aim to distill various capabilities from the teacher model to the target model. 
These capabilities include not only general instruction-following skills but also domain-specific abilities such as mathematics, coding, and agent-related tasks.

\paragraph{Instruction-Following} 
After constructing instruction data, corresponding responses can easily be developed from the teacher model. 
Training with these instruction-response pairs emulates the teacher's capability of following instructions. 
For instance, LLaMA-GPT4~\citep{peng2023instruction} utilizes GPT-4 to generate responses to instructions derived from Alpaca~\citep{alpaca}. 
% \citet{ubani2023zeroshotdataaug} discuss the zero-shot prompting of ChatGPT for data augmentation in low-resource scenarios, synthesizing instructional data by designing high-quality and diverse prompts. 
In addition to single-round data, some studies focus on collecting multi-turn trajectories from teacher models. Baize~\citep{xu-etal-2023-baize} and Ultrachat~\citep{ding-etal-2023-enhancing} using two ChatGPT APIs to play the roles of user and assistant to generate multi-round conversations. 
Parrot~\citep{sun2023parrot} trains models to simulate humans in generating instructions and uses these trained models to engage in multi-turn conversations with ChatGPT on various topics.

\paragraph{Mathematics}
Wizardmath~\citep{luo2023wizardmath} employs the Evol-Instruct method to construct a comprehensive dataset specifically for mathematical reasoning tasks.    
MetaMath~\citep{yu2024metamath} utilizes ChatGPT to bootstrap mathematical questions by rephrasing them from multiple perspectives without introducing additional knowledge.  
MAmmoTH~\citep{yue2024mammoth} produces a dataset comprising math problems and model-generated solutions distinguished by a unique combination of chain-of-thought (CoT) and program-of-thought (PoT) rationales. MathCoder~\citep{wang2024mathcoder} generates innovative and high-quality math problems along with their code-based solutions using the GPT-4 Code Interpreter.
MathGenie~\citep{lu2024mathgenie} generates diverse and reliable math problems through a process of question back-translation. 
MARIO~\citep{liao2024mario} leverages GSM8K and MATH as seed data, resulting in 26.9K solutions annotated by GPT and human experts. 
Beyond purely mathematical data, several other studies propose transferring essential reasoning abilities from commercial LLMs to small models by generating detailed CoT responses~\citep{shridhar-etal-2023-distilling, fu2023specializing, hsieh-etal-2023-distilling, magister-etal-2023-teaching, ho-etal-2023-large, li2022explanations, li2023query, zhou2024jiuzhang30, hong2024abstractionofthought}.
    
\paragraph{Coding} 
State-of-the-art LLMs, such as GPT-4, demonstrate exceptional performance in coding tasks. 
Apart from pre-training on raw code data, some approaches aim to transfer coding capabilities from teacher models through instruction tuning. 
Code Alpaca~\citep{codealpaca} and WizardCoder~\citep{luo2024wizardcoder} adhere to general automatic instruction-building paradigms. 
Code Alpaca employs Self-Instruct on 20K instruction-following data, thereby extending Alpaca's capabilities to the coding domain. 
WizardCoder adapts the Evol-Instruct method for the coding domain, generating complex code and program instructions from simple coding and programming directives. 
WaveCoder~\citep{yu2024wavecoder} and Magicoder~\citep{wei2023magicoder} create high-quality instruction data utilizing open-source code datasets. WaveCoder enhances LLMs with open-source code snippets to produce superior instruction data for coding tasks.
Magicoder creates multi-task data generated according to the techniques of the Self-Instruct. OpenCodeInterpreter~\citep{zheng2024opencodeinterpreter} utilizes GPT-3.5 and GPT-4 to improve solutions with integrated text explanations and code snippets, incorporating execution and feedback for dynamic code refinement.
    
\paragraph{Agent} 
Although open-source LLMs have achieved comparable performance to commercial models in many aspects, their capabilities in agent-related functions, such as tool usage and complex task planning, remain significantly limited.
To address this issue, ToolLLM~\citep{qin2023toolllm} has created an instruction-tuning dataset named ToolBench with ChatGPT, acquiring general tool usage capabilities in a zero-shot manner. 
Similar works include Graph-ToolFormer~\citep{zhang2023graphtoolformer}, Gorilla~\citep{patil2023gorilla}, GPT4Tools~\citep{NEURIPS2023_e3936777}, ToolAlpaca~\citep{tang2023toolalpaca}, and others. 
Beyond tool usage, some studies focus on planning tasks. 
Examples include FIREACT~\citep{chen2023fireact}, AgentTuning~\citep{zeng2023agenttuning}, ReAct Meets ActRe~\citep{aksitov2023rest}, ReST meets ReAct~\citep{yang2024react}, and ETO~\citep{song2024trial}.

\subsubsection{Preference-Guided Distillation}
\label{Preference-Guided Distillation}
Although response-guided distillation can enhance the performance of student models~\citep{DBLP:conf/emnlp/WangMAKMNADASPK22}, it does not effectively help the student model align with human preferences~\citep{xu2024survey}. 
Therefore, some works concentrate on preference-guided distillation, which aligns the student model with the preferences reflected in the output from the teacher model. 
In this paradigm, the teacher model is guided to generate preference data in the form of partial order pairs, which are then used to align the student model via direct preference optimization algorithms such as DPO~\citep{rafailovDirectPreferenceOptimization2023a}, IPO~\citep{azar2024general}, and PRO~\citep{songPreferenceRankingOptimization2024}.
Based on the methodologies for constructing partial order signals, current works primarily encompass three paradigms: 1) Score-based, which involves scoring and ranking responses; 2) Refine-based, which involves refining existing responses with AI feedback; and 3) Source-based, which focuses on learning the human preference of different data sources.

\paragraph{Score-based} Through the implementation of meticulously designed diverse instructions and model responses, along with detailed numerical and text feedback provided by GPT-4, UltraFeedback~\citep{cui2024ultrafeedback} generates a large-scale, high-quality preference dataset with fine-grained annotations. 
Additionally, Zephyr~\citep{tunstall2023zephyr} employs distilled direct preference optimization on UltraFeedback to develop small yet efficient LLMs. 
CodeUltraFeedback~\citep{weyssow2024codeultrafeedback} leverages the LLM-as-a-Judge approach of GPT, evaluating responses from a pool of 14 different LLMs and aligning them according to five coding preferences.

\paragraph{Refine-based} Other studies improve initial responses using powerful models. 
Aligner~\citep{ji2024aligner} and MetaAligner~\citep{yang2024metaaligner} utilize models such as GPT-4 to revise original responses and construct preference data. IterAlign~\citep{chen2024iteralign} automatically discovers new constitutions using an LLM and optimizes responses generated from a red team dataset to create preference data. 
Safer-Instruct~\citep{shi2024saferinstruct} employs reversed instruction tuning, instruction induction, and expert model evaluation, using both raw text and GPT-4 generated responses to build high-quality preference data. UltraInteract~\citep{yuan2024advancing} build a preference tree for each instruction where trajectories are root-to-leaf paths and paired correct and incorrect nodes or trajectories can be used for alignment.
    
\paragraph{Source-based} Learning preferences from a single model may lack diversity and amplify bias. Therefore, some works build partial order signals from different data sources. 
AlMoST~\citep{kim-etal-2023-aligning}, CycleAlign~\citep{hong2023cyclealign}, and Openchat~\citep{wang2024openchat} focus on learning comparative preferences from different data sources. \citet{kim-etal-2023-aligning} transform human preferences into a series of empirical prior rules, using LLMs of various sizes to generate preference data. 
\citet{wang2024openchat} treat different data sources as coarse-grained reward labels, generating mixed-quality data through GPT-3 and ShareGPT. \citet{hong2023cyclealign} rank responses by comparing the agreement rank of white-box and black-box models across a series of responses and constructed preference data through this ranking as context.

\subsection{Weak-to-Strong Alignment}
\label{Weak-to-Strong Alignment}
As we mentioned in Section~\ref{sec:intro}, the challenges of scalable oversight become a significant barrier for the continuous development of AI systems.
Specifically, the difficulty lies in effectively providing supervision as the capabilities of AI systems gradually surpass those of humans. 
Given the impracticality of strong-to-weak distillation approaches, \emph{weak-to-strong alignment} has emerged as one of the most promising directions for achieving automated scalable oversight~\citep{burns2023weaktostrong}.
Previous studies have mainly focused on weak-to-strong generalization between humans and AI, such as the Iterated Amplification method~\citep{christiano2018supervising}, which supervises strong learners by iteratively amplifying weak experts. 
Recent research has begun to explore using weaker models to guide stronger models to achieve superalignment~\citep{burns2023weaktostrong, liu2024cosupervised}. 
Based on the source of the alignment signals, these works can be categorized into two types: 1) using smaller but aligned models to generate signals, and 2) using weaker models to guide stronger models in generating signals. 
Moreover, some studies investigate whether models can learn from behaviors in easy tasks to improve their performance in more challenging tasks, which, although not classic behavior imitation, is still noteworthy~\citep{hase2024unreasonable, sun2024easytohard}.
In the following subsections, we will introduce the representative studies in each category respectively.

\citet{burns2023weaktostrong} employ a weaker LLM as the teacher to train a stronger LLM using a weak-to-strong approach. 
They fine-tune a larger pre-trained model based on labels generated by the smaller but aligned model and observe that the larger target model consistently outperforms the smaller supervisory model. 
Instead of relying on a single teacher, \citet{liu2024cosupervised} aim to further enhance the alignment of strong models by co-supervising a powerful student with a diverse group of professional teachers.  
\citet{somerstep2024statistical} examine weak-to-strong generalization as a transfer learning problem, achieving this through a label refinement procedure.
\citet{yang2024superficialalignment} study the multi-objective alignment in weak-to-strong generalization and discover that strong students may deceive weak teachers to gain high rewards in other dimensions, which can be mitigated by using an intermediate model. 
Additionally, Aligner~\citep{ji2024aligner} and MetaAligner~\citep{yang2024metaaligner} create partial order data by using a significantly smaller but aligned model to optimize the responses from stronger models.

In addition to directly generating signals from weak models, another possible method to achieve weak-to-strong alignment is using weak models to guide strong models in generating signals. \citet{li2024superfiltering} find the ability of both weak and strong LLMs to perceive instruction difficulty and select data is highly consistent.
Thus, smaller and weaker models can be utilized to select data for fine-tuning larger and stronger models.
Similarly, SAMI~\citep{franken2024selfsupervised} employs a weak model to write constitutions for aligning a strong baseline model.

The aforementioned works achieve weak-to-strong alignment to a certain extent and investigate potential directions for achieving superalignment.
However, weaker models may not serve as effective instructors for more complex tasks. 
Consequently, some studies attempt to align models using signals derived from simpler tasks, which are easier to generate and learn, in order to enhance performance on more difficult tasks.
For instance, \citet{hase2024unreasonable} observe that current language models generally extrapolate well from simple to complex data and can even compete with models trained directly on intricate data. 
\citet{sun2024easytohard} use reward models trained on simple tasks to assess and guide policy models on more challenging tasks, thus achieving task generalization.

\subsection{Discussion}
Current works leverage the responses or preferences from teacher models to facilitate effective generalization and scalability across various tasks, thereby significantly diminishing the necessity for manual annotation.
However, approaches exhibit notable limitations, including issues related to data quality, bias inherent in the teacher models, and inadequate exploration of superalignment.

\paragraph{Data Quality} The quality of synthetic data remains a significant concern. 
Numerous studies highlight the critical importance of data quality for alignment \citep{NEURIPS2023_ac662d74, chen2023maybe}. 
Training signals derived from teacher models are often noisy due to the inherent randomness in model generation.  
To address this issue, recent research has concentrated on two main paradigms: firstly, generating high-quality data by formulating detailed and refined principles, such as Orcas \citep{mukherjee2023orca, mitra2023orca} and AttrPrompt \citep{NEURIPS2023_ae9500c4}; and secondly, extracting relatively high-quality data from existing datasets by establishing evaluation metrics or employing filtering paradigms, such as Reflection-Tuning \citep{li2023reflectiontuning, li2024selective} and Phis \citep{li2023textbooks, javaheripi2023phi, abdin2024phi3} \footnote{Since numerous studies, e.g., \citet{wang2024survey}, conduct detailed investigations on data selection, we do not delve into this field here.}.
Additionally, some research suggests that alignment algorithms possess a certain degree of robustness \citep{gao2024impact}. Developing more robust training algorithms may thus be another approach to mitigating the issues associated with data quality.

\paragraph{Bias of the Teacher} Additionally, reliance on the teacher model may introduce biases and limitations inherent in the teacher model, which can affect the alignment effectiveness. 
Some studies propose introducing multiple teacher models to align the student model \citep{cui2024ultrafeedback, liu2024cosupervised}, thereby reducing the likelihood of the model overfitting to the biases of a single teacher model. Utilizing multiple teachers can also increase the diversity of signals, significantly enhancing alignment effectiveness \citep{song-etal-2024-scaling-data}.

\paragraph{Insufficient Understanding of Superalignment} Achieving superalignment remains a significant challenge. 
We still lack a strong scientific understanding of superalignment~\citep{burns2023weaktostrong}, hindering further exploration of weak-to-strong alignment. 
Additionally, most current approaches still require a sufficiently aligned "weak" model, and how to utilize a truly weak model for superalignment remains an issue. 
Some works present theoretical frameworks for understanding weak-to-strong generalization~\citep{charikar2024quantifying, lang2024theoretical, somerstep2024statistical}, but still have limited application scope. 
An interesting road is like ExPO~\citep{zheng2024weaktostrong}. ExPO extrapolates directly from an SFT model and an aligned model's weights, obtaining a better-aligned model without additional training, demonstrating a promising approach from weak to strong.

In conclusion, despite significant progress in instruction and behavior construction, current approaches still have significant limitations.
The core issue with strong-to-weak methods is that the alignment ceiling is constrained by the teacher model.
Conversely, works about weak-to-strong alignment remain underdeveloped, lacking theoretical analysis and generalized methodologies.  
Several critical issues must be addressed in the future, including enhancing data quality efficiently, developing more robust training algorithms, implementing multi-teacher imitation, and conducting theoretical analysis for weak-to-strong alignment in general tasks. 
Tackling these challenges will pave a feasible path for the further advancement of LLMs.
Moreover, we also provide an in-depth discussion about the underlying mechanisms of weak-to-strong alignment in Section~\ref{sec:mechanism}, which sheds light on a deeper understanding of this field.

\section{Aligning Through Model Feedback}
\label{sec:reward}
\epigraph{We all need people who will give us feedback. That's how we improve.}{Bill Gates}

\begin{figure}[t!]
\centering 
\includegraphics[width=0.7\textwidth]{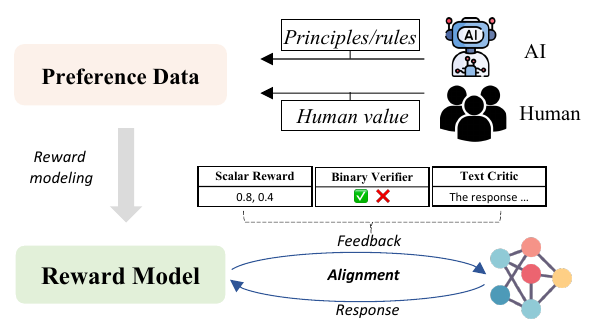}
\caption{The illustration of aligning through model feedback generated by reward models. The reward model will automatically generate the feedback of LLMs' responses in the scalar, binary or text format.}
\label{fig:reward}
\end{figure}

Human feedback reflects human values and can be used to align the LLMs, enabling LLMs to produce helpful and safe responses while correcting errors and toxic outputs. Unfortunately, accessing human feedback during training is challenging due to inefficiency and high costs.
To address this issue, model feedback is introduced as a way to estimate human feedback. This approach is commonly utilized in reinforcement learning, where a reward model generates feedback. 
Compared to relying on the limited feedback data generated by humans, the reward model can perform feedback prediction across a broader distribution, thus achieving more efficient alignment.
Aligning through automated generated model feedback offers an effective method for aligning LLMs with human values, presenting a promising path towards achieving automated alignment.
In this section, we explain how to leverage model-provided feedback to align it with human values. As shown in Figure~\ref{fig:reward}, related methods can be divided into three types based on the form of feedback signals: scalar~(\S~\ref{subsec:numerical}), binary~(\S~\ref{subsec:binary}), and text signals~(\S~\ref{subsec:text}).

\subsection{Scalar Reward}\label{subsec:numerical}
Scalar signals are commonly generated by a reward model that takes the response of LLMs as input to generate scalar signals for estimating human preferences. 
Reward model is frequently employed in reinforcement learning to align LLM with human values. 
In this way, LLMs can automatically align with human values by utilizing the large amount and diverse feedback provided by the reward model. 
To achieve more effective automated alignment, recent studies focus on how to train a higher-quality reward model and reduce the reliance on human annotation during the training of the reward model through model generation or pre-training.
Besides, the scalar signals generated by reward model can also be used to optimize the generation of LLM during decoding and filter training data for instruct-tuning.

\subsubsection{Reinforcement Learning from Human Feedback}
Reinforcement Learning from Human Feedback (RLHF) is a crucial paradigm for aligning LLMs with human values~\citep{NIPS2017_d5e2c0ad,stiennonLearningSummarizeHuman2020,ouyangTrainingLanguageModels2022b}. It typically involves three steps: 1) supervised fine-tuning (SFT), where LLMs are trained on annotated data to improve their responses to prompts; 2) training reward models to anticipate human feedback on model responses; and 3) employing reinforcement learning algorithms like Proximal Policy Optimization (PPO) ~\citep{schulman2017proximal} to align the model. In RLHF, the reward model, which is usually trained on preference data annotated by humans, produces scalar signals that mimic human feedback, serving as the guiding signal for learning. The performance of the reward model determines the potential upper bound of the model's alignment, so training the reward model is of vital importance~\citep{zheng2023secrets}. In the following sections, we first introduce related works about enhancing reward model. Then we introduce how to generate preference data without human effort. Finally, we introduce the functions of the reward model beyond reinforcement learning, including alignment during the decoding stage and SFT data filtering. 

\subsubsection{Improvement for Reward Modeling}
To achieve more effective automated alignment, it is crucial to improve the quality of model feedback. Therefore, recent studies focus on learning high-quality reward models.
The primary challenges in training reward models involve data collection and model optimization. Collected Preference data is typically sparse and deficient in consistency and detail, and model optimization can be hindered by issues such as over-fitting.

\paragraph{Reward Model Pre-training} Due to the data sparsity of existing datasets and the expense of human annotation, it is hard to train a high-quality reward model for automated alignment. To this end, \citet{askell2021general} propose the reward model pre-training. By collecting pair data from the network, including StackExchange, Reddit, and Wikipedia, they constructed a ranked dataset to pre-train a preference model.
By leveraging reward model pre-training, the reliance on human annotation is diminished~\citep{bai2022training}, which facilitates more efficient training of the reward model and enhances the effectiveness of automated alignment.

\paragraph{Consistent Preference Data Construction} Because human annotators have different evaluation principles and subjective perspectives, the feedback is diverse and includes multiple viewpoints. Previous studies have used strategies like multiply models ensemble~\citep{rame2023rewarded,touvron2023llama}, multi-objective learning~\citep{zeng2024diversified,zhong2024panacea,guo2024controllable,yang2024rewardsincontext} to mitigate the negativity from the diversity data. In contrast to reward models that produce a single score, ~\citet{li2024aligning} introduce the Distributional Preference Reward Model (DPRM) for predicting preference distributions.

\paragraph{Fine-grained Feedback Collection}Reward models often struggle to offer fine-grained feedback for intricate situations like safety and challenging tasks such as reasoning. To address this issue, some studies concentrate on refining reward models' training.~\citet{chen2024improving} introduce a token-level reward model capable of providing precise feedback at the token level, suitable for complex tasks like reasoning.~\citet{wu2023finegrained} suggest training multiple reward models that can deliver detailed feedback at the text span level.

\paragraph{Training Optimization}The learning process of the reward model usually faces the problem of over-optimization. That is, through learning, the reward model performs poorly instead. ~\citet{pmlr-v202-gao23h} analyze this phenomenon through experiments and discover the scaling law of the reward model to guide learning.~\citet{pmlr-v202-zhu23f} provide a theoretical analysis of the reward model training in RLHF and show the importance of introducing pessimism when training. Moreover, some other works have adopted various ways to improve the performance of the reward model, including normalization~\citep{zheng2023secrets} and iterative learning~\citep{touvron2023llama}.

Although the purpose of the reward model is to predict human feedback, modeling the reward is challenging. Therefore, how to build a more comprehensive reward model to achieve automated alignment is an important research question.

\subsubsection{Reinforcement Learning from AI Feedback}
Reward models are commonly trained using human feedback which is hard and expensive to annotate. 
For the purpose of reducing human effort and improving the automation in alignment, some works use the existing large language models to generate the preference data. Reinforcement Learning from AI Feedback (RLAIF)~\citep{lee2023rlaif} trains the reward model with preference data of LLM and it can achieve comparable or superior performance to RLHF. The methods are mainly divided into two types, including ranking the responses of multiple models and directly generating positive and negative responses. In this way, the automated alignment can be achieved in the entire procedure of reinforcement learning without human effort.

\paragraph{Ranking Multiple Responses}With the improved capabilities of LLMs, directly using them to rank multiple responses can provide preference data~\citep{tunstall2023zephyr,hong2023cyclealign,guo2024direct,pace2024west,yuan2024self}. This ranked preference data can also be produced with minimal human supervision, such as through human-defined principles~\citep{bai2022constitutional,sun2023salmon} or rules~\citep{kim-etal-2023-aligning}. To improve the quality of generated preference data,~\citet{shi2024saferinstruct} propose a carefully designed pipeline including reversed instruction tuning, instruction induction, and expert model evaluation.
~\citet{liu2024direct} propose to score the response using contrastive prompt pairs as input, which can achieve better performance compared to directly generated feedback with a single prompt.

\paragraph{Generating Positive and Negative Responses}Some works use LLM to generate preference data directly by prompting it to output positive response and negative response~\citep{chen2024grath}. ~\citet{yang2024rlcd} use an indirect approach by using different prompts to generate positive and negative responses, respectively. 

Although promising, the main challenge of it is the quality of preference data. Since the LLMs are commonly interfered with many aspects during generating, such as position bias~\citep{zheng2024judging,wang2023large}, the generation of high-quality preference data still requires further exploration. With the continuous improvements in LLMs, using LLM to reduce human effort will be a key strategy for automatically aligning models in the future.

\subsubsection{Reward Model-Guided Decoding}
In addition to learning directly from preference data, the generation of LLM can be enhanced by the scalar signals provided by reward model. This allows alignment to be directly achieved in the output rather than within the model via re-weighting the probabilities of tokens~\citep{mudgal2024controlled}.~\citet{lango-dusek-2023-critic} propose a critic-driven decoding approach to adjust the probability of token during generating via a binary classifier as the critic model.~\citet{deng-raffel-2023-reward} propose Reward-Augmented Decoding (RAD) that employs an attribute-specific reward model to re-weight the top-k highest probabilities when decoding. To achieve flexible alignment in different tasks, \citet{liu2024decodingtime} propose decoding-time realignment (DeRa) to control the alignment level during decoding.

Performing automated alignment at only the decoding stage is a simple method to avoid consuming a large amount of computing resources. However, alignment during decoding usually requires more time to conduct inference, and the quality of response still needs to be further improved.

\subsubsection{Filtering SFT Data using Reward Model}
High-quality SFT plays a crucial role in enhancing LLM performance~\citep{NEURIPS2023_ac662d74}. Therefore some studies use reward models to filter the training data. The main paradigms are categorized into learning from best response and learning from ranked results. Learning from the best response is often referred to as Best of N or Reject sampling~\citep{touvron2023llama,yuan2023scaling}. This approach typically involves using a reward model to choose the high-quality data among multiple responses for refining the model~\citep{dong2023raft}. In addition to learning from top responses, the LLM can also learn from ranked data.~\citet{yuan2023rrhf} propose Rank Responses to align Human Feedback (RRHF) to align LLM using ranking loss.~\citet{lu2022quark} propose using a reward model to grade data based on their scores and employing various reward tokens to regulate their generation. This approach helps prevent the learning of undesirable behaviors.

Besides reinforcement learning, SFT is also an important way to achieve alignment. By filtering data through the reward model, the LLM can automatically align with human values through SFT. Similar to the previous problem, the quality of SFT data is highly dependent on the quality of the reward model and needs further research.

\subsection{Binary Verifier}\label{subsec:binary}
For some objective tasks, such as mathematical problems, the reward model usually transforms into a verifier with binary signals. Given that mathematical problems usually require complex step-by-step reasoning, the verifier can be divided into outcome verifier and process verifier. Outcome verifier is used to estimate the correctness of the final answer. Process verifier accesses the intermediate step, which requires a large number of supervision data. Through the binary verifier, the LLMs can achieve automated alignment on these objective tasks.

\paragraph{Outcome Verifier}To improve the reasoning ability of LLMs, some studies focus on selecting reasoning paths generated by LLMs using golden answers for training~\citep{NEURIPS2022_639a9a17,singh2024beyond}. Since golden answers are costly to acquire, an outcome verifier is utilized to forecast the correctness of generated answers. This verifier is typically trained using LLM-generated correct and incorrect rationales~\citep{cobbe2021training}, and is used to fine-tune the LLM through different strategies including direct tuning~\citep{liu2023improving} and iterative training~\citep{hosseini2024vstar}.
Since outcome verifier cannot assess the correctness of reasoning steps, ~\citet{havrilla2024glore} propose Stepwise Outcome Reward Models (SORMs) that predict whether a step will lead to the correct answer. Besides training, \citet{yu2024ovm} propose the Outcome-supervised Value Model (OVM) which is used to guide decoding.

\paragraph{Process Verifier}
Even if the final answer is correct, there may still be errors in the reasoning process, limiting the effectiveness of the outcome verifier. To address this problem, the process verifier is employed to assess the correctness of reasoning step for more detailed verification~\citep{lightman2023lets,uesato2022solving}. The process verifier can be used to train a more effective reasoner~\citep{ying2024internlmmath,shao2024deepseekmath}. 
Inspired by the human reasoning mechanism,~\citet{zhu-etal-2023-solving} propose Cooperative Reasoning (CoRe) to produce synthesized training data for reasoning where process verifier is used to generate the feedback of the generation of generator.
Many studies are dedicated to training the verifier using automatically generated data due to the difficulty of collecting step-wise supervised signals.
~\citet{wang2024mathshepherd} and~\citet{wang2024multistep} train the process verifier with automatically constructed data that is collected by Monte Carlo Sampling.
Moreover, the process verifier can be applied in decoding to select the correct reasoning path~\citep{khalifa-etal-2023-grace}. Some studies focus on how to complete the final reasoning path efficiently.~\citet{ma2023lets} propose a heuristic greedy search algorithm based on the verifier's feedback.~\citet{li-etal-2023-making} propose using the verifier to filter the reasoning steps generated using diverse prompts.

Binary verifier is important for achieving automated alignment in objective tasks such as math. However, training the verifier, especially the process verifier is very difficult and requires a large amount of annotated data. Therefore, in the future, in order to further achieve automated alignment, researchers can focus on how to automatically construct a process verifier.

\subsection{Text Critic}\label{subsec:text}
Text signals contain more semantics than scalar and binary signals, enabling models to intuitively align with humans. By integrating text feedback, LLMs can improve their output alignment with humans. These refined outputs can then be used as supervised data for further aligning the LLMs~\citep{scheurer2022training,chen2024learning}. Text signals generated by the critique model have shown the potential to improve the outputs of LLMs, with feedback typically obtained through prompting LLMs. The text critic can be other LLMs (such as GPT-4)~\citep{koutcheme2024open,an2024learning} or LLM itself (i.e., self-critique)~\citep{saunders2022self,wang2023shepherd}. Since the self-critique of LLM remains a challenge~\citep{luo2023critique}, text signals for alignment are still underexplored.

Existing studies mainly focus on using the text critic to achieve automated alignment by improving the output of LLMs. In the future, exploring how to use the text critic to achieve more diverse automated alignments, such as in training, is an important research direction. 

\section{Aligning Through Environment Feedback}
\label{sec:environment}
\epigraph{We do not learn from experience... we learn from reflecting on experience.}{John Dewey}

\begin{figure*}[!tp]
\centering
\includegraphics[width=0.96\textwidth]{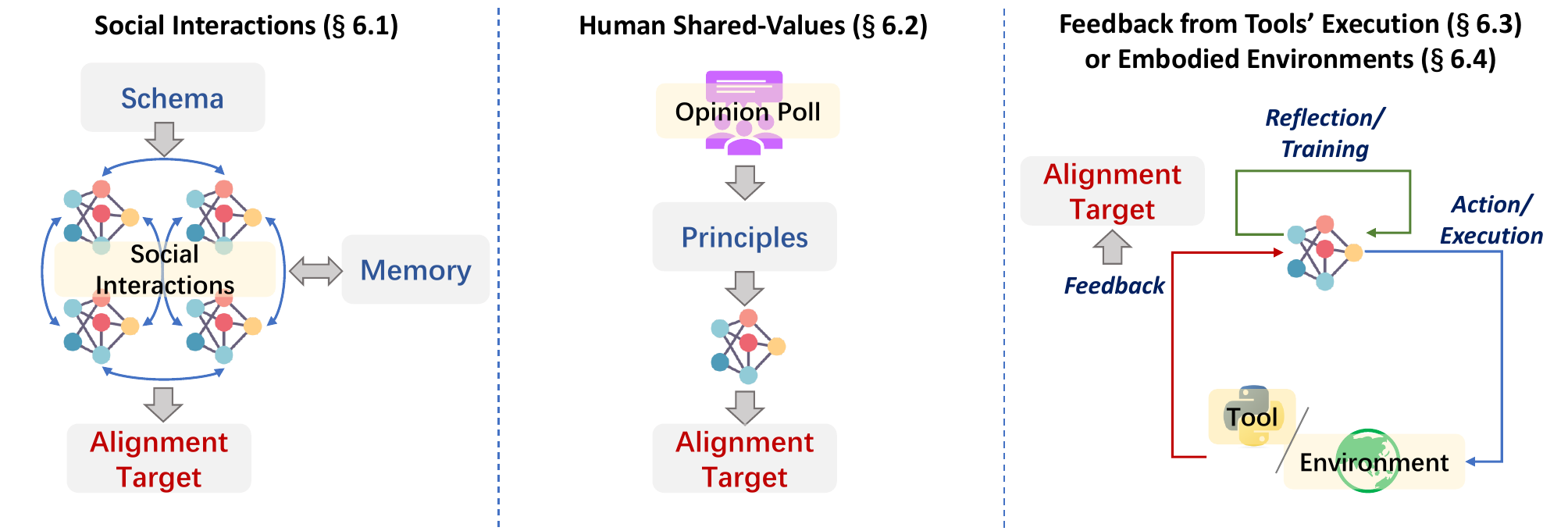}
\caption{Alignment through various types of environment feedback by social interactions, human shared-values, signals from tool execution or embodied environment.}
\label{Fig.sec6.1}
\end{figure*}

This section involves automatically obtaining alignment goals or feedback from the existing environments to achieve automated alignment of the target model. 
As illustrated in Figure~\ref{Fig.sec6.1}, according to the category of environments the model interacts with, we overview four lines of current research in this section and systematically go through the representative works of each part:

\begin{itemize}[leftmargin=1em]
    \item \textbf{Social Interactions} (\S~\ref{subsec:socialinter}), where models communicate with each other to build up a multi-agent system and collect alignment signals through such interactive communication.
    \item \textbf{Human Shared-Values} (\S~\ref{subsec:human}), where models receive judgements from human societies to calibrate their internal values.
    \item \textbf{Tool Execution Feedback} (\S~\ref{subsec:execution}), where models interact with external tools to receive instant feedback and achieve various intelligence by learning to use tools smartly.
    \item \textbf{Embodied Environment} (\S~\ref{subsec:embodied}), where models act as language interfaces in a physical world and get reward based on task goals.
\end{itemize}

\subsection{Social Interactions}\label{subsec:socialinter}

Social interaction is one of the fundamental properties of human society, in which many social norms are conveyed and followed in implicit ways. 
Recent advancements of large language models provide the opportunity to construct LLM-based agent system to simulate such interactions among human society, enabling build of sandbox environments to gather alignment signals with more scalability and akin to the real world~\citep{park2023generative}. 
Through the simulated social interactions such as moral discussions, studies offer promising avenues for enhancing AI systems' alignment with human values and ethical principles.

For instance, Stable Alignment~\citep{liu2023training} draws inspiration from how humans learn to navigate social norms and reach consensus on value judgments of societal issues through discussions.
They introduce an LLM-based multi-agent framework to simulate social interactions within human society, 
highlighted by a three-tiered method comprising imitation, self-critic and realignment. 
Alignment datasets are extracted during the simulation process, correspondingly.
\citet{wang2024sotopiapi} extend this method to multi-turn interactions under realistic social scenarios. They propose an interactive learning method to teach language agents to reach goals through social interactions among characters playing different social roles.
\citet{pang2024selfalignment} introduce social scene simulation which employs LLM to play different social roles in scenario relative to the instruction, enabling the model to take social consequences behind the instruction into account to revise its initial response accordingly.
Building upon similar methodology of simulated social interactions, other works explore the intricacies of moral discussions~\citep{sun-etal-2023-moraldial}, political debates~\citep{taubenfeld2024systematic} and task-oriented dialogue~\citep{ulmer2024bootstrapping} wherein AI systems are guided by some predefined discussion schema.

% Human Shared-Values
% Human Collective Intelligence
\subsection{Human Shared-Values}\label{subsec:human}

Another way compensate to simulated social interactions involves resorting to collective efforts from human to derive principles that AI should align to.  % crowd-sourcing
% However, distinct from previous crowd-sourcing annotations in traditional NLP or RLHF data collection, alignment signals are collected in more primitive forms to achieve scalable oversight. % broader scalability of supervision and oversight.
% These signals are often termed as norms~\citep{ammanabrolu-etal-2022-aligning}, Rules of Thumb (RoTs)~\citep{forbes-etal-2020-social, ziems-etal-2022-moral}, or Constitutions~\citep{bai2022constitutional} in different contexts.
Currently, in this line of works, alignment signals are usually termed as norms~\citep{ammanabrolu-etal-2022-aligning}, Rules of Thumb (RoTs)~\citep{forbes-etal-2020-social, ziems-etal-2022-moral}, or Constitutions~\citep{bai2022constitutional} in different contexts, which provide greater scalability compared to the data-point-level annotations in traditional NLP and RLHF data collections.
% Through such process, consensus on what basic principles AI systems should follow can be achieved among groups of people, and shared-values can be reached by synthesizing various human ideas on values into a target for aligning models~\citep{klingefjord2024human}.
% The expectation is to achieve consensus on what basic principles AI systems should follow among groups of people, and reach the human shared-values by synthesizing various human ideas into a set of target for model alignment~\citep{klingefjord2024human}.
%collective intelligence can be reached
In Constitutional AI (CAI)~\citep{bai2022constitutional}, researchers have designed a set of AI Constitutions. However, such in-house constitutions compiled by members from a research team fall short to broadly represent values of different groups of people.

% Toward this goal, current works are carried out to mainly solve two aspects of research questions.
% One of the questions is how to extract alignment signals from human values. 
% For this question, Constitutional AI (CAI)~\citep{bai2022constitutional} provides a basic attempt to convert the AI Constitutions into supervision and preference datapoints. 
% In their setting, human supervision relies entirely on a set of principles governing AI behavior, called Constitutions.
% A small number of examples are used for few-shot prompting LLMs to create SFT data or preference feedback according to those constitutional principles.

% The other question is how to elicit values from human public more effectively.
% To enable broader public to collectively shape the behavior of AI systems, follow-up efforts are dedicated to 
In order to achieve consensus on what basic principles AI systems should follow among groups of people and reach human shared-values by synthesizing various human ideas into a set of target for model alignment, follow-up efforts are dedicated to enable broader public to collectively shape the behavior of AI systems.  %  AI democratization, aiming to 
For example, in a program titled Democratic inputs to AI\footnote{\url{openai.com/blog/democratic-inputs-to-ai}}, an open call for the design of prototype systems to promote the democratic process of AI was issued.
Specifically, it addresses enabling a representative cross-section of people to exchange views and ultimately reach a consensus on the rules AI systems should follow.
Subsequently, in the Collective Constitutional AI (CCAI) project~\citep{collective_constitutional_AI, Huang2024collective}, researchers establish a multi-stage process for sourcing and integrating public input into LMs based on questionnaire. % participants to draft AI constitutions.
% design questionnaire to ask participants to draft a constitution for AI system.
Through the way of collective constitution drafting, participants can assess the existing constitutional principles, rate their acceptance scores, and propose new constitutions they believe AI should be better to adhere to, thus reduce the bias of aligned LLMs.
Beyond these pioneering research, large language models themselves are also involved to help conclude shared-values from human crowds more efficiently.
\citet{klingefjord2024human} propose Moral Graph Elicitation (MGE) to elicit and reconcile values from human participants, which utilizes an LLM to interview participants about their values in particular contexts.

\subsection{Tool Execution Feedback}\label{subsec:execution}

Tools are pivotal in expanding the capabilities of large language models, allowing them to transcend the limitations beyond their basic capabilities and interact more effectively with their surroundings.
Additionally, accurate and elaborate feedback from tool execution provides direct signals for LLMs to validate and enhance their initial outputs~\citep{chen2023teaching, gou2024critic}, which helps to reduce the reliance on human feedback.
Furthermore, the process of learning to plan and use tools and can be refined from tools execution feedback~\citep{wang2024llms}, where LLMs receive feedback on their actions and learn from both successes and failures in an interactive way.

Tools' execution feedback can serve as additional signals beyond human-labor corpus to better ground models to tool usage and reduce their hallucinations during interacting with tools.
Code generation tasks provide such examples. 
CodeRL~\citep{le2022coderl} conducts unit tests by code compiler to receive feedback signals and use them to train a critic model, with which the code generation model is further trained using deep reinforcement learning.
Self-debugging~\citep{chen2023teaching} designs an interactive code debugging pipeline by adding a code explanation stage beside the code execution feedback, where LLMs are asked to debug its own generated code given these feedback information. 
Similarly, SelfEvolve~\citep{jiang2023selfevolve} receives error message from interpreter and refine the answer code based on this feedback.

In addition to code generation, there are also studies designing unified frameworks to align language models with the execution feedback from multiple other sources.
CRITIC~\citep{gou2024critic} utilizes a verify-then-correct process to obtain external feedback from various external tools, including search engines, code interpreters and text APIs, thus enabling language models correcting the outputs with tool-interactive critiquing.
Taking a step further, \citet{wang2024llms} augment LLMs with a dynamic memory mechanism, which allows the LLMs to progressively learn how to accurately use tools.
\citet{qiao2024making} propose Reinforcement Learning with Execution Feedback to reinforce LLMs with execution results of tools.

\subsection{Embodied Environment}\label{subsec:embodied}

Embodied intelligence involves agents that perceive signals and take actions within some physical environments, requiring not only language understanding but also abilities of reasoning and decision making in a space with large number of states~\citep{Embodied-Intelligence-Overview}.
Leveraging large language models in embodied AI settings represents a compelling endeavor which can bring benefits from two aspects. On one hand, the powerful generalization ability of LLM can be utilized to facilitate action plan in natural language without preliminary learning. 
On the other hand, commonsense knowledge and physical understanding ability of LLMs can be aligned with feedback signals from environment.

Recent studies show the potential of LLMs to be deployed in robotics for real-world interactions while also pose challenges for grounding LLMs to understand the physical world~\citep{wang2023voyager, ahn2022i}.
To better align language models which have not been pre-tarined on those embodied environment, 
\citet{xiang2023language} adopt an embodied simulator as world model to provide feedback signals when LLMs interact with objects and execute actions in this environment with or without a task goal. Then the explored world states would be collected as embodied experiences to fine-tune the LLM offline subsequently.
In parallel, other studies utilize online reinforcement learning~\citep{carta2023grounding, tan2024true} to learn or iterative offline learning framework~\citep{song2024trial} to update policy for embodied agent through trial and error in the interactive environments.
Beyond aligning  specialist agents in isolated environments,
\citet{xi2024agentgym} introduce a novel framework called AGENTGYM to develop generally-capable LLM-based agents by facilitating their evolution through diverse environments and tasks. 
% AGENTGYM is equipped with an interactive platform of multiple agent environments, a benchmark suite of high-quality interactive trajectories across environments, along with a new algorithm that demonstrates promising results in enabling agents to self-evolve through manifold environmental feedbacks.

\subsection{Discussion}

In this section, we reviewed some current studies that can be uniformly observed from the perspective of aligning through environment feedback. While most of the works target at specific downstream applications, actively learning from environment has always been a key pursue in artificial intelligence.
Despite their valuable contributions, current methods still show limitations and leave open problems for future research. 
% The main limitation of XXXX arises from 
% The main limitation arises from the gap between simulated environments and the real world. 
One of the primary limitations arises from the gap between environment signals inspected in research and those in the real world. 

    On the one hand, simulated environments may be lake of \textit{representation ability} to fully catch the complexity of real-world.
    For example, to derive social interaction signals for alignment, most of the current works are based on simulated environments, while feedback from the real environment can be more noisy or ambiguous. For example, \citet{liu2023training} point out that current studies usually assume a static view of social norms, overlooking their dynamic and evolving natures.
    Similarly, for the embodied environment track of works, research are often conducted within sandbox environments and different settings are mutually independent. The action spaces of these simulated environments are still limited~\citep{tan2024true} compared to the real environment with potentially infinite degree of freedom.
    As a result, models trained in one setting might not generalize well to others, and ensuring models maintain alignment across different environments remains a challenge. 

    % Most of the current works derive alignment signals from simulated environments, while feedback from the real environment can be noisy or ambiguous.
    % For example, in aligning through social interaction, \citet{liu2023training} point out that current studies usually assume a static view of social norms, overlooking their dynamic and evolving natures.
    % In fact, social values are complex, context-dependent, and sometimes contradictory. Encoding these values into AI systems in a comprehensive and coherent manner is a significant challenge. Thus, how to effectively bridge these changeable and divergent social values is of critical importance in alignment with human society.
    % In addition, simulated environments can be oversimplified and cannot fully catch the complexity of real-world, which challenges the robustness and generalization of aligned models. 
    % As for the embodied environment track of works, research are often conducted within sandbox environments and different settings are mutually independent. The action spaces of these simulated environments are still limited~\citep{tan2024true} compared to the real environment with potentially infinite degree of freedom.
    % As a result, models trained in one setting might not generalize well to others, and ensuring models maintain alignment across different environments remains a problem.  

    On the other hand, signals from human shared-values could still suffer from \textit{bias} even they are linked to the real-world environment.  
    % In fact, social values are complex, context-dependent, and sometimes contradictory. Encoding these values into AI systems in a comprehensive and coherent manner is a significant challenge. 
    For example, the values sampled from a small society groups can quite differ from the universally acknowledged ones. Additionally, voices on the internet may not always represent the actual values of most people in the real world and peoples' perceptions are tend to be influenced by social media. 
    Since alignment with human shared-values is still a nascent territory, most of current works are tentative and invite limited number of participants not more than thousands. The potential challenges of current methods when scaling to hundreds of thousands of people may be unanticipated.
    As a result, how to inject human values into AI systems in a comprehensive and coherent manner to achieve as much unbiased alignment as possible is still an open problem.  %  AI democratisation and 

In conclusion, the topic of integrating environmental feedback into AI alignment opens up a promising and emerging track, while significant research problems remain to be explored. 
Future research can focus on bridging the gap between simulated and real environments to develop more adaptive, reliable, and unbiased AI. 
% the gap between simulated and real environments
% ethically aligned

% 1. 模拟环境和现实环境的gap
% 2. 现在纳入的环境存在的bias/局限性

\section{Underlying Mechanism of Automated Alignment}
\label{sec:mechanism}
As discussed in the previous sections, substantial research endeavors to enhance the efficiency, effectiveness, and reliability of automated alignment methods.
Despite these efforts, there remains a conspicuous dearth of systematic investigation into the mechanisms underlying alignment.
For example, numerous methods~(\citealp{liu2024what}; \citealp{li2024quantity}; etc.) are proposed to filter instruction data, however, it remains unclear whether specific filtering criteria or processes are necessary and the rationale behind them.
Likewise, many works propose various weak-to-strong alignment algorithms, but we still do not know the reasons behind the success of weak-to-strong generalization and whether it depends on specific conditions.
The absence of relevant research can impede comprehension of current alignment and automated alignment methods, thereby hindering further optimization of these methods.

Therefore, in this section, we will conduct a systematic investigation into the mechanisms of automated alignment. 
By systematically organizing and analyzing the underlying mechanisms of automated alignment, we can identify the shortcomings and limitations of current automated alignment methods and illuminate directions for designing improved approaches.
As previously discussed, there are an enormous amount of techniques of automated alignment. 
In this survey, we select the following three core research questions, which are crucial for achieving scalable automated alignment:
\begin{itemize}[leftmargin=1em]
    \item \textit{\textbf{RQ1:}} \textbf{What is the underlying mechanism of current alignment?} The underlying mechanism of alignment is fundamental to the study of automated alignment. It is crucial for understanding the feasibility, boundaries, and optimization directions of automated alignment.
    \item \textit{\textbf{RQ2:}} \textbf{Why does Self-feedback work?} Self-feedback is widely applied across various paradigms of automated alignment, such as serving as an inductive bias in section~\ref{sec:H2: LLMs can judge, critic, refine and more}, constructing data for preference learning in section~\ref{sec:reward} and acting as a crucial technique for automated iterative alignment~(\citealp{yuan2024self}; etc.).
    Understanding why does it work is vital for comprehending and enhancing all paradigms involving it. 
    \item \textit{\textbf{RQ3:}} \textbf{Why is \textit{weak-to-strong} feasible?} 
    As a promising direction for achieving scalable oversight, it is essential to comprehend the feasibility and underlying mechanisms of \textit{weak-to-strong} for optimizing and designing more effective \textit{weak-to-strong} methods, especially when aligning superhuman models.
\end{itemize}
For each research question, we summarize existing studies and perspectives and discuss the limitations of these analytical works as well as the areas that still require exploration.

\subsection{What is the underlying mechanism of current alignment?}
Understanding the mechanism underlying current alignment is essential for assessing the potential for automated alignment, examining key challenges faced by automated alignment and steering the optimization directions of current automated alignment methods.
Previous works~(\citealp{NEURIPS2023_ac662d74}; \citealp{ren2024learning}; \citealp{mecklenburg2024injecting}; etc.) mainly focus on the two aspects of alignment's underlying mechanism: behavioral norms transfer and knowledge learning.
Analyses and discussions are launched around which aspect is more crucial.

Some studies discover that the primary role of current alignment is the transform of behavioral norms rather than the learning of additional world knowledge.
\citet{NEURIPS2023_ac662d74} propose the ``Superficial Alignment Hypothesis'', that a model’s knowledge and capabilities are learned almost entirely during pretraining, while alignment teaches it which subdistribution of formats should be used when interacting with users. 
Moreover, subsequent research employ three distinct analytical methodologies to delve deeper into model behavior during the alignment process.

    \paragraph{Feature-based Analysis}
    By comparing the probability distribution of LLM's predicted tokens before and after alignment~(DPO\&RLHF), \citet{lin2024the} find that the distribution shift mainly occurs on stylistic tokens, while the knowledge-intensive tokens show a small distribution shift, and the distribution shift decreases as the predictions become longer, indicating that alignment is about formal alignment rather than knowledge injection.
    Meanwhile, \citet{duan2023exploring} find that ICL and IFT exhibit high convergence in the hidden states of LLM and low convergence with the hidden layer states of base LLM, suggesting that the role of alignment lies in the behavioral norms shift of the model from write continuation to response.
    Additionally, through employing gradient-based input-output attribution methods and analyzing attention heads and feed-forward layers, \citet{wu2024language} reveal that IFT enables LLMs to recognize user instructional components and tailor responses accordingly, without altering linguistic structure.
    \paragraph{Knowledge Intervention} \citet{ren2024learning} introduce a knowledge intervention framework to decouple the potential underlying factors of IFT and find that for IFT, there is little, if not even causing additional damage, benefits from the learning of world knowledge incongruent with parameter knowledge across all homogeneous, in-domain, and out-of-domain evaluations. Moreover, \citet{ren2024learning} find that the essence of effective IFT is to maintain the consistency of model parameter knowledge before and after IFT while completing the transform of behavioral norms. 
    \citet{gekhman2024does} also highlight the risks of adding new facts by observing the performance of the model in introducing different proportions of new knowledge through IFT.
    \paragraph{Empirical Assessment} 
    LIMA~\citep{NEURIPS2023_ac662d74}, AlpaGasus~\citep{chen2024alpagasus} and LTD~\citep{chen2023maybe} provide experimental support for ``Superficial Alignment Hypothesis'' by achieving impressive performance under IFT using few instruction data.
    Besides, \citet{gudibande2023false} show that alignment through behavior imitation can successfully improve LLM's style, role, and ability to follow instructions, but cannot improve LLM's performance on more complex dimensions, such as factuality and problem-solving. 

\paragraph{}
While these studies reach similar conclusions, they fail to delineate the specific alignment achieved by the models under scrutiny.
The potential differences in the underlying mechanisms of alignment across varying degrees or requirements remain unexplored.
Furthermore, current research primarily focuses on traditional NLP tasks and general conversational scenarios.
However, some studies~(\citealp{mecklenburg2024injecting}; \citealp{singhal2023expertlevel}; etc.) find that domain-specific alignment can indeed improve model performance on the corresponding domain, which suggest that alignment can play a crucial role in learning additional domain knowledge. 
The underlying mechanisms of alignment in various scenarios such as coding and mathematics remain an open question.

\subsection{Why does self-feedback work?}
Feedback capability refers to the capacity to provide information or guidance to a given input based on specific standards.
As we discussed above, such capability is extensively utilized in various paradigms of automated alignment.  
For example, in RLAIF, LLM itself substitutes for human assessment to construct preference data~(\citealp{yuan2024self}; etc.). 
Additionally, LLM can continuously optimize its output based on self-feedback, a process known as self-refine~(\citealp{bai2022constitutional}; \citealp{madaan2023selfrefine}; \citealp{tan-etal-2023-self}; etc.).
Nonetheless, there is much debate about whether and why LLMs can provide effective feedback to their own responses.
In the following, we systematically summarize each representative viewpoint.

\citet{li2024benchmarking} and \citet{lin2024criticbench} suggest that models possess certain knowledge that cannot be directly utilized for generation but can be employed for providing feedback.
\citet{li2024benchmarking} also discover significant Generator-Validator inconsistency in LLMs when generating and validating answers.
Similarly, \citet{lin2024criticbench} find that LLMs possess substantial knowledge that cannot be expressed through generation and correction but can be utilized for critique. 
Furthermore, other studies~\citep{yuan2024self,li2024selfalignment} argue that feedback capability is a byproduct of the model's ability to follow instructions. 
Therefore, during the alignment process, the model's feedback capacity improves alongside its ability to follow instructions, a phenomenon supported by experimental evidence~\citep{yuan2024self,li2024selfalignment}.
However, some other studies posit that the feedback capability of LLM is illusory, suggesting that it may rely on specific data and exhibit biases.
\citet{west2023generative} note that while current LLMs gain generative capabilities to replicate expert-level outputs through training, they lack understanding capacities relevant to critique.
Moreover, \citet{huang2024large} observe that for reasoning tasks, LLMs even experience performance degradation after self-correction at times because that they cannot properly judge the correctness of their reasoning.
Additionally, \citet{zheng2024judging} find that although powerful LLMs like GPT-4 achieve high agreement with human evaluators, similar to human-human agreement levels, LLM-based evaluation face various challenges such as position bias~\citep{zheng2024judging,wang2023large}, verbosity bias~\citep{zheng2024judging,wu2023style}, self-enhancement bias~\citep{zheng2024judging,liu2023geval}, as well as limited capability in certain scenarios such as grading math, reasoning questions~\citep{zheng2024judging}, high-quality summaries~\citep{shen2023large} and so on~\citep{lan2024criticbench}.
Researchers propose many methods to further improve the capability of self-feedback, such as meta critic~\citep{sun2024critique}, autocalibrate~\citep{liu2023calibrating}, split and merge~\citep{li2023split} and using external tools~(such as search engines, code interpreters, etc.) to cross-check~\citep{gou2024critic}, thus refining their initial response, etc.

\textbf{\textit{Discussion}}
While extensive work has been conducted on exploring models' capability to provide self-feedback, questions regarding its effectiveness boundaries and underlying reasons remain unanswered. 
Additionally, the disparities and rationales between model self-feedback and human expectations remain unexplored.
Moreover, promising results have been demonstrated in the two-step optimization model outputs~\citep{bai2022constitutional,madaan2023selfrefine,tan-etal-2023-self} of self-feedback and correction.
However, studies mainly focus on the former, leaving the latter largely unexplored.
Although researchers~\citep{lan2024criticbench,gou2024critic} find that LLMs possess the ability, akin to humans, to modify their response based on feedback, their capacity to correct based on the feedback they generate remains uncharted. 
Furthermore, the overall performance enhancement of LLMs depends on whether the number of refining incorrect responses to correct ones outweighs the reverse.
A comprehensive analysis of when self-refine can enhance or impair performance, as well as the underlying reasons for such effects, is still lacking.

\subsection{Why is \textit{weak-to-strong} feasible?}
As we discussed above, a promising approach towards scalable oversight is the concept of ``weak-to-strong'', which involves cultivating robust abilities through limited or simplified supervision.
While there are some successful practical works for ``weak-to-strong'', the underlying mechanisms of ``weak-to-strong'' remain further investigation, which limits further optimization and method design for ``weak-to-strong''.
Currently, a common perspective is that LLMs pre-trained on a large corpus can leverage their impressive generalization abilities to achieve robust capabilities even under limited or simplified supervision.
In the following, we will introduce how such generalization capability facilitates LLMs to achieve automated alignment via limited or simplified supervision.

\citet{bai2022constitutional}, \citet{sun2023principle} and \citet{franken2024selfsupervised} observe that merely supplying core alignment principles to LLMs enables them to automatically achieve significant alignment effects. 
This illustrates the models' capability to generalize from principle to behavior.
For example, \citet{bai2022constitutional} prompt LLM to refine and select optimal responses based on provided principles to complete IFT and RLAIF.
\citet{sun2023principle} utilize 16 manually designed rules to guide the base model to generate high-quality instructions followed by self-distillation to achieve self-alignment.
\citet{chen2024iteralign} further employ a stronger LLM to automatically discover these constitutions.
Similarly, \citet{franken2024selfsupervised} align a stronger base model using constitutions by a weaker instruction-fined model.
\citet{burns2023weaktostrong} find that simple methods can often significantly improve weak-to-strong generalization of LLM: for instance, when finetuning GPT-4 with a GPT-2-level supervisor and an auxiliary confidence loss, GPT-4 can achieve close to GPT-3.5-level performance in NLP tasks.
This demonstrates the models' capacity to generalize from limited supervision to stronger performance.
Additionally, \citet{sun2024easytohard} and \citet{hase2024unreasonable} find LLMs trained on simple tasks can successfully generalize to hard tasks.
Recent research also discuss the feasibility of weak-to-strong generalization under theoretical frameworks~\citep{somerstep2024statistical,lang2024theoretical,charikar2024quantifying}.

\textbf{\textbf{\textit{Discussion}}}
Current LLMs demonstrate strong generalization capabilities. 
However, the fundamental reasons behind the significant differences in generalization capabilities between current LLMs and earlier pre-trained LMs remain to be explored.
Moreover, further exploration is necessary to delineate the boundaries of generalization, i.e., what can and cannot be generalized, as well as to understand the underlying reasons, which is essential for identifying key challenges and determining future research directions of alignment and automated alignment.

% \section{Discussions}

\section{Conclusions}
\label{sec:discuss}
This survey explores various techniques for scalable  automated alignment, categorizing them into four primary domains: aligning with inductive bias, behavior imitation, model feedback, and environment feedback.
The existing research illustrates multiple pathways toward achieving automated alignment, primarily addressing critical challenges such as scalable oversight.
Despite these advancements, we notice significant research gaps when we investigate the mechanism of current alignment, particularly concerning the reliability of self-feedback and the feasibility of weak-to-strong generalization.
Addressing these underexplored questions is essential for advancing automated alignment, enabling the safe and effective application of large language models in real-world scenarios. 
Future research endeavors are expected to bridge these gaps, ensuring that LLMs operate reliably and align with intended human values.

Furthermore, in the most optimistic projections, the gradual enhancement of LLM capabilities could ultimately result in models capable of independently conducting alignment research, thereby improving their own safety. For instance, Super Alignment project \citep{openai2023super} briefly outlines an ambitious plan to develop a narrowly-focused alignment research expert, shifting the human cognitive burden from generation to evaluation of alignment research proposals. Recent advancements in latest LLMs~\citep{claude35modelcard} have shown their potential to address domain-specific expert-level problems, as evidenced by the GPQA ~\citep{rein2023gpqa} benchmark and the ability of models to approach PhD students' performance in certain settings. Moreover, progress in AI-driven research across scientific disciplines, such as autonomous chemistry experimentation agents~\citep{boiko2023autonomous}, illustrates how AI systems could eventually tackle alignment problems with speed and thoroughness surpassing human capabilities.

\bibliography{merge}

\begin{thebibliography}{316}
\providecommand{\natexlab}[1]{#1}
\providecommand{\url}[1]{\texttt{#1}}
\expandafter\ifx\csname urlstyle\endcsname\relax
  \providecommand{\doi}[1]{doi: #1}\else
  \providecommand{\doi}{doi: \begingroup \urlstyle{rm}\Url}\fi

\bibitem[Abdin et~al.(2023)Abdin, Aneja, Bubeck, Mendes, Chen, Giorno, Eldan, Gopi, Gunasekar, Javaheripi, Kauffmann, Lee, Li, Nguyen, de~Rosa, Saarikivi, Salim, Shah, Santacroce, Behl, Kalai, Wang, Ward, Witte, Zhang, and Zhang]{javaheripi2023phi}
Marah Abdin, Jyoti Aneja, Sebastien Bubeck, Caio César~Teodoro Mendes, Weizhu Chen, Allie~Del Giorno, Ronen Eldan, Sivakanth Gopi, Suriya Gunasekar, Mojan Javaheripi, Piero Kauffmann, Yin~Tat Lee, Yuanzhi Li, Anh Nguyen, Gustavo de~Rosa, Olli Saarikivi, Adil Salim, Shital Shah, Michael Santacroce, Harkirat~Singh Behl, Adam~Taumann Kalai, Xin Wang, Rachel Ward, Philipp Witte, Cyril Zhang, and Yi~Zhang.
\newblock Phi-2: The surprising power of small language models., 2023.
\newblock URL \url{https://www.microsoft.com/en-us/research/blog/phi-2-the-surprising-power-of-small-language-models}.

\bibitem[Abdin et~al.(2024)Abdin, Jacobs, Awan, Aneja, Awadallah, Awadalla, Bach, Bahree, Bakhtiari, Behl, Benhaim, Bilenko, Bjorck, Bubeck, Cai, Mendes, Chen, Chaudhary, Chopra, Giorno, de~Rosa, Dixon, Eldan, Iter, Garg, Goswami, Gunasekar, Haider, Hao, Hewett, Huynh, Javaheripi, Jin, Kauffmann, Karampatziakis, Kim, Khademi, Kurilenko, Lee, Lee, Li, Liang, Liu, Lin, Lin, Madan, Mitra, Modi, Nguyen, Norick, Patra, Perez-Becker, Portet, Pryzant, Qin, Radmilac, Rosset, Roy, Ruwase, Saarikivi, Saied, Salim, Santacroce, Shah, Shang, Sharma, Song, Tanaka, Wang, Ward, Wang, Witte, Wyatt, Xu, Xu, Yadav, Yang, Yang, Yu, Zhang, Zhang, Zhang, Zhang, Zhang, Zhang, Zhang, and Zhou]{abdin2024phi3}
Marah Abdin, Sam~Ade Jacobs, Ammar~Ahmad Awan, Jyoti Aneja, Ahmed Awadallah, Hany Awadalla, Nguyen Bach, Amit Bahree, Arash Bakhtiari, Harkirat Behl, Alon Benhaim, Misha Bilenko, Johan Bjorck, Sébastien Bubeck, Martin Cai, Caio César~Teodoro Mendes, Weizhu Chen, Vishrav Chaudhary, Parul Chopra, Allie~Del Giorno, Gustavo de~Rosa, Matthew Dixon, Ronen Eldan, Dan Iter, Amit Garg, Abhishek Goswami, Suriya Gunasekar, Emman Haider, Junheng Hao, Russell~J. Hewett, Jamie Huynh, Mojan Javaheripi, Xin Jin, Piero Kauffmann, Nikos Karampatziakis, Dongwoo Kim, Mahoud Khademi, Lev Kurilenko, James~R. Lee, Yin~Tat Lee, Yuanzhi Li, Chen Liang, Weishung Liu, Eric Lin, Zeqi Lin, Piyush Madan, Arindam Mitra, Hardik Modi, Anh Nguyen, Brandon Norick, Barun Patra, Daniel Perez-Becker, Thomas Portet, Reid Pryzant, Heyang Qin, Marko Radmilac, Corby Rosset, Sambudha Roy, Olatunji Ruwase, Olli Saarikivi, Amin Saied, Adil Salim, Michael Santacroce, Shital Shah, Ning Shang, Hiteshi Sharma, Xia Song, Masahiro Tanaka, Xin Wang, Rachel
  Ward, Guanhua Wang, Philipp Witte, Michael Wyatt, Can Xu, Jiahang Xu, Sonali Yadav, Fan Yang, Ziyi Yang, Donghan Yu, Chengruidong Zhang, Cyril Zhang, Jianwen Zhang, Li~Lyna Zhang, Yi~Zhang, Yue Zhang, Yunan Zhang, and Xiren Zhou.
\newblock Phi-3 technical report: A highly capable language model locally on your phone, 2024.

\bibitem[Agarwal et~al.(2024)Agarwal, Singh, Zhang, Bohnet, Chan, Anand, Abbas, Nova, Co-Reyes, Chu, et~al.]{agarwal2024many}
Rishabh Agarwal, Avi Singh, Lei~M Zhang, Bernd Bohnet, Stephanie Chan, Ankesh Anand, Zaheer Abbas, Azade Nova, John~D Co-Reyes, Eric Chu, et~al.
\newblock Many-shot in-context learning.
\newblock \emph{ArXiv preprint}, abs/2404.11018, 2024.
\newblock URL \url{https://arxiv.org/abs/2404.11018}.

\bibitem[Ahn et~al.(2022)Ahn, Brohan, Brown, Chebotar, Cortes, David, Finn, Fu, Gopalakrishnan, Hausman, Herzog, Ho, Hsu, Ibarz, Ichter, Irpan, Jang, Ruano, Jeffrey, Jesmonth, Joshi, Julian, Kalashnikov, Kuang, Lee, Levine, Lu, Luu, Parada, Pastor, Quiambao, Rao, Rettinghouse, Reyes, Sermanet, Sievers, Tan, Toshev, Vanhoucke, Xia, Xiao, Xu, Xu, Yan, and Zeng]{ahn2022i}
Michael Ahn, Anthony Brohan, Noah Brown, Yevgen Chebotar, Omar Cortes, Byron David, Chelsea Finn, Chuyuan Fu, Keerthana Gopalakrishnan, Karol Hausman, Alex Herzog, Daniel Ho, Jasmine Hsu, Julian Ibarz, Brian Ichter, Alex Irpan, Eric Jang, Rosario~Jauregui Ruano, Kyle Jeffrey, Sally Jesmonth, Nikhil~J Joshi, Ryan Julian, Dmitry Kalashnikov, Yuheng Kuang, Kuang-Huei Lee, Sergey Levine, Yao Lu, Linda Luu, Carolina Parada, Peter Pastor, Jornell Quiambao, Kanishka Rao, Jarek Rettinghouse, Diego Reyes, Pierre Sermanet, Nicolas Sievers, Clayton Tan, Alexander Toshev, Vincent Vanhoucke, Fei Xia, Ted Xiao, Peng Xu, Sichun Xu, Mengyuan Yan, and Andy Zeng.
\newblock Do as i can, not as i say: Grounding language in robotic affordances, 2022.

\bibitem[Aksitov et~al.(2023)Aksitov, Miryoosefi, Li, Li, Babayan, Kopparapu, Fisher, Guo, Prakash, Srinivasan, Zaheer, Yu, and Kumar]{aksitov2023rest}
Renat Aksitov, Sobhan Miryoosefi, Zonglin Li, Daliang Li, Sheila Babayan, Kavya Kopparapu, Zachary Fisher, Ruiqi Guo, Sushant Prakash, Pranesh Srinivasan, Manzil Zaheer, Felix Yu, and Sanjiv Kumar.
\newblock Rest meets react: Self-improvement for multi-step reasoning llm agent, 2023.

\bibitem[Amini and Gallinari(2002)]{amini2002semi}
Massih-Reza Amini and Patrick Gallinari.
\newblock Semi-supervised logistic regression.
\newblock In \emph{ECAI}, volume~2, page~11, 2002.

\bibitem[Ammanabrolu et~al.(2022)Ammanabrolu, Jiang, Sap, Hajishirzi, and Choi]{ammanabrolu-etal-2022-aligning}
Prithviraj Ammanabrolu, Liwei Jiang, Maarten Sap, Hannaneh Hajishirzi, and Yejin Choi.
\newblock Aligning to social norms and values in interactive narratives.
\newblock In \emph{Proceedings of the 2022 Conference of the North American Chapter of the Association for Computational Linguistics: Human Language Technologies}, pages 5994--6017, Seattle, United States, 2022. Association for Computational Linguistics.
\newblock \doi{10.18653/v1/2022.naacl-main.439}.
\newblock URL \url{https://aclanthology.org/2022.naacl-main.439}.

\bibitem[An et~al.(2024)An, Ma, Lin, Zheng, Lou, and Chen]{an2024learning}
Shengnan An, Zexiong Ma, Zeqi Lin, Nanning Zheng, Jian-Guang Lou, and Weizhu Chen.
\newblock Learning from mistakes makes llm better reasoner, 2024.

\bibitem[Anthropic(2023)]{collective_constitutional_AI}
Anthropic.
\newblock Collective constitutional ai: Aligning a language model with public input, 2023.
\newblock URL \url{https://www.anthropic.com/news/collective-constitutional-ai-aligning-a-language-model-with-public-input}.

\bibitem[Anthropic(2024{\natexlab{a}})]{anthropic2024measuring}
Anthropic.
\newblock Measuring the persuasiveness of language models, 2024{\natexlab{a}}.
\newblock \url{https://www.anthropic.com/research/measuring-model-persuasiveness/}.

\bibitem[Anthropic(2024{\natexlab{b}})]{claude35modelcard}
Anthropic.
\newblock Claude 3.5 sonnet model card addendum, 2024{\natexlab{b}}.
\newblock URL \url{https://www-cdn.anthropic.com/fed9cc193a14b84131812372d8d5857f8f304c52/Model_Card_Claude_3_Addendum.pdf}.

\bibitem[Anwar et~al.(2024)Anwar, Saparov, Rando, Paleka, Turpin, Hase, Lubana, Jenner, Casper, Sourbut, et~al.]{anwar2024foundational}
Usman Anwar, Abulhair Saparov, Javier Rando, Daniel Paleka, Miles Turpin, Peter Hase, Ekdeep~Singh Lubana, Erik Jenner, Stephen Casper, Oliver Sourbut, et~al.
\newblock Foundational challenges in assuring alignment and safety of large language models.
\newblock \emph{ArXiv preprint}, abs/2404.09932, 2024.
\newblock URL \url{https://arxiv.org/abs/2404.09932}.

\bibitem[Askell et~al.(2021)Askell, Bai, Chen, Drain, Ganguli, Henighan, Jones, Joseph, Mann, DasSarma, et~al.]{askell2021general}
Amanda Askell, Yuntao Bai, Anna Chen, Dawn Drain, Deep Ganguli, Tom Henighan, Andy Jones, Nicholas Joseph, Ben Mann, Nova DasSarma, et~al.
\newblock A general language assistant as a laboratory for alignment.
\newblock \emph{ArXiv preprint}, abs/2112.00861, 2021.
\newblock URL \url{https://arxiv.org/abs/2112.00861}.

\bibitem[Azar et~al.(2024)Azar, Guo, Piot, Munos, Rowland, Valko, and Calandriello]{azar2024general}
Mohammad~Gheshlaghi Azar, Zhaohan~Daniel Guo, Bilal Piot, Remi Munos, Mark Rowland, Michal Valko, and Daniele Calandriello.
\newblock A general theoretical paradigm to understand learning from human preferences.
\newblock In \emph{International Conference on Artificial Intelligence and Statistics}, pages 4447--4455. PMLR, 2024.

\bibitem[Bai et~al.(2022{\natexlab{a}})Bai, Jones, Ndousse, Askell, Chen, DasSarma, Drain, Fort, Ganguli, Henighan, et~al.]{bai2022training}
Yuntao Bai, Andy Jones, Kamal Ndousse, Amanda Askell, Anna Chen, Nova DasSarma, Dawn Drain, Stanislav Fort, Deep Ganguli, Tom Henighan, et~al.
\newblock Training a helpful and harmless assistant with reinforcement learning from human feedback.
\newblock \emph{ArXiv preprint}, abs/2204.05862, 2022{\natexlab{a}}.
\newblock URL \url{https://arxiv.org/abs/2204.05862}.

\bibitem[Bai et~al.(2022{\natexlab{b}})Bai, Kadavath, Kundu, Askell, Kernion, Jones, Chen, Goldie, Mirhoseini, McKinnon, et~al.]{bai2022constitutional}
Yuntao Bai, Saurav Kadavath, Sandipan Kundu, Amanda Askell, Jackson Kernion, Andy Jones, Anna Chen, Anna Goldie, Azalia Mirhoseini, Cameron McKinnon, et~al.
\newblock Constitutional ai: Harmlessness from ai feedback.
\newblock \emph{ArXiv preprint}, abs/2212.08073, 2022{\natexlab{b}}.
\newblock URL \url{https://arxiv.org/abs/2212.08073}.

\bibitem[Bansal et~al.(2018)Bansal, Pachocki, Sidor, Sutskever, and Mordatch]{bansal2017emergent}
Trapit Bansal, Jakub Pachocki, Szymon Sidor, Ilya Sutskever, and Igor Mordatch.
\newblock Emergent complexity via multi-agent competition.
\newblock In \emph{6th International Conference on Learning Representations, {ICLR} 2018, Vancouver, BC, Canada, April 30 - May 3, 2018, Conference Track Proceedings}. OpenReview.net, 2018.
\newblock URL \url{https://openreview.net/forum?id=Sy0GnUxCb}.

\bibitem[Besta et~al.(2024)Besta, Blach, Kubicek, Gerstenberger, Podstawski, Gianinazzi, Gajda, Lehmann, Niewiadomski, Nyczyk, et~al.]{besta2024graph}
Maciej Besta, Nils Blach, Ales Kubicek, Robert Gerstenberger, Michal Podstawski, Lukas Gianinazzi, Joanna Gajda, Tomasz Lehmann, Hubert Niewiadomski, Piotr Nyczyk, et~al.
\newblock Graph of thoughts: Solving elaborate problems with large language models.
\newblock In \emph{Proceedings of the AAAI Conference on Artificial Intelligence}, volume~38, pages 17682--17690, 2024.

\bibitem[Boiko et~al.(2023)Boiko, MacKnight, Kline, and Gomes]{boiko2023autonomous}
Daniil~A Boiko, Robert MacKnight, Ben Kline, and Gabe Gomes.
\newblock Autonomous chemical research with large language models.
\newblock \emph{Nature}, 624\penalty0 (7992):\penalty0 570--578, 2023.

\bibitem[Bousmalis et~al.(2023)Bousmalis, Vezzani, Rao, Devin, Lee, Bauza, Davchev, Zhou, Gupta, Raju, Laurens, Fantacci, Dalibard, Zambelli, Martins, Pevceviciute, Blokzijl, Denil, Batchelor, Lampe, Parisotto, Żołna, Reed, Colmenarejo, Scholz, Abdolmaleki, Groth, Regli, Sushkov, Rothörl, Chen, Aytar, Barker, Ortiz, Riedmiller, Springenberg, Hadsell, Nori, and Heess]{bousmalis2023robocat}
Konstantinos Bousmalis, Giulia Vezzani, Dushyant Rao, Coline Devin, Alex~X. Lee, Maria Bauza, Todor Davchev, Yuxiang Zhou, Agrim Gupta, Akhil Raju, Antoine Laurens, Claudio Fantacci, Valentin Dalibard, Martina Zambelli, Murilo Martins, Rugile Pevceviciute, Michiel Blokzijl, Misha Denil, Nathan Batchelor, Thomas Lampe, Emilio Parisotto, Konrad Żołna, Scott Reed, Sergio~Gómez Colmenarejo, Jon Scholz, Abbas Abdolmaleki, Oliver Groth, Jean-Baptiste Regli, Oleg Sushkov, Tom Rothörl, José~Enrique Chen, Yusuf Aytar, Dave Barker, Joy Ortiz, Martin Riedmiller, Jost~Tobias Springenberg, Raia Hadsell, Francesco Nori, and Nicolas Heess.
\newblock Robocat: A self-improving generalist agent for robotic manipulation, 2023.

\bibitem[Bowman et~al.(2022)Bowman, Hyun, Perez, Chen, Pettit, Heiner, Luko{\v{s}}i{\=u}t{\.e}, Askell, Jones, Chen, et~al.]{bowman2022measuring}
Samuel~R Bowman, Jeeyoon Hyun, Ethan Perez, Edwin Chen, Craig Pettit, Scott Heiner, Kamil{\.e} Luko{\v{s}}i{\=u}t{\.e}, Amanda Askell, Andy Jones, Anna Chen, et~al.
\newblock Measuring progress on scalable oversight for large language models.
\newblock \emph{ArXiv preprint}, abs/2211.03540, 2022.
\newblock URL \url{https://arxiv.org/abs/2211.03540}.

\bibitem[Brooks et~al.(2024)Brooks, Peebles, Holmes, DePue, Guo, Jing, Schnurr, Taylor, Luhman, Luhman, Ng, Wang, and Ramesh]{videoworldsimulators2024}
Tim Brooks, Bill Peebles, Connor Holmes, Will DePue, Yufei Guo, Li~Jing, David Schnurr, Joe Taylor, Troy Luhman, Eric Luhman, Clarence Ng, Ricky Wang, and Aditya Ramesh.
\newblock Video generation models as world simulators.
\newblock 2024.
\newblock URL \url{https://openai.com/research/video-generation-models-as-world-simulators}.

\bibitem[Brown et~al.(2020)Brown, Mann, Ryder, Subbiah, Kaplan, Dhariwal, Neelakantan, Shyam, Sastry, Askell, Agarwal, Herbert{-}Voss, Krueger, Henighan, Child, Ramesh, Ziegler, Wu, Winter, Hesse, Chen, Sigler, Litwin, Gray, Chess, Clark, Berner, McCandlish, Radford, Sutskever, and Amodei]{brown2020language}
Tom~B. Brown, Benjamin Mann, Nick Ryder, Melanie Subbiah, Jared Kaplan, Prafulla Dhariwal, Arvind Neelakantan, Pranav Shyam, Girish Sastry, Amanda Askell, Sandhini Agarwal, Ariel Herbert{-}Voss, Gretchen Krueger, Tom Henighan, Rewon Child, Aditya Ramesh, Daniel~M. Ziegler, Jeffrey Wu, Clemens Winter, Christopher Hesse, Mark Chen, Eric Sigler, Mateusz Litwin, Scott Gray, Benjamin Chess, Jack Clark, Christopher Berner, Sam McCandlish, Alec Radford, Ilya Sutskever, and Dario Amodei.
\newblock Language models are few-shot learners.
\newblock In Hugo Larochelle, Marc'Aurelio Ranzato, Raia Hadsell, Maria{-}Florina Balcan, and Hsuan{-}Tien Lin, editors, \emph{Advances in Neural Information Processing Systems 33: Annual Conference on Neural Information Processing Systems 2020, NeurIPS 2020, December 6-12, 2020, virtual}, 2020.
\newblock URL \url{https://proceedings.neurips.cc/paper/2020/hash/1457c0d6bfcb4967418bfb8ac142f64a-Abstract.html}.

\bibitem[Brown-Cohen et~al.(2023)Brown-Cohen, Irving, and Piliouras]{brown2023scalable}
Jonah Brown-Cohen, Geoffrey Irving, and Georgios Piliouras.
\newblock Scalable ai safety via doubly-efficient debate.
\newblock \emph{ArXiv preprint}, abs/2311.14125, 2023.
\newblock URL \url{https://arxiv.org/abs/2311.14125}.

\bibitem[Burns et~al.(2023)Burns, Izmailov, Kirchner, Baker, Gao, Aschenbrenner, Chen, Ecoffet, Joglekar, Leike, Sutskever, and Wu]{burns2023weaktostrong}
Collin Burns, Pavel Izmailov, Jan~Hendrik Kirchner, Bowen Baker, Leo Gao, Leopold Aschenbrenner, Yining Chen, Adrien Ecoffet, Manas Joglekar, Jan Leike, Ilya Sutskever, and Jeff Wu.
\newblock Weak-to-strong generalization: Eliciting strong capabilities with weak supervision, 2023.
\newblock URL \url{https://arxiv.org/abs/2312.09390}.

\bibitem[Carta et~al.(2023)Carta, Romac, Wolf, Lamprier, Sigaud, and Oudeyer]{carta2023grounding}
Thomas Carta, Clément Romac, Thomas Wolf, Sylvain Lamprier, Olivier Sigaud, and Pierre-Yves Oudeyer.
\newblock Grounding large language models in interactive environments with online reinforcement learning, 2023.

\bibitem[Charikar et~al.(2024)Charikar, Pabbaraju, and Shiragur]{charikar2024quantifying}
Moses Charikar, Chirag Pabbaraju, and Kirankumar Shiragur.
\newblock Quantifying the gain in weak-to-strong generalization, 2024.

\bibitem[Chaudhary(2023)]{codealpaca}
Sahil Chaudhary.
\newblock Code alpaca: An instruction-following llama model for code generation.
\newblock \url{https://github.com/sahil280114/codealpaca}, 2023.

\bibitem[Chen et~al.(2024{\natexlab{a}})Chen, Scheurer, Campos, Korbak, Chan, Bowman, Cho, and Perez]{chen2024learning}
Angelica Chen, J{\'e}r{\'e}my Scheurer, Jon~Ander Campos, Tomasz Korbak, Jun~Shern Chan, Samuel~R. Bowman, Kyunghyun Cho, and Ethan Perez.
\newblock Learning from natural language feedback.
\newblock \emph{Transactions on Machine Learning Research}, 2024{\natexlab{a}}.
\newblock ISSN 2835-8856.
\newblock URL \url{https://openreview.net/forum?id=xo3hI5MwvU}.

\bibitem[Chen et~al.(2023{\natexlab{a}})Chen, Shu, Shareghi, Collier, Narasimhan, and Yao]{chen2023fireact}
Baian Chen, Chang Shu, Ehsan Shareghi, Nigel Collier, Karthik Narasimhan, and Shunyu Yao.
\newblock Fireact: Toward language agent fine-tuning, 2023{\natexlab{a}}.

\bibitem[Chen et~al.(2023{\natexlab{b}})Chen, Zhang, Zhang, Yang, Hu, Ma, Yanggong, and Zhao]{chen2023maybe}
Hao Chen, Yiming Zhang, Qi~Zhang, Hantao Yang, Xiaomeng Hu, Xuetao Ma, Yifan Yanggong, and Junbo Zhao.
\newblock Maybe only 0.5\% data is needed: A preliminary exploration of low training data instruction tuning.
\newblock \emph{arXiv preprint arXiv:2305.09246}, 2023{\natexlab{b}}.

\bibitem[Chen et~al.(2023{\natexlab{c}})Chen, Li, Yan, Wang, Gunaratna, Yadav, Tang, Srinivasan, Zhou, Huang, and Jin]{chen2024alpagasus}
Lichang Chen, Shiyang Li, Jun Yan, Hai Wang, Kalpa Gunaratna, Vikas Yadav, Zheng Tang, Vijay Srinivasan, Tianyi Zhou, Heng Huang, and Hongxia Jin.
\newblock Alpagasus: Training a better alpaca with fewer data, 2023{\natexlab{c}}.
\newblock URL \url{https://arxiv.org/abs/2307.08701}.

\bibitem[Chen et~al.(2024{\natexlab{b}})Chen, Guan, Lu, Lin, Han, and Sun]{chen-et-al-2024-reinstruct}
Shu Chen, Xinyan Guan, Yaojie Lu, Hongyu Lin, Xianpei Han, and Le~Sun.
\newblock Reinstruct: Building instruction data from unlabelled corpus.
\newblock In \emph{Proceedings of the 62nd Annual Meeting of the Association for Computational Linguistics}, 2024{\natexlab{b}}.

\bibitem[Chen et~al.(2024{\natexlab{c}})Chen, Song, and Li]{chen2024grath}
Weixin Chen, Dawn Song, and Bo~Li.
\newblock Grath: Gradual self-truthifying for large language models, 2024{\natexlab{c}}.

\bibitem[Chen et~al.(2024{\natexlab{d}})Chen, He, Lin, Han, Wang, Cao, Sun, and Sun]{chen2024spiral}
Xiaoyang Chen, Ben He, Hongyu Lin, Xianpei Han, Tianshu Wang, Boxi Cao, Le~Sun, and Yingfei Sun.
\newblock Spiral of silences: How is large language model killing information retrieval?--a case study on open domain question answering.
\newblock \emph{arXiv preprint arXiv:2404.10496}, 2024{\natexlab{d}}.

\bibitem[Chen et~al.(2023{\natexlab{d}})Chen, Lin, Schärli, and Zhou]{chen2023teaching}
Xinyun Chen, Maxwell Lin, Nathanael Schärli, and Denny Zhou.
\newblock Teaching large language models to self-debug, 2023{\natexlab{d}}.

\bibitem[Chen et~al.(2024{\natexlab{e}})Chen, Wen, Nag, Luo, Yin, Li, Li, and Wang]{chen2024iteralign}
Xiusi Chen, Hongzhi Wen, Sreyashi Nag, Chen Luo, Qingyu Yin, Ruirui Li, Zheng Li, and Wei Wang.
\newblock Iteralign: Iterative constitutional alignment of large language models.
\newblock \emph{ArXiv preprint}, abs/2403.18341, 2024{\natexlab{e}}.
\newblock URL \url{https://arxiv.org/abs/2403.18341}.

\bibitem[Chen et~al.(2023{\natexlab{e}})Chen, Jiang, Huang, Shi, and Qi]{chen2023tegit}
Yongrui Chen, Haiyun Jiang, Xinting Huang, Shuming Shi, and Guilin Qi.
\newblock Tegit: Generating high-quality instruction-tuning data with text-grounded task design, 2023{\natexlab{e}}.

\bibitem[Chen et~al.(2024{\natexlab{f}})Chen, Zhou, Zhao, Wan, Zhang, Zhang, and Wen]{chen2024improving}
Zhipeng Chen, Kun Zhou, Wayne~Xin Zhao, Junchen Wan, Fuzheng Zhang, Di~Zhang, and Ji-Rong Wen.
\newblock Improving large language models via fine-grained reinforcement learning with minimum editing constraint, 2024{\natexlab{f}}.

\bibitem[Chen et~al.(2024{\natexlab{g}})Chen, Deng, Yuan, Ji, and Gu]{chen2024self}
Zixiang Chen, Yihe Deng, Huizhuo Yuan, Kaixuan Ji, and Quanquan Gu.
\newblock Self-play fine-tuning converts weak language models to strong language models.
\newblock \emph{ArXiv preprint}, abs/2401.01335, 2024{\natexlab{g}}.
\newblock URL \url{https://arxiv.org/abs/2401.01335}.

\bibitem[Cheng et~al.(2023)Cheng, Yang, Li, Dai, and Du]{cheng2023adversarial}
Pengyu Cheng, Yifan Yang, Jian Li, Yong Dai, and Nan Du.
\newblock Adversarial preference optimization.
\newblock \emph{arXiv preprint arXiv:2311.08045}, 2023.

\bibitem[Cheng et~al.(2024)Cheng, Hu, Xu, Zhang, Dai, Han, and Du]{cheng2024self}
Pengyu Cheng, Tianhao Hu, Han Xu, Zhisong Zhang, Yong Dai, Lei Han, and Nan Du.
\newblock Self-playing adversarial language game enhances llm reasoning.
\newblock \emph{ArXiv preprint}, abs/2404.10642, 2024.
\newblock URL \url{https://arxiv.org/abs/2404.10642}.

\bibitem[Chiang et~al.(2023)Chiang, Li, Lin, Sheng, Wu, Zhang, Zheng, Zhuang, Zhuang, Gonzalez, Stoica, and Xing]{vicuna2023}
Wei-Lin Chiang, Zhuohan Li, Zi~Lin, Ying Sheng, Zhanghao Wu, Hao Zhang, Lianmin Zheng, Siyuan Zhuang, Yonghao Zhuang, Joseph~E. Gonzalez, Ion Stoica, and Eric~P. Xing.
\newblock Vicuna: An open-source chatbot impressing gpt-4 with 90\%* chatgpt quality, 2023.
\newblock URL \url{https://lmsys.org/blog/2023-03-30-vicuna/}.

\bibitem[Christiano et~al.(2018)Christiano, Shlegeris, and Amodei]{christiano2018supervising}
Paul Christiano, Buck Shlegeris, and Dario Amodei.
\newblock Supervising strong learners by amplifying weak experts.
\newblock \emph{ArXiv preprint}, abs/1810.08575, 2018.
\newblock URL \url{https://arxiv.org/abs/1810.08575}.

\bibitem[Christiano et~al.(2017)Christiano, Leike, Brown, Martic, Legg, and Amodei]{NIPS2017_d5e2c0ad}
Paul~F. Christiano, Jan Leike, Tom~B. Brown, Miljan Martic, Shane Legg, and Dario Amodei.
\newblock Deep reinforcement learning from human preferences.
\newblock In Isabelle Guyon, Ulrike von Luxburg, Samy Bengio, Hanna~M. Wallach, Rob Fergus, S.~V.~N. Vishwanathan, and Roman Garnett, editors, \emph{Advances in Neural Information Processing Systems 30: Annual Conference on Neural Information Processing Systems 2017, December 4-9, 2017, Long Beach, CA, {USA}}, pages 4299--4307, 2017.
\newblock URL \url{https://proceedings.neurips.cc/paper/2017/hash/d5e2c0adad503c91f91df240d0cd4e49-Abstract.html}.

\bibitem[Cobbe et~al.(2021)Cobbe, Kosaraju, Bavarian, Chen, Jun, Kaiser, Plappert, Tworek, Hilton, Nakano, Hesse, and Schulman]{cobbe2021training}
Karl Cobbe, Vineet Kosaraju, Mohammad Bavarian, Mark Chen, Heewoo Jun, Lukasz Kaiser, Matthias Plappert, Jerry Tworek, Jacob Hilton, Reiichiro Nakano, Christopher Hesse, and John Schulman.
\newblock Training verifiers to solve math word problems, 2021.

\bibitem[Cui et~al.(2024)Cui, Yuan, Ding, Yao, Zhu, Ni, Xie, Liu, and Sun]{cui2024ultrafeedback}
Ganqu Cui, Lifan Yuan, Ning Ding, Guanming Yao, Wei Zhu, Yuan Ni, Guotong Xie, Zhiyuan Liu, and Maosong Sun.
\newblock Ultrafeedback: Boosting language models with high-quality feedback, 2024.
\newblock URL \url{https://openreview.net/forum?id=pNkOx3IVWI}.

\bibitem[Dai et~al.(2023)Dai, Sun, Dong, Hao, Ma, Sui, and Wei]{dai-etal-2023-gpt}
Damai Dai, Yutao Sun, Li~Dong, Yaru Hao, Shuming Ma, Zhifang Sui, and Furu Wei.
\newblock Why can {GPT} learn in-context? language models secretly perform gradient descent as meta-optimizers.
\newblock In Anna Rogers, Jordan Boyd-Graber, and Naoaki Okazaki, editors, \emph{Findings of the Association for Computational Linguistics: ACL 2023}, pages 4005--4019, Toronto, Canada, 2023. Association for Computational Linguistics.
\newblock \doi{10.18653/v1/2023.findings-acl.247}.
\newblock URL \url{https://aclanthology.org/2023.findings-acl.247}.

\bibitem[Deng and Raffel(2023)]{deng-raffel-2023-reward}
Haikang Deng and Colin Raffel.
\newblock Reward-augmented decoding: Efficient controlled text generation with a unidirectional reward model.
\newblock In Houda Bouamor, Juan Pino, and Kalika Bali, editors, \emph{Proceedings of the 2023 Conference on Empirical Methods in Natural Language Processing}, pages 11781--11791, Singapore, 2023. Association for Computational Linguistics.
\newblock \doi{10.18653/v1/2023.emnlp-main.721}.
\newblock URL \url{https://aclanthology.org/2023.emnlp-main.721}.

\bibitem[DiGiovanni and Zell(2021)]{digiovanni2021survey}
Anthony DiGiovanni and Ethan~C Zell.
\newblock Survey of self-play in reinforcement learning.
\newblock \emph{ArXiv preprint}, abs/2107.02850, 2021.
\newblock URL \url{https://arxiv.org/abs/2107.02850}.

\bibitem[Ding et~al.(2023)Ding, Chen, Xu, Qin, Hu, Liu, Sun, and Zhou]{ding-etal-2023-enhancing}
Ning Ding, Yulin Chen, Bokai Xu, Yujia Qin, Shengding Hu, Zhiyuan Liu, Maosong Sun, and Bowen Zhou.
\newblock Enhancing chat language models by scaling high-quality instructional conversations.
\newblock In Houda Bouamor, Juan Pino, and Kalika Bali, editors, \emph{Proceedings of the 2023 Conference on Empirical Methods in Natural Language Processing}, pages 3029--3051, Singapore, 2023. Association for Computational Linguistics.
\newblock \doi{10.18653/v1/2023.emnlp-main.183}.
\newblock URL \url{https://aclanthology.org/2023.emnlp-main.183}.

\bibitem[Dong et~al.(2023)Dong, Xiong, Goyal, Zhang, Chow, Pan, Diao, Zhang, SHUM, and Zhang]{dong2023raft}
Hanze Dong, Wei Xiong, Deepanshu Goyal, Yihan Zhang, Winnie Chow, Rui Pan, Shizhe Diao, Jipeng Zhang, KaShun SHUM, and Tong Zhang.
\newblock {RAFT}: Reward ranked finetuning for generative foundation model alignment.
\newblock \emph{Transactions on Machine Learning Research}, 2023.
\newblock ISSN 2835-8856.
\newblock URL \url{https://openreview.net/forum?id=m7p5O7zblY}.

\bibitem[Duan et~al.(2023)Duan, Tang, Yang, Abbasi, and Tam]{duan2023exploring}
Hanyu Duan, Yixuan Tang, Yi~Yang, Ahmed Abbasi, and Kar~Yan Tam.
\newblock Exploring the relationship between in-context learning and instruction tuning, 2023.
\newblock URL \url{https://arxiv.org/abs/2311.10367}.

\bibitem[Dubey et~al.(2024)Dubey, Jauhri, Pandey, Kadian, Al-Dahle, Letman, Mathur, Schelten, Yang, Fan, et~al.]{dubey2024llama}
Abhimanyu Dubey, Abhinav Jauhri, Abhinav Pandey, Abhishek Kadian, Ahmad Al-Dahle, Aiesha Letman, Akhil Mathur, Alan Schelten, Amy Yang, Angela Fan, et~al.
\newblock The llama 3 herd of models.
\newblock \emph{arXiv preprint arXiv:2407.21783}, 2024.

\bibitem[Fernandes et~al.(2023)Fernandes, Deutsch, Finkelstein, Riley, Martins, Neubig, Garg, Clark, Freitag, and Firat]{fernandes-etal-2023-devil}
Patrick Fernandes, Daniel Deutsch, Mara Finkelstein, Parker Riley, Andr{\'e} Martins, Graham Neubig, Ankush Garg, Jonathan Clark, Markus Freitag, and Orhan Firat.
\newblock The devil is in the errors: Leveraging large language models for fine-grained machine translation evaluation.
\newblock In Philipp Koehn, Barry Haddow, Tom Kocmi, and Christof Monz, editors, \emph{Proceedings of the Eighth Conference on Machine Translation}, pages 1066--1083, Singapore, 2023. Association for Computational Linguistics.
\newblock \doi{10.18653/v1/2023.wmt-1.100}.
\newblock URL \url{https://aclanthology.org/2023.wmt-1.100}.

\bibitem[Fernando et~al.(2023)Fernando, Banarse, Michalewski, Osindero, and Rockt{\"a}schel]{fernando2023promptbreeder}
Chrisantha Fernando, Dylan Banarse, Henryk Michalewski, Simon Osindero, and Tim Rockt{\"a}schel.
\newblock Promptbreeder: Self-referential self-improvement via prompt evolution.
\newblock \emph{ArXiv preprint}, abs/2309.16797, 2023.
\newblock URL \url{https://arxiv.org/abs/2309.16797}.

\bibitem[Ferreira et~al.(2023)Ferreira, Lee, and D{\'o}rea]{ferreira2023using}
Rafael~EP Ferreira, Yong~Jae Lee, and Jo{\~a}o~RR D{\'o}rea.
\newblock Using pseudo-labeling to improve performance of deep neural networks for animal identification.
\newblock \emph{Scientific Reports}, 13\penalty0 (1):\penalty0 13875, 2023.
\newblock URL \url{https://doi.org/10.1038/s41598-023-40977-x}.

\bibitem[Forbes et~al.(2020)Forbes, Hwang, Shwartz, Sap, and Choi]{forbes-etal-2020-social}
Maxwell Forbes, Jena~D. Hwang, Vered Shwartz, Maarten Sap, and Yejin Choi.
\newblock Social chemistry 101: Learning to reason about social and moral norms.
\newblock In \emph{Proceedings of the 2020 Conference on Empirical Methods in Natural Language Processing (EMNLP)}, pages 653--670, Online, 2020. Association for Computational Linguistics.
\newblock \doi{10.18653/v1/2020.emnlp-main.48}.
\newblock URL \url{https://aclanthology.org/2020.emnlp-main.48}.

\bibitem[Fränken et~al.(2024)Fränken, Zelikman, Rafailov, Gandhi, Gerstenberg, and Goodman]{franken2024selfsupervised}
Jan-Philipp Fränken, Eric Zelikman, Rafael Rafailov, Kanishk Gandhi, Tobias Gerstenberg, and Noah~D. Goodman.
\newblock Self-supervised alignment with mutual information: Learning to follow principles without preference labels, 2024.
\newblock URL \url{https://arxiv.org/abs/2404.14313}.

\bibitem[Fu et~al.(2023{\natexlab{a}})Fu, Peng, Khot, and Lapata]{fu2023improving}
Yao Fu, Hao Peng, Tushar Khot, and Mirella Lapata.
\newblock Improving language model negotiation with self-play and in-context learning from ai feedback.
\newblock \emph{ArXiv preprint}, abs/2305.10142, 2023{\natexlab{a}}.
\newblock URL \url{https://arxiv.org/abs/2305.10142}.

\bibitem[Fu et~al.(2023{\natexlab{b}})Fu, Peng, Ou, Sabharwal, and Khot]{fu2023specializing}
Yao Fu, Hao Peng, Litu Ou, Ashish Sabharwal, and Tushar Khot.
\newblock Specializing smaller language models towards multi-step reasoning, 2023{\natexlab{b}}.

\bibitem[Ganguli et~al.(2022)Ganguli, Lovitt, Kernion, Askell, Bai, Kadavath, Mann, Perez, Schiefer, Ndousse, et~al.]{ganguli2022red}
Deep Ganguli, Liane Lovitt, Jackson Kernion, Amanda Askell, Yuntao Bai, Saurav Kadavath, Ben Mann, Ethan Perez, Nicholas Schiefer, Kamal Ndousse, et~al.
\newblock Red teaming language models to reduce harms: Methods, scaling behaviors, and lessons learned.
\newblock \emph{ArXiv preprint}, abs/2209.07858, 2022.
\newblock URL \url{https://arxiv.org/abs/2209.07858}.

\bibitem[Gao et~al.(2023)Gao, Schulman, and Hilton]{pmlr-v202-gao23h}
Leo Gao, John Schulman, and Jacob Hilton.
\newblock Scaling laws for reward model overoptimization.
\newblock In Andreas Krause, Emma Brunskill, Kyunghyun Cho, Barbara Engelhardt, Sivan Sabato, and Jonathan Scarlett, editors, \emph{Proceedings of the 40th International Conference on Machine Learning}, volume 202 of \emph{Proceedings of Machine Learning Research}, pages 10835--10866. PMLR, 2023.
\newblock URL \url{https://proceedings.mlr.press/v202/gao23h.html}.

\bibitem[Gao et~al.(2024)Gao, Alon, and Metzler]{gao2024impact}
Yang Gao, Dana Alon, and Donald Metzler.
\newblock Impact of preference noise on the alignment performance of generative language models, 2024.

\bibitem[Garrabrant and Demski(2018)]{garrabrant2018embedded}
Scott Garrabrant and Abram Demski.
\newblock Embedded agency, 2018.
\newblock URL \url{https://www.alignmentforum.org/s/Rm6oQRJJmhGCcLvxh/p/i3BTagvt3HbPMx6PN}.

\bibitem[Gekhman et~al.(2024)Gekhman, Yona, Aharoni, Eyal, Feder, Reichart, and Herzig]{gekhman2024does}
Zorik Gekhman, Gal Yona, Roee Aharoni, Matan Eyal, Amir Feder, Roi Reichart, and Jonathan Herzig.
\newblock Does fine-tuning llms on new knowledge encourage hallucinations?, 2024.
\newblock URL \url{https://arxiv.org/abs/2405.05904}.

\bibitem[Gim et~al.(2023)Gim, Chen, Lee, Sarda, Khandelwal, and Zhong]{gim2023prompt}
In~Gim, Guojun Chen, Seung-seob Lee, Nikhil Sarda, Anurag Khandelwal, and Lin Zhong.
\newblock Prompt cache: Modular attention reuse for low-latency inference.
\newblock \emph{ArXiv preprint}, abs/2311.04934, 2023.
\newblock URL \url{https://arxiv.org/abs/2311.04934}.

\bibitem[Goodfellow et~al.(2014)Goodfellow, Pouget{-}Abadie, Mirza, Xu, Warde{-}Farley, Ozair, Courville, and Bengio]{goodfellow2014generative}
Ian~J. Goodfellow, Jean Pouget{-}Abadie, Mehdi Mirza, Bing Xu, David Warde{-}Farley, Sherjil Ozair, Aaron~C. Courville, and Yoshua Bengio.
\newblock Generative adversarial nets.
\newblock In Zoubin Ghahramani, Max Welling, Corinna Cortes, Neil~D. Lawrence, and Kilian~Q. Weinberger, editors, \emph{Advances in Neural Information Processing Systems 27: Annual Conference on Neural Information Processing Systems 2014, December 8-13 2014, Montreal, Quebec, Canada}, pages 2672--2680, 2014.
\newblock URL \url{https://proceedings.neurips.cc/paper/2014/hash/5ca3e9b122f61f8f06494c97b1afccf3-Abstract.html}.

\bibitem[Gou et~al.(2024)Gou, Shao, Gong, Shen, Yang, Duan, and Chen]{gou2024critic}
Zhibin Gou, Zhihong Shao, Yeyun Gong, Yelong Shen, Yujiu Yang, Nan Duan, and Weizhu Chen.
\newblock Critic: Large language models can self-correct with tool-interactive critiquing, 2024.
\newblock URL \url{https://arxiv.org/abs/2305.11738}.

\bibitem[Grandvalet and Bengio(2004)]{grandvalet2004semi}
Yves Grandvalet and Yoshua Bengio.
\newblock Semi-supervised learning by entropy minimization.
\newblock In \emph{Advances in Neural Information Processing Systems 17 [Neural Information Processing Systems, {NIPS} 2004, December 13-18, 2004, Vancouver, British Columbia, Canada]}, pages 529--536, 2004.
\newblock URL \url{https://proceedings.neurips.cc/paper/2004/hash/96f2b50b5d3613adf9c27049b2a888c7-Abstract.html}.

\bibitem[Gu et~al.(2024)Gu, Knoll, and Jin]{gu2024teaching}
Shangding Gu, Alois Knoll, and Ming Jin.
\newblock Teaching {LLM}s to teach themselves better instructions via reinforcement learning, 2024.
\newblock URL \url{https://openreview.net/forum?id=wlRp8IdLkN}.

\bibitem[Gudibande et~al.(2023)Gudibande, Wallace, Snell, Geng, Liu, Abbeel, Levine, and Song]{gudibande2023false}
Arnav Gudibande, Eric Wallace, Charlie Snell, Xinyang Geng, Hao Liu, Pieter Abbeel, Sergey Levine, and Dawn Song.
\newblock The false promise of imitating proprietary llms, 2023.
\newblock URL \url{https://arxiv.org/abs/2305.15717}.

\bibitem[Guo et~al.(2024{\natexlab{a}})Guo, Zhang, Liu, Liu, Khalman, Llinares, Rame, Mesnard, Zhao, Piot, et~al.]{guo2024direct}
Shangmin Guo, Biao Zhang, Tianlin Liu, Tianqi Liu, Misha Khalman, Felipe Llinares, Alexandre Rame, Thomas Mesnard, Yao Zhao, Bilal Piot, et~al.
\newblock Direct language model alignment from online ai feedback.
\newblock \emph{ArXiv preprint}, abs/2402.04792, 2024{\natexlab{a}}.
\newblock URL \url{https://arxiv.org/abs/2402.04792}.

\bibitem[Guo et~al.(2024{\natexlab{b}})Guo, Cui, Yuan, Ding, Wang, Chen, Sun, Xie, Zhou, Lin, Liu, and Sun]{guo2024controllable}
Yiju Guo, Ganqu Cui, Lifan Yuan, Ning Ding, Jiexin Wang, Huimin Chen, Bowen Sun, Ruobing Xie, Jie Zhou, Yankai Lin, Zhiyuan Liu, and Maosong Sun.
\newblock Controllable preference optimization: Toward controllable multi-objective alignment, 2024{\natexlab{b}}.

\bibitem[Hase et~al.(2024)Hase, Bansal, Clark, and Wiegreffe]{hase2024unreasonable}
Peter Hase, Mohit Bansal, Peter Clark, and Sarah Wiegreffe.
\newblock The unreasonable effectiveness of easy training data for hard tasks, 2024.
\newblock URL \url{https://arxiv.org/abs/2401.06751}.

\bibitem[Havrilla et~al.(2024)Havrilla, Raparthy, Nalmpantis, Dwivedi-Yu, Zhuravinskyi, Hambro, and Railneau]{havrilla2024glore}
Alex Havrilla, Sharath Raparthy, Christoforus Nalmpantis, Jane Dwivedi-Yu, Maksym Zhuravinskyi, Eric Hambro, and Roberta Railneau.
\newblock Glore: When, where, and how to improve llm reasoning via global and local refinements, 2024.

\bibitem[Hazra and Anjaria(2022)]{hazra2022applications}
Tanmoy Hazra and Kushal Anjaria.
\newblock Applications of game theory in deep learning: a survey.
\newblock \emph{Multimedia Tools and Applications}, 81\penalty0 (6):\penalty0 8963--8994, 2022.

\bibitem[He et~al.(2023)He, Cui, Chen, Hu, and Zhu]{he2023investigating}
Guande He, Peng Cui, Jianfei Chen, Wenbo Hu, and Jun Zhu.
\newblock Investigating uncertainty calibration of aligned language models under the multiple-choice setting.
\newblock \emph{ArXiv preprint}, abs/2310.11732, 2023.
\newblock URL \url{https://arxiv.org/abs/2310.11732}.

\bibitem[He et~al.(2020)He, Gu, Shen, and Ranzato]{He2020Revisiting}
Junxian He, Jiatao Gu, Jiajun Shen, and Marc'Aurelio Ranzato.
\newblock Revisiting self-training for neural sequence generation.
\newblock In \emph{8th International Conference on Learning Representations, {ICLR} 2020, Addis Ababa, Ethiopia, April 26-30, 2020}. OpenReview.net, 2020.
\newblock URL \url{https://openreview.net/forum?id=SJgdnAVKDH}.

\bibitem[Ho et~al.(2023)Ho, Schmid, and Yun]{ho-etal-2023-large}
Namgyu Ho, Laura Schmid, and Se-Young Yun.
\newblock Large language models are reasoning teachers.
\newblock In Anna Rogers, Jordan Boyd-Graber, and Naoaki Okazaki, editors, \emph{Proceedings of the 61st Annual Meeting of the Association for Computational Linguistics (Volume 1: Long Papers)}, pages 14852--14882, Toronto, Canada, 2023. Association for Computational Linguistics.
\newblock \doi{10.18653/v1/2023.acl-long.830}.
\newblock URL \url{https://aclanthology.org/2023.acl-long.830}.

\bibitem[Hoare(1961)]{hoare1961algorithm}
Charles Antony~Richard Hoare.
\newblock Algorithm 64: quicksort.
\newblock \emph{Communications of the ACM}, 4\penalty0 (7):\penalty0 321, 1961.

\bibitem[Hong et~al.(2023)Hong, Tu, Chen, Gao, Zhang, and Yan]{hong2023cyclealign}
Jixiang Hong, Quan Tu, Changyu Chen, Xing Gao, Ji~Zhang, and Rui Yan.
\newblock Cyclealign: Iterative distillation from black-box llm to white-box models for better human alignment, 2023.

\bibitem[Hong et~al.(2024)Hong, Zhang, Pan, Yu, and Zhang]{hong2024abstractionofthought}
Ruixin Hong, Hongming Zhang, Xiaoman Pan, Dong Yu, and Changshui Zhang.
\newblock Abstraction-of-thought makes language models better reasoners, 2024.

\bibitem[Honovich et~al.(2023)Honovich, Scialom, Levy, and Schick]{honovich-etal-2023-unnatural}
Or~Honovich, Thomas Scialom, Omer Levy, and Timo Schick.
\newblock Unnatural instructions: Tuning language models with (almost) no human labor.
\newblock In Anna Rogers, Jordan Boyd-Graber, and Naoaki Okazaki, editors, \emph{Proceedings of the 61st Annual Meeting of the Association for Computational Linguistics (Volume 1: Long Papers)}, pages 14409--14428, Toronto, Canada, 2023. Association for Computational Linguistics.
\newblock \doi{10.18653/v1/2023.acl-long.806}.
\newblock URL \url{https://aclanthology.org/2023.acl-long.806}.

\bibitem[Hosseini et~al.(2024)Hosseini, Yuan, Malkin, Courville, Sordoni, and Agarwal]{hosseini2024vstar}
Arian Hosseini, Xingdi Yuan, Nikolay Malkin, Aaron Courville, Alessandro Sordoni, and Rishabh Agarwal.
\newblock V-star: Training verifiers for self-taught reasoners, 2024.

\bibitem[Hsieh et~al.(2023)Hsieh, Li, Yeh, Nakhost, Fujii, Ratner, Krishna, Lee, and Pfister]{hsieh-etal-2023-distilling}
Cheng-Yu Hsieh, Chun-Liang Li, Chih-kuan Yeh, Hootan Nakhost, Yasuhisa Fujii, Alex Ratner, Ranjay Krishna, Chen-Yu Lee, and Tomas Pfister.
\newblock Distilling step-by-step! outperforming larger language models with less training data and smaller model sizes.
\newblock In Anna Rogers, Jordan Boyd-Graber, and Naoaki Okazaki, editors, \emph{Findings of the Association for Computational Linguistics: ACL 2023}, pages 8003--8017, Toronto, Canada, 2023. Association for Computational Linguistics.
\newblock \doi{10.18653/v1/2023.findings-acl.507}.
\newblock URL \url{https://aclanthology.org/2023.findings-acl.507}.

\bibitem[Huang et~al.(2023{\natexlab{a}})Huang, Gu, Hou, Wu, Wang, Yu, and Han]{huang2023large}
Jiaxin Huang, Shixiang~Shane Gu, Le~Hou, Yuexin Wu, Xuezhi Wang, Hongkun Yu, and Jiawei Han.
\newblock Large language models can self-improve.
\newblock In \emph{The 2023 Conference on Empirical Methods in Natural Language Processing}, 2023{\natexlab{a}}.
\newblock URL \url{https://openreview.net/forum?id=uuUQraD4XX}.

\bibitem[Huang et~al.(2023{\natexlab{b}})Huang, Chen, Mishra, Zheng, Yu, Song, and Zhou]{huang2024large}
Jie Huang, Xinyun Chen, Swaroop Mishra, Huaixiu~Steven Zheng, Adams~Wei Yu, Xinying Song, and Denny Zhou.
\newblock Large language models cannot self-correct reasoning yet, 2023{\natexlab{b}}.
\newblock URL \url{https://arxiv.org/abs/2310.01798}.

\bibitem[Huang et~al.(2024)Huang, Siddarth, Lovitt, Liao, Durmus, Tamkin, and Ganguli]{Huang2024collective}
Saffron Huang, Divya Siddarth, Liane Lovitt, Thomas~I. Liao, Esin Durmus, Alex Tamkin, and Deep Ganguli.
\newblock Collective constitutional ai: Aligning a language model with public input.
\newblock In \emph{The 2024 ACM Conference on Fairness, Accountability, and Transparency}, FAccT ’24. ACM, June 2024.
\newblock \doi{10.1145/3630106.3658979}.
\newblock URL \url{http://dx.doi.org/10.1145/3630106.3658979}.

\bibitem[Hubinger et~al.(2024)Hubinger, Denison, Mu, Lambert, Tong, MacDiarmid, Lanham, Ziegler, Maxwell, Cheng, et~al.]{hubinger2024sleeper}
Evan Hubinger, Carson Denison, Jesse Mu, Mike Lambert, Meg Tong, Monte MacDiarmid, Tamera Lanham, Daniel~M Ziegler, Tim Maxwell, Newton Cheng, et~al.
\newblock Sleeper agents: Training deceptive llms that persist through safety training.
\newblock \emph{arXiv preprint arXiv:2401.05566}, 2024.

\bibitem[Irving et~al.(2018)Irving, Christiano, and Amodei]{irving2018ai}
Geoffrey Irving, Paul Christiano, and Dario Amodei.
\newblock Ai safety via debate.
\newblock \emph{ArXiv preprint}, abs/1805.00899, 2018.
\newblock URL \url{https://arxiv.org/abs/1805.00899}.

\bibitem[Jacob et~al.(2024)Jacob, Shen, Farina, and Andreas]{jacob2024the}
Athul~Paul Jacob, Yikang Shen, Gabriele Farina, and Jacob Andreas.
\newblock The consensus game: Language model generation via equilibrium search.
\newblock In \emph{The Twelfth International Conference on Learning Representations}, 2024.
\newblock URL \url{https://openreview.net/forum?id=n9xeGcI4Yg}.

\bibitem[Ji et~al.(2023)Ji, Qiu, Chen, Zhang, Lou, Wang, Duan, He, Zhou, Zhang, Zeng, Ng, Dai, Pan, O'Gara, Lei, Xu, Tse, Fu, McAleer, Yang, Wang, Zhu, Guo, and Gao]{jiAIAlignmentComprehensive2024}
Jiaming Ji, Tianyi Qiu, Boyuan Chen, Borong Zhang, Hantao Lou, Kaile Wang, Yawen Duan, Zhonghao He, Jiayi Zhou, Zhaowei Zhang, Fanzhi Zeng, Kwan~Yee Ng, Juntao Dai, Xuehai Pan, Aidan O'Gara, Yingshan Lei, Hua Xu, Brian Tse, Jie Fu, Stephen McAleer, Yaodong Yang, Yizhou Wang, Song-Chun Zhu, Yike Guo, and Wen Gao.
\newblock {{AI Alignment}}: {{A Comprehensive Survey}}, 2023.
\newblock URL \url{https://arxiv.org/abs/2310.19852}.

\bibitem[Ji et~al.(2024)Ji, Chen, Lou, Hong, Zhang, Pan, Dai, and Yang]{ji2024aligner}
Jiaming Ji, Boyuan Chen, Hantao Lou, Donghai Hong, Borong Zhang, Xuehai Pan, Juntao Dai, and Yaodong Yang.
\newblock Aligner: Achieving efficient alignment through weak-to-strong correction, 2024.

\bibitem[Jiang et~al.(2023)Jiang, Wang, and Wang]{jiang2023selfevolve}
Shuyang Jiang, Yuhao Wang, and Yu~Wang.
\newblock Selfevolve: A code evolution framework via large language models, 2023.

\bibitem[Kadavath et~al.(2022)Kadavath, Conerly, Askell, Henighan, Drain, Perez, Schiefer, Hatfield-Dodds, DasSarma, Tran-Johnson, et~al.]{kadavath2022language}
Saurav Kadavath, Tom Conerly, Amanda Askell, Tom Henighan, Dawn Drain, Ethan Perez, Nicholas Schiefer, Zac Hatfield-Dodds, Nova DasSarma, Eli Tran-Johnson, et~al.
\newblock Language models (mostly) know what they know.
\newblock \emph{ArXiv preprint}, abs/2207.05221, 2022.
\newblock URL \url{https://arxiv.org/abs/2207.05221}.

\bibitem[Khalifa et~al.(2023)Khalifa, Logeswaran, Lee, Lee, and Wang]{khalifa-etal-2023-grace}
Muhammad Khalifa, Lajanugen Logeswaran, Moontae Lee, Honglak Lee, and Lu~Wang.
\newblock {GRACE}: Discriminator-guided chain-of-thought reasoning.
\newblock In Houda Bouamor, Juan Pino, and Kalika Bali, editors, \emph{Findings of the Association for Computational Linguistics: EMNLP 2023}, pages 15299--15328, Singapore, 2023. Association for Computational Linguistics.
\newblock \doi{10.18653/v1/2023.findings-emnlp.1022}.
\newblock URL \url{https://aclanthology.org/2023.findings-emnlp.1022}.

\bibitem[Khan et~al.(2024)Khan, Hughes, Valentine, Ruis, Sachan, Radhakrishnan, Grefenstette, Bowman, Rockt{\"a}schel, and Perez]{khan2024debating}
Akbir Khan, John Hughes, Dan Valentine, Laura Ruis, Kshitij Sachan, Ansh Radhakrishnan, Edward Grefenstette, Samuel~R Bowman, Tim Rockt{\"a}schel, and Ethan Perez.
\newblock Debating with more persuasive llms leads to more truthful answers.
\newblock \emph{ArXiv preprint}, abs/2402.06782, 2024.
\newblock URL \url{https://arxiv.org/abs/2402.06782}.

\bibitem[Khot et~al.(2023)Khot, Trivedi, Finlayson, Fu, Richardson, Clark, and Sabharwal]{khot2023decomposed}
Tushar Khot, Harsh Trivedi, Matthew Finlayson, Yao Fu, Kyle Richardson, Peter Clark, and Ashish Sabharwal.
\newblock Decomposed prompting: A modular approach for solving complex tasks.
\newblock In \emph{The Eleventh International Conference on Learning Representations}, 2023.
\newblock URL \url{https://openreview.net/forum?id=_nGgzQjzaRy}.

\bibitem[Kim et~al.(2023)Kim, Bae, Shin, Kang, Kwak, Yoo, and Seo]{kim-etal-2023-aligning}
Sungdong Kim, Sanghwan Bae, Jamin Shin, Soyoung Kang, Donghyun Kwak, Kang Yoo, and Minjoon Seo.
\newblock Aligning large language models through synthetic feedback.
\newblock In Houda Bouamor, Juan Pino, and Kalika Bali, editors, \emph{Proceedings of the 2023 Conference on Empirical Methods in Natural Language Processing}, pages 13677--13700, Singapore, 2023. Association for Computational Linguistics.
\newblock \doi{10.18653/v1/2023.emnlp-main.844}.
\newblock URL \url{https://aclanthology.org/2023.emnlp-main.844}.

\bibitem[Kirchner et~al.(2024)Kirchner, Chen, Edwards, Leike, McAleese, and Burda]{kirchner2024prover}
Jan~Hendrik Kirchner, Yining Chen, Harri Edwards, Jan Leike, Nat McAleese, and Yuri Burda.
\newblock Prover-verifier games improve legibility of llm outputs.
\newblock \emph{arXiv preprint arXiv:2407.13692}, 2024.

\bibitem[Klingefjord et~al.(2024)Klingefjord, Lowe, and Edelman]{klingefjord2024human}
Oliver Klingefjord, Ryan Lowe, and Joe Edelman.
\newblock What are human values, and how do we align ai to them?, 2024.
\newblock URL \url{https://arxiv.org/pdf/2404.10636}.

\bibitem[Kojima et~al.(2022)Kojima, Gu, Reid, Matsuo, and Iwasawa]{kojima2022large}
Takeshi Kojima, Shixiang~Shane Gu, Machel Reid, Yutaka Matsuo, and Yusuke Iwasawa.
\newblock Large language models are zero-shot reasoners.
\newblock \emph{Advances in neural information processing systems}, 35:\penalty0 22199--22213, 2022.

\bibitem[Koutcheme et~al.(2024)Koutcheme, Dainese, Sarsa, Hellas, Leinonen, and Denny]{koutcheme2024open}
Charles Koutcheme, Nicola Dainese, Sami Sarsa, Arto Hellas, Juho Leinonen, and Paul Denny.
\newblock Open source language models can provide feedback: Evaluating llms' ability to help students using gpt-4-as-a-judge, 2024.

\bibitem[Köksal et~al.(2024)Köksal, Schick, Korhonen, and Schütze]{longform}
Abdullatif Köksal, Timo Schick, Anna Korhonen, and Hinrich Schütze.
\newblock Longform: Effective instruction tuning with reverse instructions, 2024.

\bibitem[Lan et~al.(2024)Lan, Zhang, Xu, Huang, Lin, Chen, and Mao]{lan2024criticbench}
Tian Lan, Wenwei Zhang, Chen Xu, Heyan Huang, Dahua Lin, Kai Chen, and Xian-ling Mao.
\newblock Criticbench: Evaluating large language models as critic, 2024.
\newblock URL \url{https://arxiv.org/abs/2402.13764}.

\bibitem[Lang et~al.(2024)Lang, Sontag, and Vijayaraghavan]{lang2024theoretical}
Hunter Lang, David Sontag, and Aravindan Vijayaraghavan.
\newblock Theoretical analysis of weak-to-strong generalization, 2024.

\bibitem[Lango and Dusek(2023)]{lango-dusek-2023-critic}
Mateusz Lango and Ondrej Dusek.
\newblock Critic-driven decoding for mitigating hallucinations in data-to-text generation.
\newblock In Houda Bouamor, Juan Pino, and Kalika Bali, editors, \emph{Proceedings of the 2023 Conference on Empirical Methods in Natural Language Processing}, pages 2853--2862, Singapore, 2023. Association for Computational Linguistics.
\newblock \doi{10.18653/v1/2023.emnlp-main.172}.
\newblock URL \url{https://aclanthology.org/2023.emnlp-main.172}.

\bibitem[Le et~al.(2022)Le, Wang, Gotmare, Savarese, and Hoi]{le2022coderl}
Hung Le, Yue Wang, Akhilesh~Deepak Gotmare, Silvio Savarese, and Steven C.~H. Hoi.
\newblock Coderl: Mastering code generation through pretrained models and deep reinforcement learning, 2022.

\bibitem[Lee et~al.(2013)]{lee2013pseudo}
Dong-Hyun Lee et~al.
\newblock Pseudo-label: The simple and efficient semi-supervised learning method for deep neural networks.
\newblock In \emph{Workshop on challenges in representation learning, ICML}, volume~3, page 896. Atlanta, 2013.

\bibitem[Lee and Anderson(2001)]{lee2001does}
Frank~J Lee and John~R Anderson.
\newblock Does learning a complex task have to be complex?: A study in learning decomposition.
\newblock \emph{Cognitive psychology}, 42\penalty0 (3):\penalty0 267--316, 2001.

\bibitem[Lee et~al.(2023)Lee, Phatale, Mansoor, Mesnard, Ferret, Lu, Bishop, Hall, Carbune, Rastogi, and Prakash]{lee2023rlaif}
Harrison Lee, Samrat Phatale, Hassan Mansoor, Thomas Mesnard, Johan Ferret, Kellie Lu, Colton Bishop, Ethan Hall, Victor Carbune, Abhinav Rastogi, and Sushant Prakash.
\newblock Rlaif: Scaling reinforcement learning from human feedback with ai feedback, 2023.

\bibitem[Lee et~al.(2024)Lee, Wattanawong, Kim, Mangalam, Shen, Anumanchipali, Mahoney, Keutzer, and Gholami]{lee2024llm2llm}
Nicholas Lee, Thanakul Wattanawong, Sehoon Kim, Karttikeya Mangalam, Sheng Shen, Gopala Anumanchipali, Michael~W. Mahoney, Kurt Keutzer, and Amir Gholami.
\newblock Llm2llm: Boosting llms with novel iterative data enhancement, 2024.

\bibitem[Li et~al.(2023{\natexlab{a}})Li, Yuan, Yuan, Dong, Lu, Wu, Tan, Wang, and Zhou]{li2023query}
Chengpeng Li, Zheng Yuan, Hongyi Yuan, Guanting Dong, Keming Lu, Jiancan Wu, Chuanqi Tan, Xiang Wang, and Chang Zhou.
\newblock Query and response augmentation cannot help out-of-domain math reasoning generalization, 2023{\natexlab{a}}.

\bibitem[Li et~al.(2024{\natexlab{a}})Li, Zhang, Dong, Deik, Tang, and Liu]{li2024aligning}
Dexun Li, Cong Zhang, Kuicai Dong, Derrick Goh~Xin Deik, Ruiming Tang, and Yong Liu.
\newblock Aligning crowd feedback via distributional preference reward modeling, 2024{\natexlab{a}}.

\bibitem[Li et~al.(2023{\natexlab{b}})Li, Chen, Chen, He, and Zhou]{li2023reflectiontuning}
Ming Li, Lichang Chen, Jiuhai Chen, Shwai He, and Tianyi Zhou.
\newblock Reflection-tuning: Recycling data for better instruction-tuning.
\newblock In \emph{NeurIPS 2023 Workshop on Instruction Tuning and Instruction Following}, 2023{\natexlab{b}}.
\newblock URL \url{https://openreview.net/forum?id=xaqoZZqkPU}.

\bibitem[Li et~al.(2024{\natexlab{b}})Li, Chen, Chen, He, Gu, and Zhou]{li2024selective}
Ming Li, Lichang Chen, Jiuhai Chen, Shwai He, Jiuxiang Gu, and Tianyi Zhou.
\newblock Selective reflection-tuning: Student-selected data recycling for llm instruction-tuning, 2024{\natexlab{b}}.

\bibitem[Li et~al.(2024{\natexlab{c}})Li, Zhang, He, Li, Zhao, Wang, Cheng, and Zhou]{li2024superfiltering}
Ming Li, Yong Zhang, Shwai He, Zhitao Li, Hongyu Zhao, Jianzong Wang, Ning Cheng, and Tianyi Zhou.
\newblock Superfiltering: Weak-to-strong data filtering for fast instruction-tuning, 2024{\natexlab{c}}.

\bibitem[Li et~al.(2024{\natexlab{d}})Li, Zhang, Li, Chen, Chen, Cheng, Wang, Zhou, and Xiao]{li2024quantity}
Ming Li, Yong Zhang, Zhitao Li, Jiuhai Chen, Lichang Chen, Ning Cheng, Jianzong Wang, Tianyi Zhou, and Jing Xiao.
\newblock From quantity to quality: Boosting llm performance with self-guided data selection for instruction tuning, 2024{\natexlab{d}}.
\newblock URL \url{https://arxiv.org/abs/2308.12032}.

\bibitem[Li et~al.(2022)Li, Chen, Shen, Chen, Zhang, Li, Wang, Qian, Peng, Mao, Chen, and Yan]{li2022explanations}
Shiyang Li, Jianshu Chen, Yelong Shen, Zhiyu Chen, Xinlu Zhang, Zekun Li, Hong Wang, Jing Qian, Baolin Peng, Yi~Mao, Wenhu Chen, and Xifeng Yan.
\newblock Explanations from large language models make small reasoners better, 2022.

\bibitem[Li et~al.(2024{\natexlab{e}})Li, Zhang, Do, Yue, and Chen]{li2024long}
Tianle Li, Ge~Zhang, Quy~Duc Do, Xiang Yue, and Wenhu Chen.
\newblock Long-context llms struggle with long in-context learning.
\newblock \emph{arXiv preprint arXiv:2404.02060}, 2024{\natexlab{e}}.

\bibitem[Li et~al.(2024{\natexlab{f}})Li, Yu, Zhou, Schick, Levy, Zettlemoyer, Weston, and Lewis]{li2024selfalignment}
Xian Li, Ping Yu, Chunting Zhou, Timo Schick, Omer Levy, Luke Zettlemoyer, Jason~E Weston, and Mike Lewis.
\newblock Self-alignment with instruction backtranslation.
\newblock In \emph{The Twelfth International Conference on Learning Representations}, volume abs/2308.06259, 2024{\natexlab{f}}.
\newblock URL \url{https://openreview.net/forum?id=1oijHJBRsT}.

\bibitem[Li et~al.(2024{\natexlab{g}})Li, Shrivastava, Li, Hashimoto, and Liang]{li2024benchmarking}
Xiang~Lisa Li, Vaishnavi Shrivastava, Siyan Li, Tatsunori Hashimoto, and Percy Liang.
\newblock Benchmarking and improving generator-validator consistency of language models.
\newblock In \emph{The Twelfth International Conference on Learning Representations}, volume abs/2310.01846, 2024{\natexlab{g}}.
\newblock URL \url{https://openreview.net/forum?id=phBS6YpTzC}.

\bibitem[Li and Qiu(2023)]{li-qiu-2023-mot}
Xiaonan Li and Xipeng Qiu.
\newblock {M}o{T}: Memory-of-thought enables {C}hat{GPT} to self-improve.
\newblock In Houda Bouamor, Juan Pino, and Kalika Bali, editors, \emph{Proceedings of the 2023 Conference on Empirical Methods in Natural Language Processing}, pages 6354--6374, Singapore, December 2023. Association for Computational Linguistics.
\newblock \doi{10.18653/v1/2023.emnlp-main.392}.
\newblock URL \url{https://aclanthology.org/2023.emnlp-main.392}.

\bibitem[Li et~al.(2023{\natexlab{c}})Li, Lin, Zhang, Fu, Chen, Lou, and Chen]{li-etal-2023-making}
Yifei Li, Zeqi Lin, Shizhuo Zhang, Qiang Fu, Bei Chen, Jian-Guang Lou, and Weizhu Chen.
\newblock Making language models better reasoners with step-aware verifier.
\newblock In Anna Rogers, Jordan Boyd-Graber, and Naoaki Okazaki, editors, \emph{Proceedings of the 61st Annual Meeting of the Association for Computational Linguistics (Volume 1: Long Papers)}, pages 5315--5333, Toronto, Canada, 2023{\natexlab{c}}. Association for Computational Linguistics.
\newblock \doi{10.18653/v1/2023.acl-long.291}.
\newblock URL \url{https://aclanthology.org/2023.acl-long.291}.

\bibitem[Li et~al.(2023{\natexlab{d}})Li, Bubeck, Eldan, Giorno, Gunasekar, and Lee]{li2023textbooks}
Yuanzhi Li, Sébastien Bubeck, Ronen Eldan, Allie~Del Giorno, Suriya Gunasekar, and Yin~Tat Lee.
\newblock Textbooks are all you need ii: phi-1.5 technical report, 2023{\natexlab{d}}.

\bibitem[Li et~al.(2024{\natexlab{h}})Li, Wei, Zhao, Zhang, and Zhang]{li2024rain}
Yuhui Li, Fangyun Wei, Jinjing Zhao, Chao Zhang, and Hongyang Zhang.
\newblock {RAIN}: Your language models can align themselves without finetuning.
\newblock In \emph{The Twelfth International Conference on Learning Representations}, volume abs/2309.07124, 2024{\natexlab{h}}.
\newblock URL \url{https://openreview.net/forum?id=pETSfWMUzy}.

\bibitem[Li et~al.(2023{\natexlab{e}})Li, Wang, Ma, Wu, Wang, Gao, and Liu]{li2023split}
Zongjie Li, Chaozheng Wang, Pingchuan Ma, Daoyuan Wu, Shuai Wang, Cuiyun Gao, and Yang Liu.
\newblock Split and merge: Aligning position biases in large language model based evaluators, 2023{\natexlab{e}}.
\newblock URL \url{https://arxiv.org/abs/2310.01432}.

\bibitem[Liao et~al.(2024)Liao, Luo, Li, Wu, and Fan]{liao2024mario}
Minpeng Liao, Wei Luo, Chengxi Li, Jing Wu, and Kai Fan.
\newblock Mario: Math reasoning with code interpreter output -- a reproducible pipeline, 2024.

\bibitem[Lightman et~al.(2023)Lightman, Kosaraju, Burda, Edwards, Baker, Lee, Leike, Schulman, Sutskever, and Cobbe]{lightman2023lets}
Hunter Lightman, Vineet Kosaraju, Yura Burda, Harri Edwards, Bowen Baker, Teddy Lee, Jan Leike, John Schulman, Ilya Sutskever, and Karl Cobbe.
\newblock Let's verify step by step, 2023.

\bibitem[Lin et~al.(2024{\natexlab{a}})Lin, Ravichander, Lu, Dziri, Sclar, Chandu, Bhagavatula, and Choi]{lin2024the}
Bill~Yuchen Lin, Abhilasha Ravichander, Ximing Lu, Nouha Dziri, Melanie Sclar, Khyathi Chandu, Chandra Bhagavatula, and Yejin Choi.
\newblock The unlocking spell on base {LLM}s: Rethinking alignment via in-context learning.
\newblock In \emph{The Twelfth International Conference on Learning Representations}, volume abs/2312.01552, 2024{\natexlab{a}}.
\newblock URL \url{https://openreview.net/forum?id=wxJ0eXwwda}.

\bibitem[Lin et~al.(2024{\natexlab{b}})Lin, Gou, Liang, Luo, Liu, and Yang]{lin2024criticbench}
Zicheng Lin, Zhibin Gou, Tian Liang, Ruilin Luo, Haowei Liu, and Yujiu Yang.
\newblock Criticbench: Benchmarking llms for critique-correct reasoning.
\newblock \emph{ArXiv preprint}, abs/2402.14809, 2024{\natexlab{b}}.
\newblock URL \url{https://arxiv.org/abs/2402.14809}.

\bibitem[Liu et~al.(2024{\natexlab{a}})Liu, Bai, Lu, Kong, Wang, Shan, Cao, and Wen]{liu2024direct}
Aiwei Liu, Haoping Bai, Zhiyun Lu, Xiang Kong, Simon Wang, Jiulong Shan, Meng Cao, and Lijie Wen.
\newblock Direct large language model alignment through self-rewarding contrastive prompt distillation, 2024{\natexlab{a}}.
\newblock URL \url{https://arxiv.org/abs/2402.11907}.

\bibitem[Liu et~al.(2023{\natexlab{a}})Liu, Yang, Jia, Zhang, Zhou, Dai, Yang, and Vosoughi]{liu2023training}
Ruibo Liu, Ruixin Yang, Chenyan Jia, Ge~Zhang, Denny Zhou, Andrew~M. Dai, Diyi Yang, and Soroush Vosoughi.
\newblock Training socially aligned language models on simulated social interactions, 2023{\natexlab{a}}.

\bibitem[Liu et~al.(2024{\natexlab{b}})Liu, Guo, Bianco, Calandriello, Berthet, Llinares, Hoffmann, Dixon, Valko, and Blondel]{liu2024decodingtime}
Tianlin Liu, Shangmin Guo, Leonardo Bianco, Daniele Calandriello, Quentin Berthet, Felipe Llinares, Jessica Hoffmann, Lucas Dixon, Michal Valko, and Mathieu Blondel.
\newblock Decoding-time realignment of language models, 2024{\natexlab{b}}.

\bibitem[Liu et~al.(2024{\natexlab{c}})Liu, Zeng, He, Jiang, and He]{liu2024what}
Wei Liu, Weihao Zeng, Keqing He, Yong Jiang, and Junxian He.
\newblock What makes good data for alignment? a comprehensive study of automatic data selection in instruction tuning.
\newblock In \emph{The Twelfth International Conference on Learning Representations}, 2024{\natexlab{c}}.
\newblock URL \url{https://openreview.net/forum?id=BTKAeLqLMw}.

\bibitem[Liu et~al.(2023{\natexlab{b}})Liu, Iter, Xu, Wang, Xu, and Zhu]{liu2023geval}
Yang Liu, Dan Iter, Yichong Xu, Shuohang Wang, Ruochen Xu, and Chenguang Zhu.
\newblock G-eval: Nlg evaluation using gpt-4 with better human alignment, 2023{\natexlab{b}}.
\newblock URL \url{https://arxiv.org/abs/2303.16634}.

\bibitem[Liu et~al.(2023{\natexlab{c}})Liu, Singh, Freeman, Co-Reyes, and Liu]{liu2023improving}
Yixin Liu, Avi Singh, C.~Daniel Freeman, John~D. Co-Reyes, and Peter~J. Liu.
\newblock Improving large language model fine-tuning for solving math problems, 2023{\natexlab{c}}.

\bibitem[Liu and Alahi(2024)]{liu2024cosupervised}
Yuejiang Liu and Alexandre Alahi.
\newblock Co-supervised learning: Improving weak-to-strong generalization with hierarchical mixture of experts, 2024.

\bibitem[Liu et~al.(2023{\natexlab{d}})Liu, Yang, Huang, Zhang, Huang, Wei, Deng, Sun, and Zhang]{liu2023calibrating}
Yuxuan Liu, Tianchi Yang, Shaohan Huang, Zihan Zhang, Haizhen Huang, Furu Wei, Weiwei Deng, Feng Sun, and Qi~Zhang.
\newblock Calibrating llm-based evaluator, 2023{\natexlab{d}}.
\newblock URL \url{https://arxiv.org/abs/2309.13308}.

\bibitem[Lu et~al.(2022)Lu, Welleck, Hessel, Jiang, Qin, West, Ammanabrolu, and Choi]{lu2022quark}
Ximing Lu, Sean Welleck, Jack Hessel, Liwei Jiang, Lianhui Qin, Peter West, Prithviraj Ammanabrolu, and Yejin Choi.
\newblock Quark: Controllable text generation with reinforced unlearning, 2022.

\bibitem[Lu et~al.(2024)Lu, Zhou, Ren, Wang, Shi, Pan, Zhan, and Li]{lu2024mathgenie}
Zimu Lu, Aojun Zhou, Houxing Ren, Ke~Wang, Weikang Shi, Junting Pan, Mingjie Zhan, and Hongsheng Li.
\newblock Mathgenie: Generating synthetic data with question back-translation for enhancing mathematical reasoning of llms, 2024.

\bibitem[Luo et~al.(2023{\natexlab{a}})Luo, Sun, Xu, Zhao, Lou, Tao, Geng, Lin, Chen, and Zhang]{luo2023wizardmath}
Haipeng Luo, Qingfeng Sun, Can Xu, Pu~Zhao, Jianguang Lou, Chongyang Tao, Xiubo Geng, Qingwei Lin, Shifeng Chen, and Dongmei Zhang.
\newblock Wizardmath: Empowering mathematical reasoning for large language models via reinforced evol-instruct, 2023{\natexlab{a}}.

\bibitem[Luo et~al.(2023{\natexlab{b}})Luo, Lin, Liu, Shu, Zhu, Shang, and Meng]{luo2023critique}
Liangchen Luo, Zi~Lin, Yinxiao Liu, Lei Shu, Yun Zhu, Jingbo Shang, and Lei Meng.
\newblock Critique ability of large language models, 2023{\natexlab{b}}.

\bibitem[Luo et~al.(2024)Luo, Xu, Zhao, Sun, Geng, Hu, Tao, Ma, Lin, and Jiang]{luo2024wizardcoder}
Ziyang Luo, Can Xu, Pu~Zhao, Qingfeng Sun, Xiubo Geng, Wenxiang Hu, Chongyang Tao, Jing Ma, Qingwei Lin, and Daxin Jiang.
\newblock Wizardcoder: Empowering code large language models with evol-instruct.
\newblock In \emph{The Twelfth International Conference on Learning Representations}, 2024.
\newblock URL \url{https://openreview.net/forum?id=UnUwSIgK5W}.

\bibitem[Ma et~al.(2023{\natexlab{a}})Ma, Yang, Gao, Ci, Gao, Pan, and Yang]{ma2023red}
Chengdong Ma, Ziran Yang, Minquan Gao, Hai Ci, Jun Gao, Xuehai Pan, and Yaodong Yang.
\newblock Red teaming game: A game-theoretic framework for red teaming language models.
\newblock \emph{ArXiv preprint}, abs/2310.00322, 2023{\natexlab{a}}.
\newblock URL \url{https://arxiv.org/abs/2310.00322}.

\bibitem[Ma et~al.(2023{\natexlab{b}})Ma, Zhou, Liu, Yuan, Liu, You, and Yang]{ma2023lets}
Qianli Ma, Haotian Zhou, Tingkai Liu, Jianbo Yuan, Pengfei Liu, Yang You, and Hongxia Yang.
\newblock Let's reward step by step: Step-level reward model as the navigators for reasoning, 2023{\natexlab{b}}.

\bibitem[Madaan et~al.(2023)Madaan, Tandon, Gupta, Hallinan, Gao, Wiegreffe, Alon, Dziri, Prabhumoye, Yang, Gupta, Majumder, Hermann, Welleck, Yazdanbakhsh, and Clark]{madaan2023selfrefine}
Aman Madaan, Niket Tandon, Prakhar Gupta, Skyler Hallinan, Luyu Gao, Sarah Wiegreffe, Uri Alon, Nouha Dziri, Shrimai Prabhumoye, Yiming Yang, Shashank Gupta, Bodhisattwa~Prasad Majumder, Katherine Hermann, Sean Welleck, Amir Yazdanbakhsh, and Peter Clark.
\newblock Self-refine: Iterative refinement with self-feedback.
\newblock In \emph{Thirty-seventh Conference on Neural Information Processing Systems}, volume~36, pages 46534--46594, 2023.
\newblock URL \url{https://openreview.net/forum?id=S37hOerQLB}.

\bibitem[Magister et~al.(2023)Magister, Mallinson, Adamek, Malmi, and Severyn]{magister-etal-2023-teaching}
Lucie~Charlotte Magister, Jonathan Mallinson, Jakub Adamek, Eric Malmi, and Aliaksei Severyn.
\newblock Teaching small language models to reason.
\newblock In Anna Rogers, Jordan Boyd-Graber, and Naoaki Okazaki, editors, \emph{Proceedings of the 61st Annual Meeting of the Association for Computational Linguistics (Volume 2: Short Papers)}, pages 1773--1781, Toronto, Canada, 2023. Association for Computational Linguistics.
\newblock \doi{10.18653/v1/2023.acl-short.151}.
\newblock URL \url{https://aclanthology.org/2023.acl-short.151}.

\bibitem[Manakul et~al.(2023)Manakul, Liusie, and Gales]{manakul-etal-2023-selfcheckgpt}
Potsawee Manakul, Adian Liusie, and Mark Gales.
\newblock {S}elf{C}heck{GPT}: Zero-resource black-box hallucination detection for generative large language models.
\newblock In Houda Bouamor, Juan Pino, and Kalika Bali, editors, \emph{Proceedings of the 2023 Conference on Empirical Methods in Natural Language Processing}, pages 9004--9017, Singapore, 2023. Association for Computational Linguistics.
\newblock \doi{10.18653/v1/2023.emnlp-main.557}.
\newblock URL \url{https://aclanthology.org/2023.emnlp-main.557}.

\bibitem[McAleese et~al.(2024)McAleese, Pokorny, Uribe, Nitishinskaya, Trebacz, and Leike]{mcaleese2024llm}
Nat McAleese, Rai~Michael Pokorny, Juan Felipe~Ceron Uribe, Evgenia Nitishinskaya, Maja Trebacz, and Jan Leike.
\newblock Llm critics help catch llm bugs.
\newblock \emph{arXiv preprint arXiv:2407.00215}, 2024.

\bibitem[Mecklenburg et~al.(2024)Mecklenburg, Lin, Li, Holstein, Nunes, Malvar, Silva, Chandra, Aski, Yannam, Aktas, and Hendry]{mecklenburg2024injecting}
Nick Mecklenburg, Yiyou Lin, Xiaoxiao Li, Daniel Holstein, Leonardo Nunes, Sara Malvar, Bruno Silva, Ranveer Chandra, Vijay Aski, Pavan Kumar~Reddy Yannam, Tolga Aktas, and Todd Hendry.
\newblock Injecting new knowledge into large language models via supervised fine-tuning, 2024.
\newblock URL \url{https://arxiv.org/abs/2404.00213}.

\bibitem[METR(2024)]{metr2024update}
METR.
\newblock An update on our general capability evaluations.
\newblock \url{https://metr.org/blog/2024-08-06-update-on-evaluations/}, 2024.

\bibitem[Mitchell(1980)]{mitchell1980need}
Tom~M Mitchell.
\newblock The need for biases in learning generalizations.
\newblock 1980.

\bibitem[Mitra et~al.(2023)Mitra, Corro, Mahajan, Codas, Simoes, Agarwal, Chen, Razdaibiedina, Jones, Aggarwal, Palangi, Zheng, Rosset, Khanpour, and Awadallah]{mitra2023orca}
Arindam Mitra, Luciano~Del Corro, Shweti Mahajan, Andres Codas, Clarisse Simoes, Sahaj Agarwal, Xuxi Chen, Anastasia Razdaibiedina, Erik Jones, Kriti Aggarwal, Hamid Palangi, Guoqing Zheng, Corby Rosset, Hamed Khanpour, and Ahmed Awadallah.
\newblock Orca 2: Teaching small language models how to reason, 2023.

\bibitem[Mudgal et~al.(2024)Mudgal, Lee, Ganapathy, Li, Wang, Huang, Chen, Cheng, Collins, Strohman, Chen, Beutel, and Beirami]{mudgal2024controlled}
Sidharth Mudgal, Jong Lee, Harish Ganapathy, YaGuang Li, Tao Wang, Yanping Huang, Zhifeng Chen, Heng-Tze Cheng, Michael Collins, Trevor Strohman, Jilin Chen, Alex Beutel, and Ahmad Beirami.
\newblock Controlled decoding from language models, 2024.

\bibitem[Mukherjee et~al.(2023)Mukherjee, Mitra, Jawahar, Agarwal, Palangi, and Awadallah]{mukherjee2023orca}
Subhabrata Mukherjee, Arindam Mitra, Ganesh Jawahar, Sahaj Agarwal, Hamid Palangi, and Ahmed Awadallah.
\newblock Orca: Progressive learning from complex explanation traces of gpt-4, 2023.

\bibitem[Nash et~al.(1950)]{nash1950bargaining}
John~F Nash et~al.
\newblock The bargaining problem.
\newblock \emph{Econometrica}, 18\penalty0 (2):\penalty0 155--162, 1950.

\bibitem[Nash et~al.(1951)]{nash1951non}
John~F Nash et~al.
\newblock Non-cooperative games.
\newblock page 286–295, 1951.

\bibitem[Nigam and Ghani(2000)]{nigam2000analyzing}
Kamal Nigam and Rayid Ghani.
\newblock Analyzing the effectiveness and applicability of co-training.
\newblock In \emph{Proceedings of the ninth international conference on Information and knowledge management}, pages 86--93, 2000.

\bibitem[OpenAI(2023{\natexlab{a}})]{openai2023preparedness}
OpenAI.
\newblock Preparedness framework (beta), 2023{\natexlab{a}}.
\newblock URL \url{https://cdn.openai.com/openai-preparedness-framework-beta.pdf}.

\bibitem[OpenAI(2023{\natexlab{b}})]{openai2023super}
OpenAI.
\newblock Introducing superalignment, 2023{\natexlab{b}}.
\newblock URL \url{https://openai.com/index/introducing-superalignment/}.

\bibitem[OpenAI(2023{\natexlab{c}})]{openaiGPT4TechnicalReport2023}
OpenAI.
\newblock {{GPT-4 Technical Report}}.
\newblock \emph{ArXiv preprint}, abs/2303.08774, 2023{\natexlab{c}}.
\newblock URL \url{https://arxiv.org/abs/2303.08774}.

\bibitem[OpenAI(2024)]{gpt4o}
OpenAI.
\newblock Gpt-4o system card.
\newblock \url{https://openai.com/index/gpt-4o-system-card/}, 2024.

\bibitem[Ought(2017)]{factored_cognition}
Ought.
\newblock Factored cognition, 2017.
\newblock \url{https://ought.org/research/factored-cognition}.

\bibitem[Ouyang et~al.(2022)Ouyang, Wu, Jiang, Almeida, Wainwright, Mishkin, Zhang, Agarwal, Slama, Ray, Schulman, Hilton, Kelton, Miller, Simens, Askell, Welinder, Christiano, Leike, and Lowe]{ouyangTrainingLanguageModels2022b}
Long Ouyang, Jeff Wu, Xu~Jiang, Diogo Almeida, Carroll~L. Wainwright, Pamela Mishkin, Chong Zhang, Sandhini Agarwal, Katarina Slama, Alex Ray, John Schulman, Jacob Hilton, Fraser Kelton, Luke Miller, Maddie Simens, Amanda Askell, Peter Welinder, Paul Christiano, Jan Leike, and Ryan Lowe.
\newblock Training language models to follow instructions with human feedback, 2022.
\newblock URL \url{https://arxiv.org/abs/2203.02155}.

\bibitem[Pace et~al.(2024)Pace, Mallinson, Malmi, Krause, and Severyn]{pace2024west}
Aliz{\'e}e Pace, Jonathan Mallinson, Eric Malmi, Sebastian Krause, and Aliaksei Severyn.
\newblock West-of-n: Synthetic preference generation for improved reward modeling.
\newblock \emph{ArXiv preprint}, abs/2401.12086, 2024.
\newblock URL \url{https://arxiv.org/abs/2401.12086}.

\bibitem[Padmanabhan et~al.(2023)Padmanabhan, Onoe, Zhang, Durrett, and Choi]{padmanabhan2023propagating}
Shankar Padmanabhan, Yasumasa Onoe, Michael~JQ Zhang, Greg Durrett, and Eunsol Choi.
\newblock Propagating knowledge updates to {LM}s through distillation.
\newblock In \emph{Thirty-seventh Conference on Neural Information Processing Systems}, 2023.
\newblock URL \url{https://openreview.net/forum?id=DFaGf3O7jf}.

\bibitem[Pan et~al.(2023)Pan, Chan, Zou, Li, Basart, Woodside, Zhang, Emmons, and Hendrycks]{pan2023rewards}
Alexander Pan, Jun~Shern Chan, Andy Zou, Nathaniel Li, Steven Basart, Thomas Woodside, Hanlin Zhang, Scott Emmons, and Dan Hendrycks.
\newblock Do the rewards justify the means? measuring trade-offs between rewards and ethical behavior in the machiavelli benchmark.
\newblock In \emph{International Conference on Machine Learning}, pages 26837--26867. PMLR, 2023.

\bibitem[Pang et~al.(2024)Pang, Tang, Ye, Xiong, Zhang, Wang, and Chen]{pang2024selfalignment}
Xianghe Pang, Shuo Tang, Rui Ye, Yuxin Xiong, Bolun Zhang, Yanfeng Wang, and Siheng Chen.
\newblock Self-alignment of large language models via multi-agent social simulation.
\newblock In \emph{ICLR 2024 Workshop on Large Language Model (LLM) Agents}, 2024.
\newblock URL \url{https://openreview.net/forum?id=8jUdgJdxTw}.

\bibitem[Park et~al.(2023)Park, O'Brien, Cai, Morris, Liang, and Bernstein]{park2023generative}
Joon~Sung Park, Joseph~C. O'Brien, Carrie~J. Cai, Meredith~Ringel Morris, Percy Liang, and Michael~S. Bernstein.
\newblock Generative agents: Interactive simulacra of human behavior, 2023.

\bibitem[Patil et~al.(2023)Patil, Zhang, Wang, and Gonzalez]{patil2023gorilla}
Shishir~G. Patil, Tianjun Zhang, Xin Wang, and Joseph~E. Gonzalez.
\newblock Gorilla: Large language model connected with massive apis, 2023.

\bibitem[Peng et~al.(2023)Peng, Li, He, Galley, and Gao]{peng2023instruction}
Baolin Peng, Chunyuan Li, Pengcheng He, Michel Galley, and Jianfeng Gao.
\newblock Instruction tuning with gpt-4, 2023.

\bibitem[Polu and Sutskever(2020)]{polu2020generative}
Stanislas Polu and Ilya Sutskever.
\newblock Generative language modeling for automated theorem proving.
\newblock \emph{ArXiv preprint}, abs/2009.03393, 2020.
\newblock URL \url{https://arxiv.org/abs/2009.03393}.

\bibitem[Qiao et~al.(2024)Qiao, Gui, Lv, Jia, Chen, and Zhang]{qiao2024making}
Shuofei Qiao, Honghao Gui, Chengfei Lv, Qianghuai Jia, Huajun Chen, and Ningyu Zhang.
\newblock Making language models better tool learners with execution feedback, 2024.

\bibitem[Qin et~al.(2023)Qin, Liang, Ye, Zhu, Yan, Lu, Lin, Cong, Tang, Qian, Zhao, Hong, Tian, Xie, Zhou, Gerstein, Li, Liu, and Sun]{qin2023toolllm}
Yujia Qin, Shihao Liang, Yining Ye, Kunlun Zhu, Lan Yan, Yaxi Lu, Yankai Lin, Xin Cong, Xiangru Tang, Bill Qian, Sihan Zhao, Lauren Hong, Runchu Tian, Ruobing Xie, Jie Zhou, Mark Gerstein, Dahai Li, Zhiyuan Liu, and Maosong Sun.
\newblock Toolllm: Facilitating large language models to master 16000+ real-world apis, 2023.

\bibitem[Rafailov et~al.(2023)Rafailov, Sharma, Mitchell, Ermon, Manning, and Finn]{rafailovDirectPreferenceOptimization2023a}
Rafael Rafailov, Archit Sharma, Eric Mitchell, Stefano Ermon, Christopher~D. Manning, and Chelsea Finn.
\newblock Direct {{Preference Optimization}}: {{Your Language Model}} is {{Secretly}} a {{Reward Model}}, 2023.
\newblock URL \url{https://arxiv.org/abs/2305.18290}.

\bibitem[Rame et~al.(2023)Rame, Couairon, Dancette, Gaya, Shukor, Soulier, and Cord]{rame2023rewarded}
Alexandre Rame, Guillaume Couairon, Corentin Dancette, Jean-Baptiste Gaya, Mustafa Shukor, Laure Soulier, and Matthieu Cord.
\newblock Rewarded soups: towards pareto-optimal alignment by interpolating weights fine-tuned on diverse rewards.
\newblock In \emph{Thirty-seventh Conference on Neural Information Processing Systems}, 2023.
\newblock URL \url{https://openreview.net/forum?id=lSbbC2VyCu}.

\bibitem[Ramesh et~al.(2021)Ramesh, Pavlov, Goh, Gray, Voss, Radford, Chen, and Sutskever]{ramesh2021zero}
Aditya Ramesh, Mikhail Pavlov, Gabriel Goh, Scott Gray, Chelsea Voss, Alec Radford, Mark Chen, and Ilya Sutskever.
\newblock Zero-shot text-to-image generation.
\newblock In \emph{International conference on machine learning}, pages 8821--8831. Pmlr, 2021.

\bibitem[Rein et~al.(2023)Rein, Hou, Stickland, Petty, Pang, Dirani, Michael, and Bowman]{rein2023gpqa}
David Rein, Betty~Li Hou, Asa~Cooper Stickland, Jackson Petty, Richard~Yuanzhe Pang, Julien Dirani, Julian Michael, and Samuel~R Bowman.
\newblock Gpqa: A graduate-level google-proof q\&a benchmark.
\newblock \emph{arXiv preprint arXiv:2311.12022}, 2023.

\bibitem[Ren et~al.(2024)Ren, Cao, Lin, Liu, Han, Zeng, Wan, Cai, and Sun]{ren2024learning}
Mengjie Ren, Boxi Cao, Hongyu Lin, Cao Liu, Xianpei Han, Ke~Zeng, Guanglu Wan, Xunliang Cai, and Le~Sun.
\newblock Learning or self-aligning? rethinking instruction fine-tuning, 2024.
\newblock URL \url{https://arxiv.org/abs/2402.18243}.

\bibitem[Roy et~al.(2021)Roy, Posner, Barfoot, Beaudoin, Bengio, Bohg, Brock, Depatie, Fox, Koditschek, Lozano{-}P{\'{e}}rez, Mansinghka, Pal, Richards, Sadigh, Schaal, Sukhatme, Th{\'{e}}rien, Toussaint, and van~de Panne]{Embodied-Intelligence-Overview}
Nicholas Roy, Ingmar Posner, Tim~D. Barfoot, Philippe Beaudoin, Yoshua Bengio, Jeannette Bohg, Oliver Brock, Isabelle Depatie, Dieter Fox, Daniel~E. Koditschek, Tom{\'{a}}s Lozano{-}P{\'{e}}rez, Vikash Mansinghka, Christopher~J. Pal, Blake~A. Richards, Dorsa Sadigh, Stefan Schaal, Gaurav~S. Sukhatme, Denis Th{\'{e}}rien, Marc Toussaint, and Michiel van~de Panne.
\newblock From machine learning to robotics: Challenges and opportunities for embodied intelligence.
\newblock \emph{ArXiv preprint}, abs/2110.15245, 2021.
\newblock URL \url{https://arxiv.org/abs/2110.15245}.

\bibitem[Saunders et~al.(2022)Saunders, Yeh, Wu, Bills, Ouyang, Ward, and Leike]{saunders2022self}
William Saunders, Catherine Yeh, Jeff Wu, Steven Bills, Long Ouyang, Jonathan Ward, and Jan Leike.
\newblock Self-critiquing models for assisting human evaluators.
\newblock \emph{ArXiv preprint}, abs/2206.05802, 2022.
\newblock URL \url{https://arxiv.org/abs/2206.05802}.

\bibitem[Scheurer et~al.(2022)Scheurer, Campos, Chan, Chen, Cho, and Perez]{scheurer2022training}
Jérémy Scheurer, Jon~Ander Campos, Jun~Shern Chan, Angelica Chen, Kyunghyun Cho, and Ethan Perez.
\newblock Training language models with language feedback, 2022.

\bibitem[Schulman et~al.(2017)Schulman, Wolski, Dhariwal, Radford, and Klimov]{schulman2017proximal}
John Schulman, Filip Wolski, Prafulla Dhariwal, Alec Radford, and Oleg Klimov.
\newblock Proximal policy optimization algorithms, 2017.

\bibitem[Scudder(1965)]{1053799}
H.~Scudder.
\newblock Probability of error of some adaptive pattern-recognition machines.
\newblock \emph{IEEE Transactions on Information Theory}, 11\penalty0 (3):\penalty0 363--371, 1965.
\newblock \doi{10.1109/TIT.1965.1053799}.

\bibitem[Shaikh et~al.(2024)Shaikh, Lam, Hejna, Shao, Bernstein, and Yang]{shaikh2024show}
Omar Shaikh, Michelle Lam, Joey Hejna, Yijia Shao, Michael Bernstein, and Diyi Yang.
\newblock Show, don't tell: Aligning language models with demonstrated feedback.
\newblock \emph{arXiv preprint arXiv:2406.00888}, 2024.

\bibitem[Shao et~al.(2024)Shao, Wang, Zhu, Xu, Song, Zhang, Li, Wu, and Guo]{shao2024deepseekmath}
Zhihong Shao, Peiyi Wang, Qihao Zhu, Runxin Xu, Junxiao Song, Mingchuan Zhang, Y.~K. Li, Y.~Wu, and Daya Guo.
\newblock Deepseekmath: Pushing the limits of mathematical reasoning in open language models, 2024.

\bibitem[Shapley(1971)]{shapley1971cores}
Lloyd~S Shapley.
\newblock Cores of convex games.
\newblock \emph{International journal of game theory}, 1:\penalty0 11--26, 1971.

\bibitem[Sharma et~al.(2024)Sharma, Keh, Mitchell, Finn, Arora, and Kollar]{sharma2024critical}
Archit Sharma, Sedrick Keh, Eric Mitchell, Chelsea Finn, Kushal Arora, and Thomas Kollar.
\newblock A critical evaluation of ai feedback for aligning large language models.
\newblock \emph{ArXiv preprint}, abs/2402.12366, 2024.
\newblock URL \url{https://arxiv.org/abs/2402.12366}.

\bibitem[Shavit et~al.(2023)Shavit, Agarwal, Brundage, Adler, O’Keefe, Campbell, Lee, Mishkin, Eloundou, Hickey, et~al.]{shavit2023practices}
Yonadav Shavit, Sandhini Agarwal, Miles Brundage, Steven Adler, Cullen O’Keefe, Rosie Campbell, Teddy Lee, Pamela Mishkin, Tyna Eloundou, Alan Hickey, et~al.
\newblock Practices for governing agentic ai systems.
\newblock 2023.

\bibitem[Shen et~al.(2023{\natexlab{a}})Shen, Cheng, Nguyen, You, and Bing]{shen2023large}
Chenhui Shen, Liying Cheng, Xuan-Phi Nguyen, Yang You, and Lidong Bing.
\newblock Large language models are not yet human-level evaluators for abstractive summarization.
\newblock In Houda Bouamor, Juan Pino, and Kalika Bali, editors, \emph{Findings of the Association for Computational Linguistics: EMNLP 2023}, pages 4215--4233, Singapore, 2023{\natexlab{a}}. Association for Computational Linguistics.
\newblock \doi{10.18653/v1/2023.findings-emnlp.278}.
\newblock URL \url{https://aclanthology.org/2023.findings-emnlp.278}.

\bibitem[Shen et~al.(2023{\natexlab{b}})Shen, Jin, Huang, Liu, Dong, Guo, Wu, Liu, and Xiong]{shenLargeLanguageModel2023}
Tianhao Shen, Renren Jin, Yufei Huang, Chuang Liu, Weilong Dong, Zishan Guo, Xinwei Wu, Yan Liu, and Deyi Xiong.
\newblock Large {{Language Model Alignment}}: {{A Survey}}, 2023{\natexlab{b}}.
\newblock URL \url{https://arxiv.org/abs/2309.15025}.

\bibitem[Shi et~al.(2023)Shi, Chen, Misra, Scales, Dohan, Chi, Sch{\"a}rli, and Zhou]{shi2023large}
Freda Shi, Xinyun Chen, Kanishka Misra, Nathan Scales, David Dohan, Ed~H Chi, Nathanael Sch{\"a}rli, and Denny Zhou.
\newblock Large language models can be easily distracted by irrelevant context.
\newblock In \emph{International Conference on Machine Learning}, pages 31210--31227. PMLR, 2023.

\bibitem[Shi et~al.(2024)Shi, Chen, and Zhao]{shi2024saferinstruct}
Taiwei Shi, Kai Chen, and Jieyu Zhao.
\newblock Safer-instruct: Aligning language models with automated preference data, 2024.

\bibitem[Shinn et~al.(2023)Shinn, Cassano, Berman, Gopinath, Narasimhan, and Yao]{shinn2023reflexion}
Noah Shinn, Federico Cassano, Edward Berman, Ashwin Gopinath, Karthik Narasimhan, and Shunyu Yao.
\newblock Reflexion: Language agents with verbal reinforcement learning, 2023.

\bibitem[Shridhar et~al.(2023)Shridhar, Stolfo, and Sachan]{shridhar-etal-2023-distilling}
Kumar Shridhar, Alessandro Stolfo, and Mrinmaya Sachan.
\newblock Distilling reasoning capabilities into smaller language models.
\newblock In Anna Rogers, Jordan Boyd-Graber, and Naoaki Okazaki, editors, \emph{Findings of the Association for Computational Linguistics: ACL 2023}, pages 7059--7073, Toronto, Canada, 2023. Association for Computational Linguistics.
\newblock \doi{10.18653/v1/2023.findings-acl.441}.
\newblock URL \url{https://aclanthology.org/2023.findings-acl.441}.

\bibitem[Silver et~al.(2018)Silver, Hubert, Schrittwieser, Antonoglou, Lai, Guez, Lanctot, Sifre, Kumaran, Graepel, et~al.]{silver2018general}
David Silver, Thomas Hubert, Julian Schrittwieser, Ioannis Antonoglou, Matthew Lai, Arthur Guez, Marc Lanctot, Laurent Sifre, Dharshan Kumaran, Thore Graepel, et~al.
\newblock A general reinforcement learning algorithm that masters chess, shogi, and go through self-play.
\newblock \emph{Science}, 362\penalty0 (6419):\penalty0 1140--1144, 2018.

\bibitem[Singh et~al.(2024)Singh, Co-Reyes, Agarwal, Anand, Patil, Garcia, Liu, Harrison, Lee, Xu, Parisi, Kumar, Alemi, Rizkowsky, Nova, Adlam, Bohnet, Elsayed, Sedghi, Mordatch, Simpson, Gur, Snoek, Pennington, Hron, Kenealy, Swersky, Mahajan, Culp, Xiao, Bileschi, Constant, Novak, Liu, Warkentin, Bansal, Dyer, Neyshabur, Sohl-Dickstein, and Fiedel]{singh2024beyond}
Avi Singh, John~D Co-Reyes, Rishabh Agarwal, Ankesh Anand, Piyush Patil, Xavier Garcia, Peter~J Liu, James Harrison, Jaehoon Lee, Kelvin Xu, Aaron~T Parisi, Abhishek Kumar, Alexander~A Alemi, Alex Rizkowsky, Azade Nova, Ben Adlam, Bernd Bohnet, Gamaleldin~Fathy Elsayed, Hanie Sedghi, Igor Mordatch, Isabelle Simpson, Izzeddin Gur, Jasper Snoek, Jeffrey Pennington, Jiri Hron, Kathleen Kenealy, Kevin Swersky, Kshiteej Mahajan, Laura~A Culp, Lechao Xiao, Maxwell Bileschi, Noah Constant, Roman Novak, Rosanne Liu, Tris Warkentin, Yamini Bansal, Ethan Dyer, Behnam Neyshabur, Jascha Sohl-Dickstein, and Noah Fiedel.
\newblock Beyond human data: Scaling self-training for problem-solving with language models.
\newblock \emph{Transactions on Machine Learning Research}, 2024.
\newblock ISSN 2835-8856.
\newblock URL \url{https://openreview.net/forum?id=lNAyUngGFK}.
\newblock Expert Certification.

\bibitem[Singhal et~al.(2023)Singhal, Tu, Gottweis, Sayres, Wulczyn, Hou, Clark, Pfohl, {Cole-Lewis}, Neal, Schaekermann, Wang, Amin, Lachgar, Mansfield, Prakash, Green, Dominowska, y~Arcas, Tomasev, Liu, Wong, Semturs, Mahdavi, Barral, Webster, Corrado, Matias, Azizi, Karthikesalingam, and Natarajan]{singhal2023expertlevel}
Karan Singhal, Tao Tu, Juraj Gottweis, Rory Sayres, Ellery Wulczyn, Le~Hou, Kevin Clark, Stephen Pfohl, Heather {Cole-Lewis}, Darlene Neal, Mike Schaekermann, Amy Wang, Mohamed Amin, Sami Lachgar, Philip Mansfield, Sushant Prakash, Bradley Green, Ewa Dominowska, Blaise~Aguera y~Arcas, Nenad Tomasev, Yun Liu, Renee Wong, Christopher Semturs, S.~Sara Mahdavi, Joelle Barral, Dale Webster, Greg~S. Corrado, Yossi Matias, Shekoofeh Azizi, Alan Karthikesalingam, and Vivek Natarajan.
\newblock Towards expert-level medical question answering with large language models, 2023.
\newblock URL \url{https://arxiv.org/abs/2305.09617}.

\bibitem[Snell et~al.(2022)Snell, Klein, and Zhong]{snell2022learning}
Charlie Snell, Dan Klein, and Ruiqi Zhong.
\newblock Learning by distilling context.
\newblock \emph{arXiv preprint arXiv:2209.15189}, 2022.

\bibitem[Snell et~al.(2024)Snell, Lee, Xu, and Kumar]{snell2024scaling}
Charlie Snell, Jaehoon Lee, Kelvin Xu, and Aviral Kumar.
\newblock Scaling llm test-time compute optimally can be more effective than scaling model parameters.
\newblock \emph{arXiv preprint arXiv:2408.03314}, 2024.

\bibitem[Somerstep et~al.(2024)Somerstep, Polo, Banerjee, Ritov, Yurochkin, and Sun]{somerstep2024statistical}
Seamus Somerstep, Felipe~Maia Polo, Moulinath Banerjee, Ya'acov Ritov, Mikhail Yurochkin, and Yuekai Sun.
\newblock A statistical framework for weak-to-strong generalization, 2024.

\bibitem[Song et~al.(2023)Song, Yu, Li, Yu, Huang, Li, and Wang]{songPreferenceRankingOptimization2024}
Feifan Song, Bowen Yu, Minghao Li, Haiyang Yu, Fei Huang, Yongbin Li, and Houfeng Wang.
\newblock Preference {{Ranking Optimization}} for {{Human Alignment}}, 2023.
\newblock URL \url{https://arxiv.org/abs/2306.17492}.

\bibitem[Song et~al.(2024{\natexlab{a}})Song, Yu, Lang, Yu, Huang, Wang, and Li]{song-etal-2024-scaling-data}
Feifan Song, Bowen Yu, Hao Lang, Haiyang Yu, Fei Huang, Houfeng Wang, and Yongbin Li.
\newblock Scaling data diversity for fine-tuning language models in human alignment.
\newblock In Nicoletta Calzolari, Min-Yen Kan, Veronique Hoste, Alessandro Lenci, Sakriani Sakti, and Nianwen Xue, editors, \emph{Proceedings of the 2024 Joint International Conference on Computational Linguistics, Language Resources and Evaluation (LREC-COLING 2024)}, pages 14358--14369, Torino, Italia, 2024{\natexlab{a}}. ELRA and ICCL.
\newblock URL \url{https://aclanthology.org/2024.lrec-main.1251}.

\bibitem[Song et~al.(2024{\natexlab{b}})Song, Yin, Yue, Huang, Li, and Lin]{song2024trial}
Yifan Song, Da~Yin, Xiang Yue, Jie Huang, Sujian Li, and Bill~Yuchen Lin.
\newblock Trial and error: Exploration-based trajectory optimization for llm agents, 2024{\natexlab{b}}.

\bibitem[Stiennon et~al.(2020)Stiennon, Ouyang, Wu, Ziegler, Lowe, Voss, Radford, Amodei, and Christiano]{stiennonLearningSummarizeHuman2020}
Nisan Stiennon, Long Ouyang, Jeffrey Wu, Daniel~M. Ziegler, Ryan Lowe, Chelsea Voss, Alec Radford, Dario Amodei, and Paul~F. Christiano.
\newblock Learning to summarize with human feedback.
\newblock In Hugo Larochelle, Marc'Aurelio Ranzato, Raia Hadsell, Maria{-}Florina Balcan, and Hsuan{-}Tien Lin, editors, \emph{Advances in Neural Information Processing Systems 33: Annual Conference on Neural Information Processing Systems 2020, NeurIPS 2020, December 6-12, 2020, virtual}, 2020.
\newblock URL \url{https://proceedings.neurips.cc/paper/2020/hash/1f89885d556929e98d3ef9b86448f951-Abstract.html}.

\bibitem[StockFish(2023)]{stock_fish}
StockFish.
\newblock Stockfish - open source chess engine, 2023.
\newblock \url{https://stockfishchess.org/}.

\bibitem[Sun et~al.(2023{\natexlab{a}})Sun, Zhang, Mi, Wang, Liu, Cui, Wang, Liu, and Huang]{sun-etal-2023-moraldial}
Hao Sun, Zhexin Zhang, Fei Mi, Yasheng Wang, Wei Liu, Jianwei Cui, Bin Wang, Qun Liu, and Minlie Huang.
\newblock {M}oral{D}ial: A framework to train and evaluate moral dialogue systems via moral discussions.
\newblock In Anna Rogers, Jordan Boyd-Graber, and Naoaki Okazaki, editors, \emph{Proceedings of the 61st Annual Meeting of the Association for Computational Linguistics (Volume 1: Long Papers)}, pages 2213--2230, Toronto, Canada, 2023{\natexlab{a}}. Association for Computational Linguistics.
\newblock \doi{10.18653/v1/2023.acl-long.123}.
\newblock URL \url{https://aclanthology.org/2023.acl-long.123}.

\bibitem[Sun et~al.(2024{\natexlab{a}})Sun, Li, Yuan, Yuan, Li, and Liu]{sun2024critique}
Shichao Sun, Junlong Li, Weizhe Yuan, Ruifeng Yuan, Wenjie Li, and Pengfei Liu.
\newblock The critique of critique.
\newblock \emph{ArXiv preprint}, abs/2401.04518, 2024{\natexlab{a}}.
\newblock URL \url{https://arxiv.org/abs/2401.04518}.

\bibitem[Sun et~al.(2023{\natexlab{b}})Sun, Liu, Huang, Song, Zhang, Zhang, Wang, and Gai]{sun2023parrot}
Yuchong Sun, Che Liu, Jinwen Huang, Ruihua Song, Fuzheng Zhang, Di~Zhang, Zhongyuan Wang, and Kun Gai.
\newblock Parrot: Enhancing multi-turn chat models by learning to ask questions, 2023{\natexlab{b}}.

\bibitem[Sun et~al.(2023{\natexlab{c}})Sun, Shen, Zhang, Zhou, Chen, Cox, Yang, and Gan]{sun2023salmon}
Zhiqing Sun, Yikang Shen, Hongxin Zhang, Qinhong Zhou, Zhenfang Chen, David Cox, Yiming Yang, and Chuang Gan.
\newblock Salmon: Self-alignment with principle-following reward models, 2023{\natexlab{c}}.
\newblock URL \url{https://openreview.net/forum?id=xJbsmB8UMx}.

\bibitem[Sun et~al.(2023{\natexlab{d}})Sun, Shen, Zhou, Zhang, Chen, Cox, Yang, and Gan]{sun2023principle}
Zhiqing Sun, Yikang Shen, Qinhong Zhou, Hongxin Zhang, Zhenfang Chen, David Cox, Yiming Yang, and Chuang Gan.
\newblock Principle-driven self-alignment of language models from scratch with minimal human supervision.
\newblock In A.~Oh, T.~Naumann, A.~Globerson, K.~Saenko, M.~Hardt, and S.~Levine, editors, \emph{Advances in Neural Information Processing Systems}, volume~36, pages 2511--2565. Curran Associates, Inc., 2023{\natexlab{d}}.
\newblock URL \url{https://proceedings.neurips.cc/paper_files/paper/2023/file/0764db1151b936aca59249e2c1386101-Paper-Conference.pdf}.

\bibitem[Sun et~al.(2024{\natexlab{b}})Sun, Yu, Shen, Liu, Yang, Welleck, and Gan]{sun2024easytohard}
Zhiqing Sun, Longhui Yu, Yikang Shen, Weiyang Liu, Yiming Yang, Sean Welleck, and Chuang Gan.
\newblock Easy-to-hard generalization: Scalable alignment beyond human supervision, 2024{\natexlab{b}}.
\newblock URL \url{https://arxiv.org/abs/2403.09472}.

\bibitem[Tan et~al.(2024)Tan, Zhang, Liu, Zheng, Wang, and An]{tan2024true}
Weihao Tan, Wentao Zhang, Shanqi Liu, Longtao Zheng, Xinrun Wang, and Bo~An.
\newblock True knowledge comes from practice: Aligning llms with embodied environments via reinforcement learning, 2024.

\bibitem[Tan et~al.(2023)Tan, Shi, Qiu, Qu, Qi, Xu, and Qi]{tan-etal-2023-self}
Xiaoyu Tan, Shaojie Shi, Xihe Qiu, Chao Qu, Zhenting Qi, Yinghui Xu, and Yuan Qi.
\newblock Self-criticism: Aligning large language models with their understanding of helpfulness, honesty, and harmlessness.
\newblock In Mingxuan Wang and Imed Zitouni, editors, \emph{Proceedings of the 2023 Conference on Empirical Methods in Natural Language Processing: Industry Track}, pages 650--662, Singapore, 2023. Association for Computational Linguistics.
\newblock \doi{10.18653/v1/2023.emnlp-industry.62}.
\newblock URL \url{https://aclanthology.org/2023.emnlp-industry.62}.

\bibitem[Tang et~al.(2023)Tang, Deng, Lin, Han, Liang, and Sun]{tang2023toolalpaca}
Qiaoyu Tang, Ziliang Deng, Hongyu Lin, Xianpei Han, Qiao Liang, and Le~Sun.
\newblock Toolalpaca: Generalized tool learning for language models with 3000 simulated cases.
\newblock \emph{ArXiv preprint}, abs/2306.05301, 2023.
\newblock URL \url{https://arxiv.org/abs/2306.05301}.

\bibitem[Tao et~al.(2024)Tao, Lin, Chen, Li, Wu, Li, Jin, Huang, Tao, and Zhou]{tao2024survey}
Zhengwei Tao, Ting-En Lin, Xiancai Chen, Hangyu Li, Yuchuan Wu, Yongbin Li, Zhi Jin, Fei Huang, Dacheng Tao, and Jingren Zhou.
\newblock A survey on self-evolution of large language models.
\newblock \emph{arXiv preprint arXiv:2404.14387}, 2024.

\bibitem[Taori et~al.(2023)Taori, Gulrajani, Zhang, Dubois, Li, Guestrin, Liang, and Hashimoto]{alpaca}
Rohan Taori, Ishaan Gulrajani, Tianyi Zhang, Yann Dubois, Xuechen Li, Carlos Guestrin, Percy Liang, and Tatsunori~B. Hashimoto.
\newblock Stanford alpaca: An instruction-following llama model.
\newblock \url{https://github.com/tatsu-lab/stanford_alpaca}, 2023.

\bibitem[Taubenfeld et~al.(2024)Taubenfeld, Dover, Reichart, and Goldstein]{taubenfeld2024systematic}
Amir Taubenfeld, Yaniv Dover, Roi Reichart, and Ariel Goldstein.
\newblock Systematic biases in llm simulations of debates, 2024.

\bibitem[Touvron et~al.(2023)Touvron, Martin, Stone, Albert, Almahairi, Babaei, Bashlykov, Batra, Bhargava, Bhosale, et~al.]{touvron2023llama}
Hugo Touvron, Louis Martin, Kevin Stone, Peter Albert, Amjad Almahairi, Yasmine Babaei, Nikolay Bashlykov, Soumya Batra, Prajjwal Bhargava, Shruti Bhosale, et~al.
\newblock Llama 2: Open foundation and fine-tuned chat models.
\newblock \emph{ArXiv preprint}, abs/2307.09288, 2023.
\newblock URL \url{https://arxiv.org/abs/2307.09288}.

\bibitem[Tunstall et~al.(2023)Tunstall, Beeching, Lambert, Rajani, Rasul, Belkada, Huang, von Werra, Fourrier, Habib, Sarrazin, Sanseviero, Rush, and Wolf]{tunstall2023zephyr}
Lewis Tunstall, Edward Beeching, Nathan Lambert, Nazneen Rajani, Kashif Rasul, Younes Belkada, Shengyi Huang, Leandro von Werra, Clémentine Fourrier, Nathan Habib, Nathan Sarrazin, Omar Sanseviero, Alexander~M. Rush, and Thomas Wolf.
\newblock Zephyr: Direct distillation of lm alignment, 2023.

\bibitem[Uesato et~al.(2022)Uesato, Kushman, Kumar, Song, Siegel, Wang, Creswell, Irving, and Higgins]{uesato2022solving}
Jonathan Uesato, Nate Kushman, Ramana Kumar, Francis Song, Noah Siegel, Lisa Wang, Antonia Creswell, Geoffrey Irving, and Irina Higgins.
\newblock Solving math word problems with process- and outcome-based feedback, 2022.

\bibitem[Ulmer et~al.(2024)Ulmer, Mansimov, Lin, Sun, Gao, and Zhang]{ulmer2024bootstrapping}
Dennis Ulmer, Elman Mansimov, Kaixiang Lin, Justin Sun, Xibin Gao, and Yi~Zhang.
\newblock Bootstrapping llm-based task-oriented dialogue agents via self-talk, 2024.

\bibitem[von Oswald et~al.(2023)von Oswald, Niklasson, Schlegel, Kobayashi, Zucchet, Scherrer, Miller, Sandler, Vladymyrov, Pascanu, et~al.]{von2023uncovering}
Johannes von Oswald, Eyvind Niklasson, Maximilian Schlegel, Seijin Kobayashi, Nicolas Zucchet, Nino Scherrer, Nolan Miller, Mark Sandler, Max Vladymyrov, Razvan Pascanu, et~al.
\newblock Uncovering mesa-optimization algorithms in transformers.
\newblock \emph{ArXiv preprint}, abs/2309.05858, 2023.
\newblock URL \url{https://arxiv.org/abs/2309.05858}.

\bibitem[Wang et~al.(2024{\natexlab{a}})Wang, Fang, Eisner, Durme, and Su]{wang2024llms}
Boshi Wang, Hao Fang, Jason Eisner, Benjamin~Van Durme, and Yu~Su.
\newblock Llms in the imaginarium: Tool learning through simulated trial and error, 2024{\natexlab{a}}.

\bibitem[Wang et~al.(2024{\natexlab{b}})Wang, Cheng, Zhan, Li, Song, and Liu]{wang2024openchat}
Guan Wang, Sijie Cheng, Xianyuan Zhan, Xiangang Li, Sen Song, and Yang Liu.
\newblock Openchat: Advancing open-source language models with mixed-quality data.
\newblock In \emph{The Twelfth International Conference on Learning Representations}, 2024{\natexlab{b}}.
\newblock URL \url{https://openreview.net/forum?id=AOJyfhWYHf}.

\bibitem[Wang et~al.(2023{\natexlab{a}})Wang, Xie, Jiang, Mandlekar, Xiao, Zhu, Fan, and Anandkumar]{wang2023voyager}
Guanzhi Wang, Yuqi Xie, Yunfan Jiang, Ajay Mandlekar, Chaowei Xiao, Yuke Zhu, Linxi Fan, and Anima Anandkumar.
\newblock Voyager: An open-ended embodied agent with large language models, 2023{\natexlab{a}}.

\bibitem[Wang et~al.(2024{\natexlab{c}})Wang, Zhang, Du, Zhang, and Chu]{wang2024survey}
Jiahao Wang, Bolin Zhang, Qianlong Du, Jiajun Zhang, and Dianhui Chu.
\newblock A survey on data selection for llm instruction tuning, 2024{\natexlab{c}}.

\bibitem[Wang et~al.(2024{\natexlab{d}})Wang, Ren, Zhou, Lu, Luo, Shi, Zhang, Song, Zhan, and Li]{wang2024mathcoder}
Ke~Wang, Houxing Ren, Aojun Zhou, Zimu Lu, Sichun Luo, Weikang Shi, Renrui Zhang, Linqi Song, Mingjie Zhan, and Hongsheng Li.
\newblock Mathcoder: Seamless code integration in {LLM}s for enhanced mathematical reasoning.
\newblock In \emph{The Twelfth International Conference on Learning Representations}, 2024{\natexlab{d}}.
\newblock URL \url{https://openreview.net/forum?id=z8TW0ttBPp}.

\bibitem[Wang et~al.(2023{\natexlab{b}})Wang, Xu, Lan, Hu, Lan, Lee, and Lim]{wang-etal-2023-plan}
Lei Wang, Wanyu Xu, Yihuai Lan, Zhiqiang Hu, Yunshi Lan, Roy Ka-Wei Lee, and Ee-Peng Lim.
\newblock Plan-and-solve prompting: Improving zero-shot chain-of-thought reasoning by large language models.
\newblock In Anna Rogers, Jordan Boyd-Graber, and Naoaki Okazaki, editors, \emph{Proceedings of the 61st Annual Meeting of the Association for Computational Linguistics (Volume 1: Long Papers)}, pages 2609--2634, Toronto, Canada, 2023{\natexlab{b}}. Association for Computational Linguistics.
\newblock \doi{10.18653/v1/2023.acl-long.147}.
\newblock URL \url{https://aclanthology.org/2023.acl-long.147}.

\bibitem[Wang et~al.(2023{\natexlab{c}})Wang, Li, Chen, Cai, Zhu, Lin, Cao, Liu, Liu, and Sui]{wang2023large}
Peiyi Wang, Lei Li, Liang Chen, Zefan Cai, Dawei Zhu, Binghuai Lin, Yunbo Cao, Qi~Liu, Tianyu Liu, and Zhifang Sui.
\newblock Large language models are not fair evaluators, 2023{\natexlab{c}}.
\newblock URL \url{https://arxiv.org/abs/2305.17926}.

\bibitem[Wang et~al.(2024{\natexlab{e}})Wang, Li, Shao, Xu, Dai, Li, Chen, Wu, and Sui]{wang2024mathshepherd}
Peiyi Wang, Lei Li, Zhihong Shao, R.~X. Xu, Damai Dai, Yifei Li, Deli Chen, Y.~Wu, and Zhifang Sui.
\newblock Math-shepherd: Verify and reinforce llms step-by-step without human annotations, 2024{\natexlab{e}}.

\bibitem[Wang et~al.(2024{\natexlab{f}})Wang, Yu, Zhang, Qi, Sap, Neubig, Bisk, and Zhu]{wang2024sotopiapi}
Ruiyi Wang, Haofei Yu, Wenxin Zhang, Zhengyang Qi, Maarten Sap, Graham Neubig, Yonatan Bisk, and Hao Zhu.
\newblock Sotopia-$\pi$: Interactive learning of socially intelligent language agents, 2024{\natexlab{f}}.

\bibitem[Wang et~al.(2021)Wang, Liu, Xu, Zhu, and Zeng]{wang-etal-2021-want-reduce}
Shuohang Wang, Yang Liu, Yichong Xu, Chenguang Zhu, and Michael Zeng.
\newblock Want to reduce labeling cost? {GPT}-3 can help.
\newblock In \emph{Findings of the Association for Computational Linguistics: EMNLP 2021}, pages 4195--4205, Punta Cana, Dominican Republic, 2021. Association for Computational Linguistics.
\newblock \doi{10.18653/v1/2021.findings-emnlp.354}.
\newblock URL \url{https://aclanthology.org/2021.findings-emnlp.354}.

\bibitem[Wang et~al.(2023{\natexlab{d}})Wang, Yu, Tan, O'Brien, Pasunuru, {Dwivedi-Yu}, Golovneva, Zettlemoyer, {Fazel-Zarandi}, and Celikyilmaz]{wang2023shepherd}
Tianlu Wang, Ping Yu, Xiaoqing~Ellen Tan, Sean O'Brien, Ramakanth Pasunuru, Jane {Dwivedi-Yu}, Olga Golovneva, Luke Zettlemoyer, Maryam {Fazel-Zarandi}, and Asli Celikyilmaz.
\newblock Shepherd: A critic for language model generation, 2023{\natexlab{d}}.
\newblock URL \url{https://arxiv.org/abs/2308.04592}.

\bibitem[Wang et~al.(2024{\natexlab{g}})Wang, Kulikov, Golovneva, Yu, Yuan, Dwivedi-Yu, Pang, Fazel-Zarandi, Weston, and Li]{wang2024self}
Tianlu Wang, Ilia Kulikov, Olga Golovneva, Ping Yu, Weizhe Yuan, Jane Dwivedi-Yu, Richard~Yuanzhe Pang, Maryam Fazel-Zarandi, Jason Weston, and Xian Li.
\newblock Self-taught evaluators.
\newblock \emph{arXiv preprint arXiv:2408.02666}, 2024{\natexlab{g}}.

\bibitem[Wang et~al.(2024{\natexlab{h}})Wang, Duan, Yi, Yao, Zhou, Wei, Zhang, Xu, Sun, and Xie]{wangEssenceProspectInvestigation2024}
Xinpeng Wang, Shitong Duan, Xiaoyuan Yi, Jing Yao, Shanlin Zhou, Zhihua Wei, Peng Zhang, Dongkuan Xu, Maosong Sun, and Xing Xie.
\newblock On the {{Essence}} and {{Prospect}}: {{An Investigation}} of {{Alignment Approaches}} for {{Big Models}}, 2024{\natexlab{h}}.
\newblock URL \url{https://arxiv.org/abs/2403.04204}.

\bibitem[Wang and Zhou(2024)]{wang2024chain}
Xuezhi Wang and Denny Zhou.
\newblock Chain-of-thought reasoning without prompting.
\newblock \emph{ArXiv preprint}, abs/2402.10200, 2024.
\newblock URL \url{https://arxiv.org/abs/2402.10200}.

\bibitem[Wang et~al.(2023{\natexlab{e}})Wang, Wei, Schuurmans, Le, Chi, Narang, Chowdhery, and Zhou]{wang2023selfconsistency}
Xuezhi Wang, Jason Wei, Dale Schuurmans, Quoc~V Le, Ed~H. Chi, Sharan Narang, Aakanksha Chowdhery, and Denny Zhou.
\newblock Self-consistency improves chain of thought reasoning in language models.
\newblock In \emph{The Eleventh International Conference on Learning Representations}, 2023{\natexlab{e}}.
\newblock URL \url{https://openreview.net/forum?id=1PL1NIMMrw}.

\bibitem[Wang et~al.(2022)Wang, Mishra, Alipoormolabashi, Kordi, Mirzaei, Naik, Ashok, Dhanasekaran, Arunkumar, Stap, Pathak, Karamanolakis, Lai, Purohit, Mondal, Anderson, Kuznia, Doshi, Pal, Patel, Moradshahi, Parmar, Purohit, Varshney, Kaza, Verma, Puri, Karia, Doshi, Sampat, Mishra, Reddy~A, Patro, Dixit, and Shen]{DBLP:conf/emnlp/WangMAKMNADASPK22}
Yizhong Wang, Swaroop Mishra, Pegah Alipoormolabashi, Yeganeh Kordi, Amirreza Mirzaei, Atharva Naik, Arjun Ashok, Arut~Selvan Dhanasekaran, Anjana Arunkumar, David Stap, Eshaan Pathak, Giannis Karamanolakis, Haizhi Lai, Ishan Purohit, Ishani Mondal, Jacob Anderson, Kirby Kuznia, Krima Doshi, Kuntal~Kumar Pal, Maitreya Patel, Mehrad Moradshahi, Mihir Parmar, Mirali Purohit, Neeraj Varshney, Phani~Rohitha Kaza, Pulkit Verma, Ravsehaj~Singh Puri, Rushang Karia, Savan Doshi, Shailaja~Keyur Sampat, Siddhartha Mishra, Sujan Reddy~A, Sumanta Patro, Tanay Dixit, and Xudong Shen.
\newblock Super-{N}atural{I}nstructions: Generalization via declarative instructions on 1600+ {NLP} tasks.
\newblock In \emph{Proceedings of the 2022 Conference on Empirical Methods in Natural Language Processing}, pages 5085--5109, Abu Dhabi, United Arab Emirates, 2022. Association for Computational Linguistics.
\newblock URL \url{https://aclanthology.org/2022.emnlp-main.340}.

\bibitem[Wang et~al.(2023{\natexlab{f}})Wang, Kordi, Mishra, Liu, Smith, Khashabi, and Hajishirzi]{wang-etal-2023-self-instruct}
Yizhong Wang, Yeganeh Kordi, Swaroop Mishra, Alisa Liu, Noah~A. Smith, Daniel Khashabi, and Hannaneh Hajishirzi.
\newblock Self-instruct: Aligning language models with self-generated instructions.
\newblock In Anna Rogers, Jordan Boyd-Graber, and Naoaki Okazaki, editors, \emph{Proceedings of the 61st Annual Meeting of the Association for Computational Linguistics (Volume 1: Long Papers)}, pages 13484--13508, Toronto, Canada, 2023{\natexlab{f}}. Association for Computational Linguistics.
\newblock \doi{10.18653/v1/2023.acl-long.754}.
\newblock URL \url{https://aclanthology.org/2023.acl-long.754}.

\bibitem[Wang et~al.(2023{\natexlab{g}})Wang, Zhong, Li, Mi, Zeng, Huang, Shang, Jiang, and Liu]{wangAligningLargeLanguage2023a}
Yufei Wang, Wanjun Zhong, Liangyou Li, Fei Mi, Xingshan Zeng, Wenyong Huang, Lifeng Shang, Xin Jiang, and Qun Liu.
\newblock Aligning {{Large Language Models}} with {{Human}}: {{A Survey}}, 2023{\natexlab{g}}.
\newblock URL \url{https://arxiv.org/abs/2307.12966}.

\bibitem[Wang et~al.(2024{\natexlab{i}})Wang, Li, Wu, Luo, Hou, Yu, and Shang]{wang2024multistep}
Zihan Wang, Yunxuan Li, Yuexin Wu, Liangchen Luo, Le~Hou, Hongkun Yu, and Jingbo Shang.
\newblock Multi-step problem solving through a verifier: An empirical analysis on model-induced process supervision, 2024{\natexlab{i}}.

\bibitem[Wei et~al.(2022)Wei, Wang, Schuurmans, Bosma, Xia, Chi, Le, Zhou, et~al.]{wei2022chain}
Jason Wei, Xuezhi Wang, Dale Schuurmans, Maarten Bosma, Fei Xia, Ed~Chi, Quoc~V Le, Denny Zhou, et~al.
\newblock Chain-of-thought prompting elicits reasoning in large language models.
\newblock \emph{Advances in neural information processing systems}, 35:\penalty0 24824--24837, 2022.

\bibitem[Wei et~al.(2024)Wei, Yang, Song, Lu, Hu, Tran, Peng, Liu, Huang, Du, et~al.]{wei2024long}
Jerry Wei, Chengrun Yang, Xinying Song, Yifeng Lu, Nathan Hu, Dustin Tran, Daiyi Peng, Ruibo Liu, Da~Huang, Cosmo Du, et~al.
\newblock Long-form factuality in large language models.
\newblock \emph{arXiv preprint arXiv:2403.18802}, 2024.

\bibitem[Wei et~al.(2023)Wei, Wang, Liu, Ding, and Zhang]{wei2023magicoder}
Yuxiang Wei, Zhe Wang, Jiawei Liu, Yifeng Ding, and Lingming Zhang.
\newblock Magicoder: Source code is all you need, 2023.

\bibitem[Wen et~al.(2024)Wen, Zhong, Ke, Shao, Wang, and Huang]{wen2024learning}
Jiaxin Wen, Ruiqi Zhong, Pei Ke, Zhihong Shao, Hongning Wang, and Minlie Huang.
\newblock Learning task decomposition to assist humans in competitive programming.
\newblock \emph{arXiv preprint arXiv:2406.04604}, 2024.

\bibitem[Weng et~al.(2023)Weng, Zhu, Xia, Li, He, Liu, Sun, Liu, and Zhao]{weng-etal-2023-large}
Yixuan Weng, Minjun Zhu, Fei Xia, Bin Li, Shizhu He, Shengping Liu, Bin Sun, Kang Liu, and Jun Zhao.
\newblock Large language models are better reasoners with self-verification.
\newblock In Houda Bouamor, Juan Pino, and Kalika Bali, editors, \emph{Findings of the Association for Computational Linguistics: EMNLP 2023}, pages 2550--2575, Singapore, 2023. Association for Computational Linguistics.
\newblock \doi{10.18653/v1/2023.findings-emnlp.167}.
\newblock URL \url{https://aclanthology.org/2023.findings-emnlp.167}.

\bibitem[West et~al.(2023)West, Lu, Dziri, Brahman, Li, Hwang, Jiang, Fisher, Ravichander, Chandu, Newman, Koh, Ettinger, and Choi]{west2023generative}
Peter West, Ximing Lu, Nouha Dziri, Faeze Brahman, Linjie Li, Jena~D. Hwang, Liwei Jiang, Jillian Fisher, Abhilasha Ravichander, Khyathi Chandu, Benjamin Newman, Pang~Wei Koh, Allyson Ettinger, and Yejin Choi.
\newblock The generative ai paradox: "what it can create, it may not understand", 2023.
\newblock URL \url{https://arxiv.org/abs/2311.00059}.

\bibitem[Weyssow et~al.(2024)Weyssow, Kamanda, and Sahraoui]{weyssow2024codeultrafeedback}
Martin Weyssow, Aton Kamanda, and Houari Sahraoui.
\newblock Codeultrafeedback: An llm-as-a-judge dataset for aligning large language models to coding preferences, 2024.

\bibitem[Wilf(2002)]{wilf2002algorithms}
Herbert~S Wilf.
\newblock \emph{Algorithms and complexity}.
\newblock AK Peters/CRC Press, 2002.

\bibitem[Wu et~al.(2021{\natexlab{a}})Wu, Ouyang, Ziegler, Stiennon, Lowe, Leike, and Christiano]{wu2021recursively}
Jeff Wu, Long Ouyang, Daniel~M Ziegler, Nisan Stiennon, Ryan Lowe, Jan Leike, and Paul Christiano.
\newblock Recursively summarizing books with human feedback.
\newblock \emph{ArXiv preprint}, abs/2109.10862, 2021{\natexlab{a}}.
\newblock URL \url{https://arxiv.org/abs/2109.10862}.

\bibitem[Wu and Aji(2023)]{wu2023style}
Minghao Wu and Alham~Fikri Aji.
\newblock Style over substance: Evaluation biases for large language models, 2023.
\newblock URL \url{https://arxiv.org/abs/2307.03025}.

\bibitem[Wu et~al.(2024{\natexlab{a}})Wu, Waheed, Zhang, Abdul-Mageed, and Aji]{wu-etal-2024-lamini}
Minghao Wu, Abdul Waheed, Chiyu Zhang, Muhammad Abdul-Mageed, and Alham Aji.
\newblock {L}a{M}ini-{LM}: A diverse herd of distilled models from large-scale instructions.
\newblock In Yvette Graham and Matthew Purver, editors, \emph{Proceedings of the 18th Conference of the European Chapter of the Association for Computational Linguistics (Volume 1: Long Papers)}, pages 944--964, St. Julian{'}s, Malta, 2024{\natexlab{a}}. Association for Computational Linguistics.
\newblock URL \url{https://aclanthology.org/2024.eacl-long.57}.

\bibitem[Wu et~al.(2021{\natexlab{b}})Wu, Li, and Yu]{wu2021textgail}
Qingyang Wu, Lei Li, and Zhou Yu.
\newblock Textgail: Generative adversarial imitation learning for text generation.
\newblock In \emph{Thirty-Fifth {AAAI} Conference on Artificial Intelligence, {AAAI} 2021, Thirty-Third Conference on Innovative Applications of Artificial Intelligence, {IAAI} 2021, The Eleventh Symposium on Educational Advances in Artificial Intelligence, {EAAI} 2021, Virtual Event, February 2-9, 2021}, pages 14067--14075. {AAAI} Press, 2021{\natexlab{b}}.
\newblock URL \url{https://ojs.aaai.org/index.php/AAAI/article/view/17656}.

\bibitem[Wu et~al.(2024{\natexlab{b}})Wu, Li, and Liu]{wu2024progress}
Ting Wu, Xuefeng Li, and Pengfei Liu.
\newblock Progress or regress? self-improvement reversal in post-training.
\newblock In \emph{AI for Math Workshop @ ICML 2024}, 2024{\natexlab{b}}.
\newblock URL \url{https://openreview.net/forum?id=MG18DR2dAN}.

\bibitem[Wu et~al.(2023{\natexlab{a}})Wu, Yao, Chen, Pan, Wang, Liu, and Yu]{wu2024language}
Xuansheng Wu, Wenlin Yao, Jianshu Chen, Xiaoman Pan, Xiaoyang Wang, Ninghao Liu, and Dong Yu.
\newblock From language modeling to instruction following: Understanding the behavior shift in llms after instruction tuning, 2023{\natexlab{a}}.
\newblock URL \url{https://arxiv.org/abs/2310.00492}.

\bibitem[Wu et~al.(2023{\natexlab{b}})Wu, Hu, Shi, Dziri, Suhr, Ammanabrolu, Smith, Ostendorf, and Hajishirzi]{wu2023finegrained}
Zeqiu Wu, Yushi Hu, Weijia Shi, Nouha Dziri, Alane Suhr, Prithviraj Ammanabrolu, Noah~A. Smith, Mari Ostendorf, and Hannaneh Hajishirzi.
\newblock Fine-grained human feedback gives better rewards for language model training.
\newblock In \emph{Thirty-seventh Conference on Neural Information Processing Systems}, 2023{\natexlab{b}}.
\newblock URL \url{https://openreview.net/forum?id=CSbGXyCswu}.

\bibitem[Xi et~al.(2024)Xi, Ding, Chen, Hong, Guo, Wang, Yang, Liao, Guo, He, Gao, Chen, Zheng, Zou, Gui, Zhang, Qiu, Huang, Wu, and Jiang]{xi2024agentgym}
Zhiheng Xi, Yiwen Ding, Wenxiang Chen, Boyang Hong, Honglin Guo, Junzhe Wang, Dingwen Yang, Chenyang Liao, Xin Guo, Wei He, Songyang Gao, Lu~Chen, Rui Zheng, Yicheng Zou, Tao Gui, Qi~Zhang, Xipeng Qiu, Xuanjing Huang, Zuxuan Wu, and Yu-Gang Jiang.
\newblock Agentgym: Evolving large language model-based agents across diverse environments.
\newblock \emph{ArXiv preprint}, abs/2406.04151, 2024.
\newblock URL \url{https://arxiv.org/abs/2406.04151}.

\bibitem[Xiang et~al.(2023)Xiang, Tao, Gu, Shu, Wang, Yang, and Hu]{xiang2023language}
Jiannan Xiang, Tianhua Tao, Yi~Gu, Tianmin Shu, Zirui Wang, Zichao Yang, and Zhiting Hu.
\newblock Language models meet world models: Embodied experiences enhance language models, 2023.

\bibitem[Xie et~al.(2023)Xie, Kawaguchi, Zhao, Zhao, Kan, He, and Xie]{xie2023selfevaluation}
Yuxi Xie, Kenji Kawaguchi, Yiran Zhao, Xu~Zhao, Min-Yen Kan, Junxian He, and Qizhe Xie.
\newblock Self-evaluation guided beam search for reasoning.
\newblock In \emph{Thirty-seventh Conference on Neural Information Processing Systems}, 2023.
\newblock URL \url{https://openreview.net/forum?id=Bw82hwg5Q3}.

\bibitem[Xu et~al.(2023{\natexlab{a}})Xu, Sun, Zheng, Geng, Zhao, Feng, Tao, and Jiang]{xu2023wizardlm}
Can Xu, Qingfeng Sun, Kai Zheng, Xiubo Geng, Pu~Zhao, Jiazhan Feng, Chongyang Tao, and Daxin Jiang.
\newblock Wizardlm: Empowering large language models to follow complex instructions.
\newblock \emph{ArXiv preprint}, abs/2304.12244, 2023{\natexlab{a}}.
\newblock URL \url{https://arxiv.org/abs/2304.12244}.

\bibitem[Xu et~al.(2024{\natexlab{a}})Xu, Sun, Zheng, Geng, Zhao, Feng, Tao, Lin, and Jiang]{xu2024wizardlm}
Can Xu, Qingfeng Sun, Kai Zheng, Xiubo Geng, Pu~Zhao, Jiazhan Feng, Chongyang Tao, Qingwei Lin, and Daxin Jiang.
\newblock Wizard{LM}: Empowering large pre-trained language models to follow complex instructions.
\newblock In \emph{The Twelfth International Conference on Learning Representations}, 2024{\natexlab{a}}.
\newblock URL \url{https://openreview.net/forum?id=CfXh93NDgH}.

\bibitem[Xu et~al.(2023{\natexlab{b}})Xu, Guo, Duan, and McAuley]{xu-etal-2023-baize}
Canwen Xu, Daya Guo, Nan Duan, and Julian McAuley.
\newblock Baize: An open-source chat model with parameter-efficient tuning on self-chat data.
\newblock In Houda Bouamor, Juan Pino, and Kalika Bali, editors, \emph{Proceedings of the 2023 Conference on Empirical Methods in Natural Language Processing}, pages 6268--6278, Singapore, 2023{\natexlab{b}}. Association for Computational Linguistics.
\newblock \doi{10.18653/v1/2023.emnlp-main.385}.
\newblock URL \url{https://aclanthology.org/2023.emnlp-main.385}.

\bibitem[Xu et~al.(2023{\natexlab{c}})Xu, Chern, Chern, Zhang, Wang, Liu, Li, Fu, and Liu]{xu2023align}
Chunpu Xu, Steffi Chern, Ethan Chern, Ge~Zhang, Zekun Wang, Ruibo Liu, Jing Li, Jie Fu, and Pengfei Liu.
\newblock Align on the fly: Adapting chatbot behavior to established norms.
\newblock \emph{ArXiv preprint}, abs/2312.15907, 2023{\natexlab{c}}.
\newblock URL \url{https://arxiv.org/abs/2312.15907}.

\bibitem[Xu et~al.(2024{\natexlab{b}})Xu, Zhang, Li, Liu, Lan, and Kong]{xu2024sinvig}
Jie Xu, Hanbo Zhang, Xinghang Li, Huaping Liu, Xuguang Lan, and Tao Kong.
\newblock Sinvig: A self-evolving interactive visual agent for human-robot interaction, 2024{\natexlab{b}}.

\bibitem[Xu et~al.(2024{\natexlab{c}})Xu, Li, Tao, Shen, Cheng, Li, Xu, Tao, and Zhou]{xu2024survey}
Xiaohan Xu, Ming Li, Chongyang Tao, Tao Shen, Reynold Cheng, Jinyang Li, Can Xu, Dacheng Tao, and Tianyi Zhou.
\newblock A survey on knowledge distillation of large language models, 2024{\natexlab{c}}.

\bibitem[Xue et~al.(2023)Xue, Fu, Zhou, Zheng, and You]{xue2023repeat}
Fuzhao Xue, Yao Fu, Wangchunshu Zhou, Zangwei Zheng, and Yang You.
\newblock To repeat or not to repeat: Insights from scaling llm under token-crisis.
\newblock In A.~Oh, T.~Neumann, A.~Globerson, K.~Saenko, M.~Hardt, and S.~Levine, editors, \emph{Advances in Neural Information Processing Systems}, volume~36, pages 59304--59322. Curran Associates, Inc., 2023.
\newblock URL \url{https://proceedings.neurips.cc/paper_files/paper/2023/file/b9e472cd579c83e2f6aa3459f46aac28-Paper-Conference.pdf}.

\bibitem[Yang et~al.(2023{\natexlab{a}})Yang, Wang, Lu, Liu, Le, Zhou, and Chen]{yang2023large}
Chengrun Yang, Xuezhi Wang, Yifeng Lu, Hanxiao Liu, Quoc~V Le, Denny Zhou, and Xinyun Chen.
\newblock Large language models as optimizers.
\newblock \emph{ArXiv preprint}, abs/2309.03409, 2023{\natexlab{a}}.
\newblock URL \url{https://arxiv.org/abs/2309.03409}.

\bibitem[Yang et~al.(2024{\natexlab{a}})Yang, Liu, Xie, Zhang, Song, Huang, Kuang, and Ananiadou]{yang2024metaaligner}
Kailai Yang, Zhiwei Liu, Qianqian Xie, Tianlin Zhang, Nirui Song, Jimin Huang, Ziyan Kuang, and Sophia Ananiadou.
\newblock Metaaligner: Conditional weak-to-strong correction for generalizable multi-objective alignment of language models, 2024{\natexlab{a}}.

\bibitem[Yang et~al.(2024{\natexlab{b}})Yang, Klein, Celikyilmaz, Peng, and Tian]{yang2024rlcd}
Kevin Yang, Dan Klein, Asli Celikyilmaz, Nanyun Peng, and Yuandong Tian.
\newblock {RLCD}: Reinforcement learning from contrastive distillation for {LM} alignment.
\newblock In \emph{The Twelfth International Conference on Learning Representations}, 2024{\natexlab{b}}.
\newblock URL \url{https://openreview.net/forum?id=v3XXtxWKi6}.

\bibitem[Yang et~al.(2023{\natexlab{b}})Yang, Song, Li, Zhao, Ge, Li, and Shan]{NEURIPS2023_e3936777}
Rui Yang, Lin Song, Yanwei Li, Sijie Zhao, Yixiao Ge, Xiu Li, and Ying Shan.
\newblock Gpt4tools: Teaching large language model to use tools via self-instruction.
\newblock In A.~Oh, T.~Neumann, A.~Globerson, K.~Saenko, M.~Hardt, and S.~Levine, editors, \emph{Advances in Neural Information Processing Systems}, volume~36, pages 71995--72007. Curran Associates, Inc., 2023{\natexlab{b}}.
\newblock URL \url{https://proceedings.neurips.cc/paper_files/paper/2023/file/e393677793767624f2821cec8bdd02f1-Paper-Conference.pdf}.

\bibitem[Yang et~al.(2024{\natexlab{c}})Yang, Pan, Luo, Qiu, Zhong, Yu, and Chen]{yang2024rewardsincontext}
Rui Yang, Xiaoman Pan, Feng Luo, Shuang Qiu, Han Zhong, Dong Yu, and Jianshu Chen.
\newblock Rewards-in-context: Multi-objective alignment of foundation models with dynamic preference adjustment, 2024{\natexlab{c}}.

\bibitem[Yang et~al.(2024{\natexlab{d}})Yang, Shen, Shen, Gong, and Lin]{yang2024superficialalignment}
Wenkai Yang, Shiqi Shen, Guangyao Shen, Zhi Gong, and Yankai Lin.
\newblock Super(ficial)-alignment: Strong models may deceive weak models in weak-to-strong generalization, 2024{\natexlab{d}}.

\bibitem[Yang and Wang(2020)]{yang2020overview}
Yaodong Yang and Jun Wang.
\newblock An overview of multi-agent reinforcement learning from game theoretical perspective.
\newblock \emph{ArXiv preprint}, abs/2011.00583, 2020.
\newblock URL \url{https://arxiv.org/abs/2011.00583}.

\bibitem[Yang et~al.(2024{\natexlab{e}})Yang, Li, Yan, Zhang, Huang, and Liu]{yang2024react}
Zonghan Yang, Peng Li, Ming Yan, Ji~Zhang, Fei Huang, and Yang Liu.
\newblock React meets actre: When language agents enjoy training data autonomy, 2024{\natexlab{e}}.

\bibitem[Yang et~al.(2024{\natexlab{f}})Yang, Liu, Liu, Liu, Xiong, Wang, Yang, Hu, Chen, Zhang, et~al.]{yang2024towards}
Zonghan Yang, An~Liu, Zijun Liu, Kaiming Liu, Fangzhou Xiong, Yile Wang, Zeyuan Yang, Qingyuan Hu, Xinrui Chen, Zhenhe Zhang, et~al.
\newblock Towards unified alignment between agents, humans, and environment.
\newblock \emph{ArXiv preprint}, abs/2402.07744, 2024{\natexlab{f}}.
\newblock URL \url{https://arxiv.org/abs/2402.07744}.

\bibitem[Yao et~al.(2023{\natexlab{a}})Yao, Yi, Wang, Wang, and Xie]{yaoInstructionsIntrinsicHuman2023a}
Jing Yao, Xiaoyuan Yi, Xiting Wang, Jindong Wang, and Xing Xie.
\newblock From {{Instructions}} to {{Intrinsic Human Values}} -- {{A Survey}} of {{Alignment Goals}} for {{Big Models}}, 2023{\natexlab{a}}.
\newblock URL \url{https://arxiv.org/abs/2308.12014}.

\bibitem[Yao et~al.(2022)Yao, Zhao, Yu, Du, Shafran, Narasimhan, and Cao]{yao2022react}
Shunyu Yao, Jeffrey Zhao, Dian Yu, Nan Du, Izhak Shafran, Karthik~R Narasimhan, and Yuan Cao.
\newblock React: Synergizing reasoning and acting in language models.
\newblock In \emph{The Eleventh International Conference on Learning Representations}, 2022.

\bibitem[Yao et~al.(2023{\natexlab{b}})Yao, Yu, Zhao, Shafran, Griffiths, Cao, and Narasimhan]{yao2023tree}
Shunyu Yao, Dian Yu, Jeffrey Zhao, Izhak Shafran, Thomas~L. Griffiths, Yuan Cao, and Karthik~R Narasimhan.
\newblock Tree of thoughts: Deliberate problem solving with large language models.
\newblock In \emph{Thirty-seventh Conference on Neural Information Processing Systems}, 2023{\natexlab{b}}.
\newblock URL \url{https://openreview.net/forum?id=5Xc1ecxO1h}.

\bibitem[Yao et~al.(2021)Yao, Zhong, Zhang, Han, Wang, Zhang, Xiao, Zeng, Liu, and Sun]{yao2021adversarial}
Yuan Yao, Haoxi Zhong, Zhengyan Zhang, Xu~Han, Xiaozhi Wang, Kai Zhang, Chaojun Xiao, Guoyang Zeng, Zhiyuan Liu, and Maosong Sun.
\newblock Adversarial language games for advanced natural language intelligence.
\newblock In \emph{Thirty-Fifth {AAAI} Conference on Artificial Intelligence, {AAAI} 2021, Thirty-Third Conference on Innovative Applications of Artificial Intelligence, {IAAI} 2021, The Eleventh Symposium on Educational Advances in Artificial Intelligence, {EAAI} 2021, Virtual Event, February 2-9, 2021}, pages 14248--14256. {AAAI} Press, 2021.
\newblock URL \url{https://ojs.aaai.org/index.php/AAAI/article/view/17676}.

\bibitem[Yin et~al.(2023)Yin, Liu, Yin, Zhong, Bansal, Han, and Chang]{yin-etal-2023-dynosaur}
Da~Yin, Xiao Liu, Fan Yin, Ming Zhong, Hritik Bansal, Jiawei Han, and Kai-Wei Chang.
\newblock Dynosaur: A dynamic growth paradigm for instruction-tuning data curation.
\newblock In Houda Bouamor, Juan Pino, and Kalika Bali, editors, \emph{Proceedings of the 2023 Conference on Empirical Methods in Natural Language Processing}, pages 4031--4047, Singapore, 2023. Association for Computational Linguistics.
\newblock \doi{10.18653/v1/2023.emnlp-main.245}.
\newblock URL \url{https://aclanthology.org/2023.emnlp-main.245}.

\bibitem[Ying et~al.(2024)Ying, Zhang, Li, Zhou, Shao, Fei, Ma, Hong, Liu, Wang, Wang, Wu, Li, Zhou, Liu, Zhang, Zhang, Yan, Qiu, Wang, Chen, and Lin]{ying2024internlmmath}
Huaiyuan Ying, Shuo Zhang, Linyang Li, Zhejian Zhou, Yunfan Shao, Zhaoye Fei, Yichuan Ma, Jiawei Hong, Kuikun Liu, Ziyi Wang, Yudong Wang, Zijian Wu, Shuaibin Li, Fengzhe Zhou, Hongwei Liu, Songyang Zhang, Wenwei Zhang, Hang Yan, Xipeng Qiu, Jiayu Wang, Kai Chen, and Dahua Lin.
\newblock Internlm-math: Open math large language models toward verifiable reasoning, 2024.

\bibitem[Yu et~al.(2024{\natexlab{a}})Yu, Gao, and Wang]{yu2024ovm}
Fei Yu, Anningzhe Gao, and Benyou Wang.
\newblock Ovm, outcome-supervised value models for planning in mathematical reasoning, 2024{\natexlab{a}}.

\bibitem[Yu et~al.(2024{\natexlab{b}})Yu, Jiang, Shi, YU, Liu, Zhang, Kwok, Li, Weller, and Liu]{yu2024metamath}
Longhui Yu, Weisen Jiang, Han Shi, Jincheng YU, Zhengying Liu, Yu~Zhang, James Kwok, Zhenguo Li, Adrian Weller, and Weiyang Liu.
\newblock Metamath: Bootstrap your own mathematical questions for large language models.
\newblock In \emph{The Twelfth International Conference on Learning Representations}, 2024{\natexlab{b}}.
\newblock URL \url{https://openreview.net/forum?id=N8N0hgNDRt}.

\bibitem[Yu et~al.(2023)Yu, Zhuang, Zhang, Meng, Ratner, Krishna, Shen, and Zhang]{NEURIPS2023_ae9500c4}
Yue Yu, Yuchen Zhuang, Jieyu Zhang, Yu~Meng, Alexander~J Ratner, Ranjay Krishna, Jiaming Shen, and Chao Zhang.
\newblock Large language model as attributed training data generator: A tale of diversity and bias.
\newblock In A.~Oh, T.~Neumann, A.~Globerson, K.~Saenko, M.~Hardt, and S.~Levine, editors, \emph{Advances in Neural Information Processing Systems}, volume~36, pages 55734--55784. Curran Associates, Inc., 2023.
\newblock URL \url{https://proceedings.neurips.cc/paper_files/paper/2023/file/ae9500c4f5607caf2eff033c67daa9d7-Paper-Datasets_and_Benchmarks.pdf}.

\bibitem[Yu et~al.(2024{\natexlab{c}})Yu, Zhang, Shang, Huang, Xu, Zhao, Hu, and Yin]{yu2024wavecoder}
Zhaojian Yu, Xin Zhang, Ning Shang, Yangyu Huang, Can Xu, Yishujie Zhao, Wenxiang Hu, and Qiufeng Yin.
\newblock Wavecoder: Widespread and versatile enhanced instruction tuning with refined data generation, 2024{\natexlab{c}}.

\bibitem[Yuan et~al.(2023{\natexlab{a}})Yuan, Yuan, Tan, Wang, Huang, and Huang]{yuan2023rrhf}
Hongyi Yuan, Zheng Yuan, Chuanqi Tan, Wei Wang, Songfang Huang, and Fei Huang.
\newblock {RRHF}: Rank responses to align language models with human feedback.
\newblock In \emph{Thirty-seventh Conference on Neural Information Processing Systems}, 2023{\natexlab{a}}.
\newblock URL \url{https://openreview.net/forum?id=EdIGMCHk4l}.

\bibitem[Yuan et~al.(2024{\natexlab{a}})Yuan, Cui, Wang, Ding, Wang, Deng, Shan, Chen, Xie, Lin, Liu, Zhou, Peng, Liu, and Sun]{yuan2024advancing}
Lifan Yuan, Ganqu Cui, Hanbin Wang, Ning Ding, Xingyao Wang, Jia Deng, Boji Shan, Huimin Chen, Ruobing Xie, Yankai Lin, Zhenghao Liu, Bowen Zhou, Hao Peng, Zhiyuan Liu, and Maosong Sun.
\newblock Advancing llm reasoning generalists with preference trees, 2024{\natexlab{a}}.
\newblock URL \url{https://arxiv.org/abs/2404.02078}.

\bibitem[Yuan et~al.(2024{\natexlab{b}})Yuan, He, Dong, Wang, Zhao, Xia, Xu, Zhou, Li, Zhang, et~al.]{yuan2024r}
Tongxin Yuan, Zhiwei He, Lingzhong Dong, Yiming Wang, Ruijie Zhao, Tian Xia, Lizhen Xu, Binglin Zhou, Fangqi Li, Zhuosheng Zhang, et~al.
\newblock R-judge: Benchmarking safety risk awareness for llm agents.
\newblock \emph{arXiv preprint arXiv:2401.10019}, 2024{\natexlab{b}}.

\bibitem[Yuan et~al.(2024{\natexlab{c}})Yuan, Pang, Cho, Sukhbaatar, Xu, and Weston]{yuan2024self}
Weizhe Yuan, Richard~Yuanzhe Pang, Kyunghyun Cho, Sainbayar Sukhbaatar, Jing Xu, and Jason Weston.
\newblock Self-rewarding language models.
\newblock \emph{ArXiv preprint}, abs/2401.10020, 2024{\natexlab{c}}.
\newblock URL \url{https://arxiv.org/abs/2401.10020}.

\bibitem[Yuan et~al.(2023{\natexlab{b}})Yuan, Yuan, Li, Dong, Lu, Tan, Zhou, and Zhou]{yuan2023scaling}
Zheng Yuan, Hongyi Yuan, Chengpeng Li, Guanting Dong, Keming Lu, Chuanqi Tan, Chang Zhou, and Jingren Zhou.
\newblock Scaling relationship on learning mathematical reasoning with large language models, 2023{\natexlab{b}}.

\bibitem[Yue et~al.(2024{\natexlab{a}})Yue, Qu, Zhang, Fu, Huang, Sun, Su, and Chen]{yue2024mammoth}
Xiang Yue, Xingwei Qu, Ge~Zhang, Yao Fu, Wenhao Huang, Huan Sun, Yu~Su, and Wenhu Chen.
\newblock {MA}mmo{TH}: Building math generalist models through hybrid instruction tuning.
\newblock In \emph{The Twelfth International Conference on Learning Representations}, 2024{\natexlab{a}}.
\newblock URL \url{https://openreview.net/forum?id=yLClGs770I}.

\bibitem[Yue et~al.(2024{\natexlab{b}})Yue, Zheng, Zhang, and Chen]{yue2024mammoth2}
Xiang Yue, Tuney Zheng, Ge~Zhang, and Wenhu Chen.
\newblock Mammoth2: Scaling instructions from the web.
\newblock \emph{ArXiv preprint}, abs/2405.03548, 2024{\natexlab{b}}.
\newblock URL \url{https://arxiv.org/abs/2405.03548}.

\bibitem[Zelikman et~al.(2022)Zelikman, Wu, Mu, and Goodman]{NEURIPS2022_639a9a17}
Eric Zelikman, Yuhuai Wu, Jesse Mu, and Noah Goodman.
\newblock Star: Bootstrapping reasoning with reasoning.
\newblock In S.~Koyejo, S.~Mohamed, A.~Agarwal, D.~Belgrave, K.~Cho, and A.~Oh, editors, \emph{Advances in Neural Information Processing Systems}, volume~35, pages 15476--15488. Curran Associates, Inc., 2022.
\newblock URL \url{https://proceedings.neurips.cc/paper_files/paper/2022/file/639a9a172c044fbb64175b5fad42e9a5-Paper-Conference.pdf}.

\bibitem[Zelikman et~al.(2024)Zelikman, Harik, Shao, Jayasiri, Haber, and Goodman]{zelikman2024quiet}
Eric Zelikman, Georges Harik, Yijia Shao, Varuna Jayasiri, Nick Haber, and Noah~D Goodman.
\newblock Quiet-star: Language models can teach themselves to think before speaking.
\newblock \emph{ArXiv preprint}, abs/2403.09629, 2024.
\newblock URL \url{https://arxiv.org/abs/2403.09629}.

\bibitem[Zeng et~al.(2023)Zeng, Liu, Lu, Wang, Liu, Dong, and Tang]{zeng2023agenttuning}
Aohan Zeng, Mingdao Liu, Rui Lu, Bowen Wang, Xiao Liu, Yuxiao Dong, and Jie Tang.
\newblock Agenttuning: Enabling generalized agent abilities for llms, 2023.

\bibitem[Zeng et~al.(2024{\natexlab{a}})Zeng, Dai, Cheng, Wang, Hu, Chen, Du, and Xu]{zeng2024diversified}
Dun Zeng, Yong Dai, Pengyu Cheng, Longyue Wang, Tianhao Hu, Wanshun Chen, Nan Du, and Zenglin Xu.
\newblock On diversified preferences of large language model alignment, 2024{\natexlab{a}}.

\bibitem[Zeng et~al.(2024{\natexlab{b}})Zeng, Xu, Zhao, Lou, and Chen]{zeng2024automatic}
Weihao Zeng, Can Xu, Yingxiu Zhao, Jian-Guang Lou, and Weizhu Chen.
\newblock Automatic instruction evolving for large language models, 2024{\natexlab{b}}.

\bibitem[Zhang and Parkes(2023)]{zhang2023chainofthought}
Hugh Zhang and David Parkes.
\newblock Chain-of-thought reasoning is a policy improvement operator.
\newblock In \emph{NeurIPS 2023 Workshop on Instruction Tuning and Instruction Following}, 2023.
\newblock URL \url{https://openreview.net/forum?id=bH64KCBzqS}.

\bibitem[Zhang(2023)]{zhang2023graphtoolformer}
Jiawei Zhang.
\newblock Graph-toolformer: To empower llms with graph reasoning ability via prompt augmented by chatgpt, 2023.

\bibitem[Zhang et~al.(2024)Zhang, Huang, Shi, Guo, Peng, Yan, Zhou, and Qiu]{zhang2024calibrating}
Mozhi Zhang, Mianqiu Huang, Rundong Shi, Linsen Guo, Chong Peng, Peng Yan, Yaqian Zhou, and Xipeng Qiu.
\newblock Calibrating the confidence of large language models by eliciting fidelity.
\newblock \emph{ArXiv preprint}, abs/2404.02655, 2024.
\newblock URL \url{https://arxiv.org/abs/2404.02655}.

\bibitem[Zhang et~al.(2016)Zhang, Gan, and Carin]{zhang2016generating}
Yizhe Zhang, Zhe Gan, and Lawrence Carin.
\newblock Generating text via adversarial training.
\newblock In \emph{NIPS workshop on Adversarial Training}, volume~21, pages 21--32. Academia. edu, 2016.

\bibitem[Zheng et~al.(2024{\natexlab{a}})Zheng, Wang, Ji, Huang, and Peng]{zheng2024weaktostrong}
Chujie Zheng, Ziqi Wang, Heng Ji, Minlie Huang, and Nanyun Peng.
\newblock Weak-to-strong extrapolation expedites alignment, 2024{\natexlab{a}}.

\bibitem[Zheng et~al.(2024{\natexlab{b}})Zheng, Chiang, Sheng, Zhuang, Wu, Zhuang, Lin, Li, Li, Xing, et~al.]{zheng2024judging}
Lianmin Zheng, Wei-Lin Chiang, Ying Sheng, Siyuan Zhuang, Zhanghao Wu, Yonghao Zhuang, Zi~Lin, Zhuohan Li, Dacheng Li, Eric Xing, et~al.
\newblock Judging llm-as-a-judge with mt-bench and chatbot arena.
\newblock \emph{Advances in Neural Information Processing Systems}, 36:\penalty0 46595--46623, 2024{\natexlab{b}}.
\newblock URL \url{https://proceedings.neurips.cc/paper_files/paper/2023/hash/91f18a1287b398d378ef22505bf41832-Abstract-Datasets_and_Benchmarks.html}.

\bibitem[Zheng et~al.(2023)Zheng, Dou, Gao, Hua, Shen, Wang, Liu, Jin, Liu, Zhou, Xiong, Chen, Xi, Xu, Lai, Zhu, Chang, Yin, Weng, Cheng, Huang, Sun, Yan, Gui, Zhang, Qiu, and Huang]{zheng2023secrets}
Rui Zheng, Shihan Dou, Songyang Gao, Yuan Hua, Wei Shen, Binghai Wang, Yan Liu, Senjie Jin, Qin Liu, Yuhao Zhou, Limao Xiong, Lu~Chen, Zhiheng Xi, Nuo Xu, Wenbin Lai, Minghao Zhu, Cheng Chang, Zhangyue Yin, Rongxiang Weng, Wensen Cheng, Haoran Huang, Tianxiang Sun, Hang Yan, Tao Gui, Qi~Zhang, Xipeng Qiu, and Xuanjing Huang.
\newblock Secrets of rlhf in large language models part i: Ppo, 2023.

\bibitem[Zheng et~al.(2024{\natexlab{c}})Zheng, Guo, Liu, Zhang, Yao, Xu, Wang, Xi, Gui, Zhang, et~al.]{zheng2024toward}
Rui Zheng, Hongyi Guo, Zhihan Liu, Xiaoying Zhang, Yuanshun Yao, Xiaojun Xu, Zhaoran Wang, Zhiheng Xi, Tao Gui, Qi~Zhang, et~al.
\newblock Toward optimal llm alignments using two-player games.
\newblock \emph{arXiv preprint arXiv:2406.10977}, 2024{\natexlab{c}}.

\bibitem[Zheng et~al.(2024{\natexlab{d}})Zheng, Zhang, Shen, Liu, Lin, Fu, Chen, and Yue]{zheng2024opencodeinterpreter}
Tianyu Zheng, Ge~Zhang, Tianhao Shen, Xueling Liu, Bill~Yuchen Lin, Jie Fu, Wenhu Chen, and Xiang Yue.
\newblock Opencodeinterpreter: Integrating code generation with execution and refinement, 2024{\natexlab{d}}.

\bibitem[Zhong et~al.(2024)Zhong, Ma, Zhang, Yang, Chen, Zhang, Qi, and Yang]{zhong2024panacea}
Yifan Zhong, Chengdong Ma, Xiaoyuan Zhang, Ziran Yang, Haojun Chen, Qingfu Zhang, Siyuan Qi, and Yaodong Yang.
\newblock Panacea: Pareto alignment via preference adaptation for llms, 2024.

\bibitem[Zhou et~al.(2023{\natexlab{a}})Zhou, Liu, Xu, Iyer, Sun, Mao, Ma, Efrat, Yu, YU, Zhang, Ghosh, Lewis, Zettlemoyer, and Levy]{NEURIPS2023_ac662d74}
Chunting Zhou, Pengfei Liu, Puxin Xu, Srinivasan Iyer, Jiao Sun, Yuning Mao, Xuezhe Ma, Avia Efrat, Ping Yu, LILI YU, Susan Zhang, Gargi Ghosh, Mike Lewis, Luke Zettlemoyer, and Omer Levy.
\newblock Lima: Less is more for alignment.
\newblock In A.~Oh, T.~Neumann, A.~Globerson, K.~Saenko, M.~Hardt, and S.~Levine, editors, \emph{Advances in Neural Information Processing Systems}, volume~36, pages 55006--55021. Curran Associates, Inc., 2023{\natexlab{a}}.
\newblock URL \url{https://proceedings.neurips.cc/paper_files/paper/2023/file/ac662d74829e4407ce1d126477f4a03a-Paper-Conference.pdf}.

\bibitem[Zhou et~al.(2023{\natexlab{b}})Zhou, Sch{\"a}rli, Hou, Wei, Scales, Wang, Schuurmans, Cui, Bousquet, Le, and Chi]{zhou2023leasttomost}
Denny Zhou, Nathanael Sch{\"a}rli, Le~Hou, Jason Wei, Nathan Scales, Xuezhi Wang, Dale Schuurmans, Claire Cui, Olivier Bousquet, Quoc~V Le, and Ed~H. Chi.
\newblock Least-to-most prompting enables complex reasoning in large language models.
\newblock In \emph{The Eleventh International Conference on Learning Representations}, 2023{\natexlab{b}}.
\newblock URL \url{https://openreview.net/forum?id=WZH7099tgfM}.

\bibitem[Zhou et~al.(2024)Zhou, Zhang, Wang, Chen, Zhao, Sha, Sheng, Wang, and Wen]{zhou2024jiuzhang30}
Kun Zhou, Beichen Zhang, Jiapeng Wang, Zhipeng Chen, Wayne~Xin Zhao, Jing Sha, Zhichao Sheng, Shijin Wang, and Ji-Rong Wen.
\newblock Jiuzhang3.0: Efficiently improving mathematical reasoning by training small data synthesis models, 2024.

\bibitem[Zhu et~al.(2023{\natexlab{a}})Zhu, Jordan, and Jiao]{pmlr-v202-zhu23f}
Banghua Zhu, Michael Jordan, and Jiantao Jiao.
\newblock Principled reinforcement learning with human feedback from pairwise or k-wise comparisons.
\newblock In Andreas Krause, Emma Brunskill, Kyunghyun Cho, Barbara Engelhardt, Sivan Sabato, and Jonathan Scarlett, editors, \emph{Proceedings of the 40th International Conference on Machine Learning}, volume 202 of \emph{Proceedings of Machine Learning Research}, pages 43037--43067. PMLR, 2023{\natexlab{a}}.
\newblock URL \url{https://proceedings.mlr.press/v202/zhu23f.html}.

\bibitem[Zhu et~al.(2023{\natexlab{b}})Zhu, Wang, Zhang, Zhang, Huang, Gan, Zhang, and Yang]{zhu-etal-2023-solving}
Xinyu Zhu, Junjie Wang, Lin Zhang, Yuxiang Zhang, Yongfeng Huang, Ruyi Gan, Jiaxing Zhang, and Yujiu Yang.
\newblock Solving math word problems via cooperative reasoning induced language models.
\newblock In Anna Rogers, Jordan Boyd-Graber, and Naoaki Okazaki, editors, \emph{Proceedings of the 61st Annual Meeting of the Association for Computational Linguistics (Volume 1: Long Papers)}, pages 4471--4485, Toronto, Canada, 2023{\natexlab{b}}. Association for Computational Linguistics.
\newblock \doi{10.18653/v1/2023.acl-long.245}.
\newblock URL \url{https://aclanthology.org/2023.acl-long.245}.

\bibitem[Ziems et~al.(2022)Ziems, Yu, Wang, Halevy, and Yang]{ziems-etal-2022-moral}
Caleb Ziems, Jane Yu, Yi-Chia Wang, Alon Halevy, and Diyi Yang.
\newblock The moral integrity corpus: A benchmark for ethical dialogue systems.
\newblock In \emph{Proceedings of the 60th Annual Meeting of the Association for Computational Linguistics (Volume 1: Long Papers)}, pages 3755--3773, Dublin, Ireland, 2022. Association for Computational Linguistics.
\newblock \doi{10.18653/v1/2022.acl-long.261}.
\newblock URL \url{https://aclanthology.org/2022.acl-long.261}.

\end{thebibliography}

\end{document}